\documentclass[12pt]{article}
\usepackage{amsmath}
\usepackage{graphicx}
\usepackage{enumerate}
\usepackage{natbib}
\usepackage{url} 
\usepackage{pdfpages}
\RequirePackage{amsthm,amsmath,mathrsfs,multirow,morefloats,booktabs,subfigure,pdfpages,color,algorithm,array,hhline,makecell,epstopdf,algorithm}
\RequirePackage{algorithm,algpseudocode}
\algrenewcommand\algorithmicrequire{\textbf{Input:}}
\algrenewcommand\algorithmicensure{\textbf{Output:}}
\usepackage{amsmath,amssymb}

\theoremstyle{plain}
\newtheorem{thm}{Theorem}[section]
\newtheorem{defin}[thm]{Definition}
\newtheorem{lem}[thm]{Lemma}
\newtheorem{assum}[thm]{Assumption}
\newtheorem{rem}[thm]{Remark}
\newtheorem{cor}[thm]{Corollary}
\newtheorem{prop}[thm]{Proposition}
\allowdisplaybreaks[4]
\newcommand{\blind}{1}

\begin{document}

\def\spacingset#1{\renewcommand{\baselinestretch}%
{#1}\small\normalsize} \spacingset{1}


\if1\blind
{
  \title{\bf Directed degree corrected mixed membership model and estimating community memberships in directed networks}
  \author{Huan Qing\\ 
    School of Mathematics, China University of Mining and Technology}
  \maketitle
} \fi

\if0\blind
{
  \bigskip
  \bigskip
  \bigskip
  \begin{center}
    {\LARGE\bf Impact of regularization on spectral clustering under the mixed membership stochasticblock model}
\end{center}
  \medskip
} \fi

\bigskip
\begin{abstract}
\spacingset{1.2} 
This paper considers the problem of modeling and estimating community memberships of nodes in a directed network where every row (column) node is associated with a vector determining its membership in each row (column) community. To model such directed network, we propose directed degree corrected mixed membership (DiDCMM) model by considering degree heterogeneity. DiDCMM is identifiable under popular conditions for mixed membership network when considering degree heterogeneity. Based on the cone structure inherent in the normalized version of the left singular vectors and the simplex structure inherent in the right singular vectors of the population adjacency matrix, we build an efficient algorithm called DiMSC to infer the community membership vectors for both row nodes and column nodes. By taking the advantage of DiMSC's equivalence algorithm which returns same estimations as DiMSC and the recent development on row-wise singular vector deviation, we show that the proposed algorithm is asymptotically consistent under mild conditions by providing error bounds for the inferred membership vectors of each row node and each column node under DiDCMM. The theory is supplemented by a simulation study.
\end{abstract}

\noindent%
\vfill
\spacingset{1.2} 
\section{Introduction}\label{sec:intro}
In most real-world networks, a node may belong to more than one community at a time, the problem of estimating mixed memberships for undirected network has received a lot of attention \cite{MMSB,ball2011efficient,wang2011community,gopalan2013efficient,anandkumar2014a,kaufmann2017a,panov2017consistent,OCCAM,MixedSCORE,GeoNMF,MaoSVM,mao2020estimating}, and references therein. Here, we introduce partial of these works with encouraging results on models, consistency or closely related with works in this paper when estimating mixed memberships. \cite{MMSB} extends the popular model stochastic blockmodel \cite{SBM} for non-overlapping undirected networks to mixed membership undirected networks and designs the mixed membership stochastic blockmodel (MMSB). Based on the MMSB model, \cite{MixedSCORE} designs a model called degree corrected mixed membership (DCMM) model by considering degree heterogeneities, where DCMM can also be seen as an extension of the non-overlapping model degree corrected stochastic blockmodel (DCSBM) \cite{DCSBM}, and \cite{MixedSCORE} also develops an efficient and provably consistent algorithm which is an extension of the one given in \cite{SCORE} from no-overlapping community detection to the mixed membership estimating problem. \cite{mao2020estimating} presents an algorithm under MMSB and establishs per-node rates for mixed memberships by sharp row-wise eigenvector deviation. \cite{OCCAM} proposes a model OCCAM which is also an extension of MMSB by considering degree heterogeneity. To fit OCCAM, \cite{OCCAM} develops an algorithm requiring relatively small fraction of mixed nodes when building theoretical frameworks. \cite{MaoSVM} finds the cone structure inherent in the normalization of the eigen-decomposition of the population adjacency matrix under DCMM as well as OCCAM, and develops an algorithm to hunt corners in the cone structure.

Though the above works are encouraging and appealing, they focus on undirected networks. In reality, there exists substantial directed networks such as citation networks, protein-protein interaction networks and the hyperlink network of websites. Clustering directed networks is receiving more and more attentions \cite{wang1987stochastic,reichardt2007role,yang2010directed,malliaros2013clustering,DISIM,DSCORE,zhou2019analysis}.  Recent years, a lot of works with encouraging results have been developed for directed networks. \cite{DISIM} proposes Stochastic co-Blockmodel (ScBM) and its extension DC-ScBM by considering degree heterogeneity to model non-overlapping directed network, where ScBM and DCScBM can model directed networks whose row nodes may be different from column nodes and the number of row communities may also be different from the number of column communities. \cite{DSCORE} studies the theoretical guarantees for the algorithm D-SCORE \citep{ji2016coauthorship} and its variants designed under DC-ScBM when the number of row communities equals that of column communities. \cite{zhou2019analysis} studies the spectral clustering algorithms designed by a data-driven regularization of the adjacency matrix under ScBM. Based on the facts that the above works only consider non-overlapping directed networks, \cite{qing2021bipartite} develops a model bipartite mixed membership stochastic blockmodel (BiMMSB) which is an extension of ScBM and models directed networks with mixed memberships. 

In this paper, we focus on the directed network with mixed membership. Our contributions in this paper are as follows:
\begin{itemize}
   \item [(i)] We propose a novel generative model for directed networks with mixed membership, the Directed Degree Corrected Mixed Membership (DiDCMM) model. DiDCMM models a directed network with mixed memberships when row nodes have degree heterogeneities while column nodes do not.  We present the identifiability of DiDCMM under popular conditions which are also required by models modeling mixed membership network when considering degree heterogeneity. Meanwhile, our results also show that modeling a directed network with mixed membership when considering degree heterogeneity for both row and column nodes needs nontrivial conditions.
   \item [(ii)] To fit DiDCMM, we present an algorithm  called DiMSC, which is designed based on the investigation that there exist an Ideal Cone structure inherent in the normalized version of the left singular vectors and an Ideal Simplex structure inherent in the right singular vectors of the population adjacency matrix. Two existing algorithms in \cite{gillis2015semidefinite} and \cite{MaoSVM} are used to hunt for the corners in the simplex and the cone structures, respectively. We prove that our DiMSC exactly recovers the membership matrices for both row and column nodes in the oracle case under DiDCMM, and this also supports the identifiability of DiDCMM. By taking the advantage of the recent row-wise singular vector deviation \cite{chen2020spectral} and the equivalence algorithm of DiMSC, we obtain the upper bounds of error rates for each row node and each column node, and show that our method produces asymptotically consistent parameter estimations under mild conditions by delicate spectral analysis.
\end{itemize}
The article is organized as follows. Section \ref{sec2} presents the DiDCMM model and its identifiability. In Section \ref{sec3}, we present DiMSC algorithm from the oracle case to the real case, and show that DiMSC exactly recovers membership matrices in the oracle case. An equivalence algorithm of DiMSC is also presented in Section \ref{sec3}. The main results about the consistency of DiMSC are presented in Section \ref{sec4}. Section \ref{sec5} contains simulations. Section \ref{sec6} offers some final remarks. The proofs of the main results are relegated to the appendix.

\textbf{\textit{Notations.}}
We take the following general notations in this paper. For a vector $x$ and fixed $q>0$, $\|x\|_{q}$ denotes its $l_{q}$-norm. We drop the subscript if $q=2$ occasionally. For a matrix $M$, $M'$ denotes the transpose of the matrix $M$, $\|M\|$ denotes the spectral norm, $\|M\|_{F}$ denotes the Frobenius norm, and $\|M\|_{2\rightarrow\infty}$ denotes the maximum $l_{2}$-norm of all the rows of $M$. Let $\mathrm{rank}(M)$ denote the rank of matrix $M$. Let $\sigma_{i}(M)$ be the $i$-th largest singular value of matrix $M$, and $\lambda_{i}(M)$ denote the $i$-th largest eigenvalue of the matrix $M$ ordered by the magnitude. $M(i,:)$ and $M(:,j)$ denote the $i$-th row and the $j$-th column of matrix $M$, respectively. $M(S_{r},:)$ and $M(:,S_{c})$ denote the rows and columns in the index sets $S_{r}$ and $S_{c}$ of matrix $M$, respectively. For any matrix $M$, we simply use $Y=\mathrm{max}(0, M)$ to represent $Y_{ij}=\mathrm{max}(0, M_{ij})$ for any $i,j$. For any matrix $M\in\mathbb{R}^{m\times m}$, let $\mathrm{diag}(M)$ be the $m\times m$ diagonal matrix whose $i$-th diagonal entry is $M(i,i)$. $\mathbf{1}$ and $\mathbf{0}$ are  column vectors with all entries being ones and zeros, respectively.  $e_{i}$ is a column vector whose $i$-th entry is 1 while other entries are zero. In this paper, $C$ is a positive constant which may vary occasionally.

\section{The directed degree corrected  mixed membership model}\label{sec2}
Consider a directed network $\mathcal{N}=(\mathcal{V}_{r}, \mathcal{V}_{c}, \mathcal{E})$, where $\mathcal{V}_{r}=\{1,2,\ldots, n_{r}\}$ is the set of row nodes, $\mathcal{V}_{c}=\{1,2,\ldots, n_{c}\}$ is the set of column nodes ($n_r$ and $n_c$ indicate the number of row nodes and the number of column nodes, respectively), and $\mathcal{E}$ is the set of edges. We assume that the row nodes of the directed network $\mathcal{N}$ belong to $K$ perceivable communities (call row communities in this paper)
\begin{align}\label{DefinSC}
\mathcal{C}^{(1)}_{r},\mathcal{C}^{(2)}_{r},\ldots,\mathcal{C}^{(K)}_{r},
\end{align}
and the column nodes of the directed network $\mathcal{N}$ belong to $K$ perceivable communities (call column communities in this paper)
\begin{align}\label{DefinRC}
\mathcal{C}^{(1)}_{c},\mathcal{C}^{(2)}_{c},\ldots,\mathcal{C}^{(K)}_{c}.
\end{align}
Define an $n_{r}\times K$ row nodes membership matrix $\Pi_{r}$ and an $n_{c}\times K$ column nodes membership matrix $\Pi_{c}$ such that
$\Pi_{r}(i,:)$ is a $1\times K$ Probability Mass Function (PMF) for row  node $i$, $\Pi_{c}(j,:)$ is a $1\times K$ PMF for column node $j$, and
\begin{align}\label{DefineSPMF}
&\Pi_{r}(i,k)\mathrm{~is~the~weight~of~row~node~}i~\mathrm{on~}\mathcal{C}^{(k)}_{r}, 1\leq k\leq K,\\
&\Pi_{c}(j,k)\mathrm{~is~the~weight~of~column~node~}j~\mathrm{on~}\mathcal{C}^{(k)}_{c}, 1\leq k\leq K.
\end{align}
Call row node $i$ `pure' if $\Pi_{r}(i,:)$ is degenerate (i.e., one entry is 1, all others $K-1$ entries are 0) and `mixed' otherwise. Same definitions hold for column nodes.

Let $A\in \{0,1\}^{n_{r}\times n_{c}}$ be the bi-adjacency matrix of $\mathcal{N}$ such that for each entry, $A(i,j)=1$ if there is a directional edge from row node $i$ to column node $j$, and $A(i,j)=0$ otherwise. So, the $i$-th row of $A$ records how row node $i$ sends edges, and the $j$-th column of $A$ records how column node $j$ receives edges. Let $P$ be a $K\times K$ matrix such that
\begin{align}\label{ConB}
P(k,l)\geq 0\mathrm{~for~}1\leq k,l\leq K.
\end{align}
Note that since we consider directed network in this paper, $P$ may be asymmetric.

Without loss of generality, suppose that row nodes have degree heterogeneities while column nodes do not. Note that, in a directed network, if column nodes have degree heterogeneities while row nodes do not, to detect memberships of both row nodes and column nodes, set the transpose of the adjacency matrix as input when applying our algorithm DiMSC. Meanwhile, in a direct network, if both row and column nodes have degree heterogeneity, to model such directed network with mixed memberships, we need nontrivial constraints on the degree heterogeneities between row nodes and column nodes for model identifiability. Since such constraints are nontrivial, it is meaningless to model such directed network, see Remark 1 in the Supplementary Materials for detail.

Let $\theta_{r}$ be an $n_{r}\times 1$ vector whose $i$-th entry is the positive degree heterogeneity of row node $i$. For all pairs of $(i,j)$ with $1\leq i\leq n_{r},1\leq j\leq n_{c}$, DiDCMM models the entries of $A$ such that $A(i,j)$ are independent Bernoulli random variables satisfying
\begin{align}\label{DefinP}
\mathbb{P}(A(i,j)=1)=\theta_{r}(i)\sum_{k=1}^{K}\sum_{l=1}^{K}\Pi_{r}(i,k)\Pi_{c}(j,l)P(k,l).
\end{align}
Introduce the degree heterogeneity diagonal matrix $\Theta_{r}\in \mathbb{R}^{n_{r}\times n_{r}}$ for row nodes such that
\begin{align}\label{DefinThetaR}
\Theta_{r}(i,i)=\theta_{r}(i)\qquad\mathrm{for~}1\leq i\leq n_{r}.
\end{align}
\begin{defin}
Call model (\ref{DefinSC})-(\ref{DefinP}) the Directed Degree Corrected Mixed Membership (DiDCMM) model, and denote it by $DiDCMM_{n_{r},n_{c}}(K,P,\Pi_{r}, \Pi_{c}, \Theta_{r})$.
\end{defin}
The following conditions are sufficient for the identifiability of DiDCMM:
\begin{itemize}
  \item (I1) $\mathrm{rank}(P)=K$ and $P$ has unit diagonals.
  \item (I2) There is at least one pure node for each of the $K$ row and $K$ column communities.
\end{itemize}
Note that our model allows $P_{\mathrm{max}}\geq 1$ where  $P_{\mathrm{max}}=\mathrm{max}_{1\leq k,l\leq K}P(k,l)$. Similar as Eq (2.14) \cite{MixedSCORE}, let $P_{\mathrm{max}}\leq C$ for convenience. For mixed membership community detection, when considering nodes degrees, pure nodes condition for each community and condition (I1) are also requirement for model identifiability in \cite{MixedSCORE,MaoSVM,OCCAM}. Let $\Omega=\mathbb{E}[A]$ be the expectation of the adjacency matrix $A$. Under DiDCMM, we have
\begin{align*}
\Omega=\Theta_{r}\Pi_{r}P\Pi_{c}'.
\end{align*}
We refer $\Omega$ as the population adjacency matrix. For $1\leq k\leq K$, let $\mathcal{I}^{(k)}_{r}=\{i\in\{1,2,\ldots, n_{r}\}: \Pi_{r}(i,k)=1\}$ and $\mathcal{I}^{(k)}_{c}=\{j\in \{1,2,\ldots, n_{c}\}: \Pi_{c}(j,k)=1\}$. By condition (I2), $I^{(k)}_{r}$ and $I^{(k)}_{c}$ are non empty for all $1\leq k\leq K$. For $1\leq k\leq K$, select one row node from $\mathcal{I}^{(k)}_{r}$ to construct the index set $\mathcal{I}_{r}$, i.e., $\mathcal{I}_{r}$ is the indices of row nodes corresponding to $K$ pure row nodes, one from each community. And $\mathcal{I}_{c}$ is defined similarly. W.L.O.G., let $\Pi_{r}(\mathcal{I}_{r},:)=I_{K}$ and $\Pi_{c}(\mathcal{I}_{c},:)=I_{K}$ (Lemma 2.1 \cite{mao2020estimating} also has similar setting to design their spectral algorithms under MMSB.) where $I_{K}$ is the $K\times K$ identity matrix. The proposition below shows that the DiDCMM model is identifiable.
\begin{prop}\label{id}
(Identifiability).	When conditions (I1) and (I2) hold, DiDCMM is identifiable: for eligible  $(P, \Pi_{r}, \Pi_{c}, \Theta_{r})$ and $(\tilde{P}, \tilde{\Pi}_{r}, \tilde{\Pi}_{c}, \tilde{\Theta}_{r})$, set $\Omega=\Theta_{r}\Pi_{r}P\Pi_{c}'$ and $\tilde{\Omega}=\tilde{\Theta}_{r}\tilde{\Pi}_{r}\tilde{P}\tilde{\Pi}_{c}'$. If $\Omega=\tilde{\Omega}$, then $\Theta_{r}=\tilde{\Theta}_{r}, \Pi_{r}=\tilde{\Pi}_{r}, \Pi_{c}=\tilde{\Pi}_{c}$ and $P=\tilde{P}$.
\end{prop}
Unless specified, we treat conditions (I1) and (I2) as default from now on.

\section{Algorithm}\label{sec3}
The primary goal of the proposed algorithm is to estimate the row membership matrix $\Pi_{r}$ and column membership matrix $\Pi_{c}$ from the observed adjacency matrix $A$ with given $K$. We start by considering the ideal case when $\Omega$ is known, and then we extend what we learn in the ideal case to the real case.
\subsection{The Ideal Simplex (IS), the Ideal Cone (IC) and the Ideal DiMSC}
Note that $\mathrm{rank}(\Omega)=K$ under conditions (I1) and (I2),  and $K$ is much smaller than $\mathrm{min}\{n_{r}, n_{c}\}$. Let $\Omega=U\Lambda V'$ be the compact singular value decomposition of $\Omega$ such that  $U\in\mathbb{R}^{n_{r}\times K}, \Lambda\in\mathbb{R}^{K\times K}, V\in\mathbb{R}^{n_{c}\times K}$, $U'U=I_{K}, V'V=I_{K}$. The goal of the ideal case is to use $U, \Lambda$ and $V$ to exactly recover $\Pi_{r}$ and $\Pi_{c}$. As stated in \cite{MixedSCORE}, $\theta_{r}$ is one of the major nuisance, and similar as \cite{RSC}, we remove the effect of $\theta_{r}$ by normalizing each rows of $U$ to have unit $l_{2}$ norm. Set $U_{*}\in \mathbb{R}^{n_{r}\times K}$ by $U_{*}(i,:)=\frac{U(i,:)}{\|U(i,:)\|_{F}}$ and let $N_{U}$ be the $n_{r}\times n_{r}$ diagonal matrix such that $N_{U}(i,i)=\frac{1}{\|U(i,:)\|_{F}}$ for $1\leq i\leq n_{r}$. Then $U_{*}$ can be rewritten as $U_{*}=N_{U}U$. The existences of the Ideal Cone (IC for short) structure inherent in $U_{*}$ and the Ideal Simplex (IS for short) structure inherent in $V$ are guarantee by the following lemma.
\begin{lem}\label{ISIC}
(Ideal Simplex and Ideal Cone). Under $DiDCMM_{n_{r},n_{c}}(K,P,\Pi_{r}, \Pi_{c}, \Theta_{r})$, there exist an unique $K\times K$ matrix $B_{r}$ and an unique $K\times K$ matrix $B_{c}$ such that
\begin{itemize}
\item $U=\Theta_{r}\Pi_{r}B_{r}$ where $B_{r}=\Theta^{-1}_{r}(\mathcal{I}_{r},\mathcal{I}_{r})U(\mathcal{I}_{r},:)$, and $U_{*}=YU_{*}(\mathcal{I}_{r},:)$
where $Y=N_{M}\Pi_{r}\Theta^{-1}_{r}(\mathcal{I}_{r},\mathcal{I}_{r})N_{U}^{-1}(\mathcal{I}_{r},\mathcal{I}_{r})$ with $N_{M}$ being an $n_{r}\times n_{r}$ diagonal matrix whose diagonal entries are positive. Meanwhile, $U_{*}(i,:)=U_{*}(\bar{i},:)$ if $\Pi_{r}(i,:)=\Pi_{r}(\bar{i},:)$ for $1\leq i,\bar{i}\leq n_{r}$.
  \item $V=\Pi_{c}B_{c}$ where $B_{c}=V(\mathcal{I}_{c},:)$. Meanwhile, $V(j,:)=V(\bar{j},:)$ if $\Pi_{c}(j,:)=\Pi_{c}(\bar{j},:)$ for $1\leq j,\bar{j}\leq n_{c}$.
\end{itemize}
\end{lem}
Lemma \ref{ISIC} says that the rows of $V$ form a $K$-simplex in $\mathbb{R}^{K}$ which we call the Ideal Simplex (IS), with the $K$ rows of $B_{c}$ being the vertices. Such IS is also found in \cite{MixedSCORE,mao2020estimating}. Lemma \ref{ISIC} also shows that the form of $U_{*}=YU_{*}(\mathcal{I}_{r},:)$ is actually the Ideal Cone structure mentioned in \cite{MaoSVM}.

For column nodes (recall that column nodes have no degree heterogeneities), since $B_{c}$ is full rank, if $V$ and $B_{c}$ are known in advance ideally, we can exactly recover  $\Pi_{c}$ by setting $\Pi_{c}=VB_{c}'(B_{c}B_{c}')^{-1}\equiv VB^{-1}_{c}$. For convenience to transfer the ideal case to the real case, set
$Z_{c}=VB^{-1}_{c}$. Since $Z_{c}\equiv\Pi_{c}$, we have
\begin{align*}
\Pi_{c}(j,:)=\frac{Z_{c}(j,:)}{\|Z_{c}(j,:)\|_{1}},1\leq j\leq n_{c}.
\end{align*}
With given $V$, since it enjoys IS structure $V=\Pi_{c}B_{c}\equiv \Pi_{c}V(\mathcal{I}_{c},:)$, as long as we can obtain $V(\mathcal{I}_{c},:)$ (i.e., $B_{c}$), we can recover $\Pi_{c}$ exactly. As mentioned in \cite{MixedSCORE,mao2020estimating}, for such IS, the successive projection (SP) algorithm \cite{gillis2015semidefinite} (i.e., Algorithm 1 in the Supplementary Materials) can be applied to $V$ with $K$ column communities to find the column corner matrix $B_{c}$. The above analysis gives how to recover $\Pi_{c}$ with given $\Omega$ and $K$ under DiDCMM ideally.

Next, we aim to recover $\Pi_{r}$ from $U$ with given $K$. Since $\mathrm{rank}(U_{*})=K$, $\mathrm{rank}(U_{*}(\mathcal{I}_{r},:))=K$. As $U_{*}(\mathcal{I}_{r},:)\in\mathbb{R}^{K\times K}$, the inverse of $U_{*}(\mathcal{I}_{r},:)$ exists. Therefore, Lemma \ref{ISIC} also gives that
\begin{align}\label{Y1}
Y=U_{*}U^{-1}_{*}(\mathcal{I}_{r},:).
\end{align}
Since $U_{*}=N_{U}U$ and $Y=N_{M}\Pi_{r}\Theta^{-1}_{r}(\mathcal{I}_{r},\mathcal{I}_{r})N_{U}^{-1}(\mathcal{I}_{r},\mathcal{I}_{r})$, we have $N_{U}^{-1}N_{M}\Pi_{r}\Theta^{-1}_{r}(\mathcal{I}_{r},\mathcal{I}_{r})N_{U}^{-1}(\mathcal{I}_{r},\mathcal{I}_{r})=UU^{-1}_{*}(\mathcal{I}_{r},:)$, i.e.,
\begin{align}\label{ZYJ1}
N_{U}^{-1}N_{M}\Pi_{r}=UU^{-1}_{*}(\mathcal{I}_{r},:)N_{U}(\mathcal{I}_{r},\mathcal{I}_{r})\Theta_{r}(\mathcal{I}_{r},\mathcal{I}_{r}).
\end{align}
By the proof of Lemma \ref{id}, we know that $\Theta_{r}(\mathcal{I}_{r},\mathcal{I}_{r})=\mathrm{diag}(U(\mathcal{I}_{r},:)\Lambda V'(\mathcal{I}_{c},:))$ when condition (I1) holds.

For convenience, set  $J_{*}=N_{U}(\mathcal{I}_{r},\mathcal{I}_{r})\Theta_{r}(\mathcal{I}_{r},\mathcal{I}_{r})\equiv\mathrm{diag}(U_{*}(\mathcal{I}_{r},:)\Lambda V'(\mathcal{I}_{c},:)),
Z_{r}=N_{U}^{-1}N_{M}\Pi_{r}, Y_{*}=UU^{-1}_{*}(\mathcal{I}_{r},:)$. By Eq (\ref{ZYJ1}), we have
\begin{align}\label{Z1}
Z_{r}=Y_{*}J_{*}\equiv UU^{-1}_{*}(\mathcal{I}_{r},:)\mathrm{diag}(U_{*}(\mathcal{I}_{r},:)\Lambda V'(\mathcal{I}_{c},:)).
\end{align}
Meanwhile, since $N_{U}^{-1}N_{M}$ is an $n_{r}\times n_{r}$ positive diagonal matrix, we have
\begin{align}\label{Pi}
\Pi_{r}(i,:)=\frac{Z_{r}(i,:)}{\|Z_{r}(i,:)\|_{1}}, 1\leq i\leq n_{r}.
\end{align}
With given $\Omega$ and $K$, we can obtain $U,V$, thus the above analysis shows that once the two index sets $\mathcal{I}_{r}$ and $\mathcal{I}_{c}$ are known, we can exactly recover $\Pi_{r}$ by Eq. (\ref{Pi}) and Eq. (\ref{Z1}).

\begin{figure}
	\centering
	\subfigure[$U_{*}$: Ideal Cone]{\includegraphics[width=0.45\textwidth]{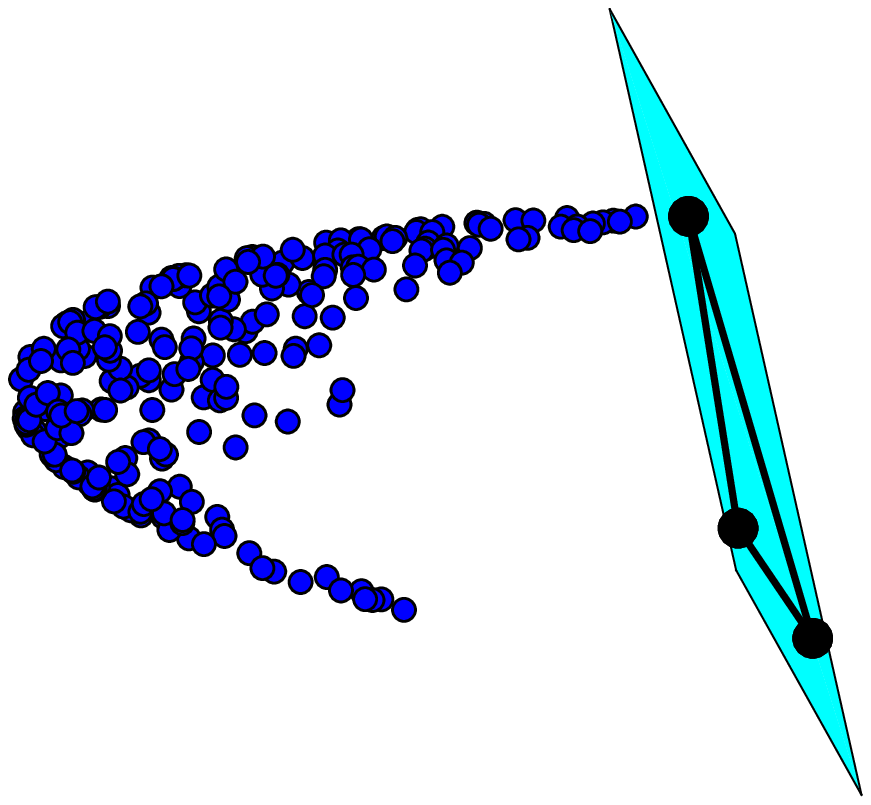}}
	\subfigure[$V$: Ideal Simplex]{\includegraphics[width=0.45\textwidth]{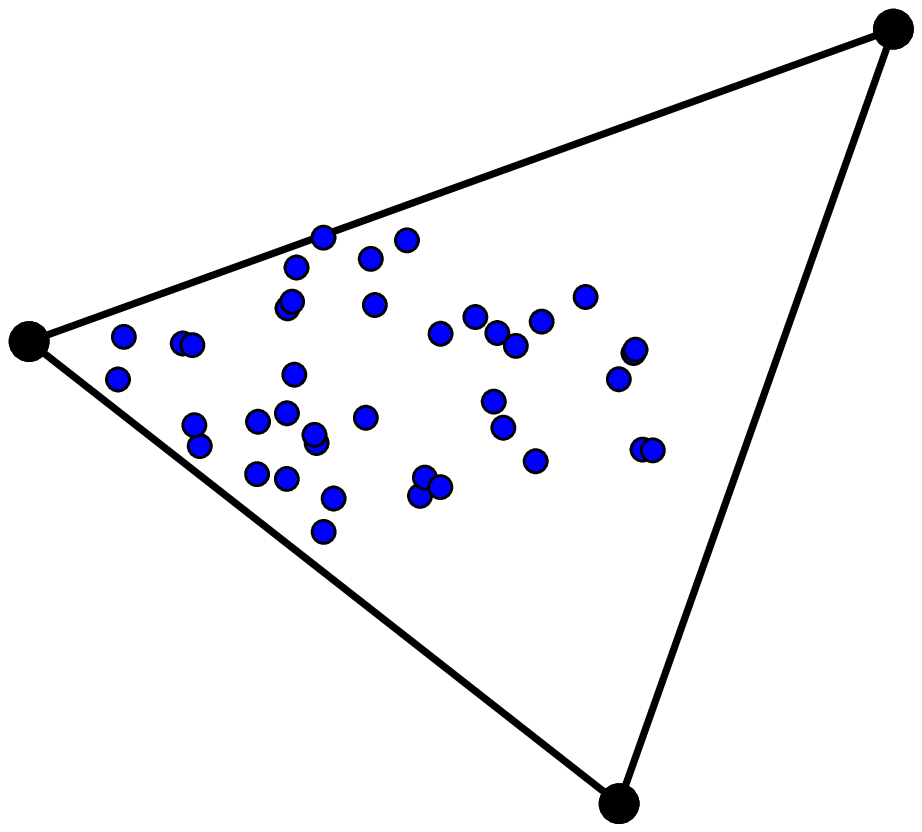}}
\caption{Panel (a): plot of $U_{*}$ and the hyperplane formed by $U_{*}(\mathcal{I}_{r},:)$. Blue points denote rows respective to mixed row nodes of $U_{*}$, and black points denote the $K$ rows of the corner matrix $U_{*}(\mathcal{I}_{r},:)$. The plane in panel (a) is the hyperplane formed by the triangle of the 3 rows of $U_{*}(\mathcal{I}_{r},:)$. Panel (b): plot of $V$ and the Ideal Simplex formed by $V(\mathcal{I}_{c},:)$. Blue points denote rows respective to mixed column nodes of $V$, and black points denote the $K$ rows of the corner matrix $V(\mathcal{I}_{c},:)$. Since $K=3$, for visualization, we have projected these points from $\mathbb{R}^{3}$ to $\mathbb{R}^{2}$.}
\label{PlotUstarV}
\end{figure}

With given $\Omega$ and $K$, to recover $\Pi_{r}$ in the ideal case, we need to obtain $Z_{r}$ by Eq. (\ref{Z1}), which means that the only difficulty is in finding the index set $\mathcal{I}_{r}$ since $V(\mathcal{I}_{c},:)$ can be obtained by SP algorithm from the IS structure $V=\Pi_{c}V(\mathcal{I}_{c},:)$. From Lemma \ref{ISIC}, we know that $U_{*}=YU_{*}(\mathcal{I}_{r},:)$ forms the IC structure. In \cite{MaoSVM}, their SVM-cone algorithm (i.e., Algorithm 2 in the Supplementary Materials) can exactly obtain the row nodes corner matrix $U_{*}(\mathcal{I}_{r},:)$ from the Ideal Cone $U_{*}=YU_{*}(\mathcal{I}_{r},:)$ as long as the condition  $(U_{*}(\mathcal{I}_{r},:)U'_{*}(\mathcal{I}_{r},:))^{-1}\mathbf{1}>0$ holds (see Lemma \ref{LSVM}).
\begin{lem}\label{LSVM}
	Under $DiDCMM_{n_{r},n_{c}}(K,P,\Pi_{r}, \Pi_{c}, \Theta_{r})$, $(U_{*}(\mathcal{I}_{r},:)U'_{*}(\mathcal{I}_{r},:))^{-1}\mathbf{1}>0$ holds.
\end{lem}
Based on the above analysis, we are now ready to give the following four-stage algorithm which we call Ideal DiMSC. Input $\Omega, K$. Output: $\Pi_{r}$ and $\Pi_{c}$.
\begin{itemize}
  \item Let $\Omega=U\Lambda V'$ be the compact SVD of $\Omega$ such that $U\in\mathbb{R}^{n_{r}\times K},V\in\mathbb{R}^{n_{c}\times K}, \Lambda\in\mathbb{R}^{K\times K},U'U=I,V'V=I$. Let $U_{*}=N_{U}U$, where $N_{U}$ is an $n_{r}\times n_{r}$ diagonal matrix whose $i$-th diagonal entry is $\frac{1}{\|U(i,:)\|_{F}}$ for $1\leq i\leq n_{r}$.
  \item Run SP algorithm on $V$ assuming that there are $K$ column communities to obtain the column corner matrix $V(\mathcal{I}_{c},:)$ (i.e.,$B_{c}$). Run SVM-cone algorithm on $U_{*}$ assuming that there are $K$ row communities to obtain $\mathcal{I}_{r}$.
  \item Set $J_{*}=\mathrm{diag}(U_{*}(\mathcal{I}_{r},:)\Lambda V'(\mathcal{I}_{c},:)), Y_{*}=UU^{-1}_{*}(\mathcal{I}_{r},:), Z_{r}=Y_{*}J_{*}$ and $Z_{c}=VV^{-1}(\mathcal{I}_{c},:)$.
  \item Recover $\Pi_{r}$ and $\Pi_{c}$ by setting $\Pi_{r}(i,:)=\frac{Z_{r}(i,:)}{\|Z_{r}(i,:)\|_{1}}$ for $1\leq i\leq n_{r}$, and $\Pi_{c}(j,:)=\frac{Z_{c}(j,:)}{\|Z_{c}(j,:)\|_{1}}$ for $1\leq j\leq n_{c}$.
\end{itemize}
The following theorem guarantees that DiMSC exactly recover nodes memberships.
\begin{thm}\label{IdealDiMSC}
(Ideal DiMSC). Under $DiDCMM_{n_{r},n_{c}}(K,P,\Pi_{r}, \Pi_{c}, \Theta_{r})$, the Ideal DiMSC exactly recovers the row nodes membership matrix $\Pi_{r}$ and the column nodes membership matrix $\Pi_{c}$.
\end{thm}
To demonstrate that $U_{*}$ has the Ideal Cone structure, we drew panel (a) of Figure \ref{PlotUstarV}. The simulated data used for panel (a) is generated from $DiDCMM_{n_{r},n_{c}}(K,P,\Pi_{r}, \Pi_{c}, \Theta_{r})$ with $n_{r}=600, n_{c}=400, K=3$, each row (and column) community has 120 pure nodes.  Among the 800 nodes, 600 are pure nodes with each cluster has 200 pure nodes. For the 240 mixed row nodes, we set $\Pi_{r}(i,1)=\mathrm{rand}(1)/2, \Pi_{r}(i,2)=\mathrm{rand}(1)/2, \Pi_{r}(i,3)=1-\Pi(j,1)-\Pi(j,2)$ where $\mathrm{rand}(1)$ is any random number in $(0,1)$ and $i$ is a mixed row node. Similar, for the 40 mixed column nodes, set $\Pi_{c}(j,1)=\mathrm{rand}(1)/2, \Pi_{c}(j,2)=\mathrm{rand}(1)/2, \Pi_{c}(j,3)=1-\Pi(j,1)-\Pi(j,2)$. For the degree heterogeneity parameter, set $\theta_{r}(i)=\mathrm{rand}(1)$ for all row nodes $i$. The matrix $P$ is set as
\[P=\begin{bmatrix}
    1&0.4&0.3\\
    0.2&1&0.1\\
    0.1&0.4&1\\
\end{bmatrix}.
\]
Under such setting, after computing $\Omega$ and obtaining $U_{*}, V$ from $\Omega$, we can plot Figure \ref{PlotUstarV}. Panel (a) shows that all rows respective to mixed row nodes of $U_{*}$ are located at one side of the hyperplane formed by the $K$ rows of $U_{*}(\mathcal{I}_{r},:)$, and this phenomenon occurs since each row of $U_{*}$ is a scaled convex combination of the $K$ rows of $U_{*}(\mathcal{I}_{r},:)$ guaranteed by the IC structure $U_{*}=YU_{*}(\mathcal{I}_{r},;)$. Thus panel (a) shows the existence of the Ideal Cone structure formed by $U_{*}$. Similarly,  to demonstrate that $V$ has the Ideal Simplex structure, we drew panel (b) of Figure \ref{PlotUstarV}, where panel (b) is obtained under the same setting as panel (a). Panel (b) shows that rows respective to mixed column nodes of $V$ are located  inside of the simplex formed by the $K$ rows of $V(\mathcal{I}_{c},:)$, and this phenomenon occurs since each row of $V$ is a convex linear combination of the $K$ rows of $V(\mathcal{I}_{c},:)$ guaranteed by the IS structure $V=\Pi_{c}V(\mathcal{I}_{c},;)$. Thus panel (b) shows the existence of the Ideal Simplex structure formed by $V$.
\subsection{DiMSC algorithm}
We now extend the ideal case to the real case. Set $\tilde{A}=\hat{U}\hat{\Lambda}\hat{V}'$ be the top-$K$-dimensional SVD of $A$ such that $\hat{U}\in \mathbb{R}^{n_{r}\times K}, \hat{V}\in \mathbb{R}^{n_{c}\times K}, \hat{\Lambda}\in \mathbb{R}^{K\times K},\hat{U}'\hat{U}=I_{K}, \hat{V}'\hat{V}=I_{K}$, and $\hat{\Lambda}$ contains the top $K$ singular values of $A$. Let $\hat{U}_{*}$ be the row-wise normalization of $\hat{U}$ such that $\hat{U}=N_{\hat{U}}\hat{U}$ where $N_{\hat{U}}\in\mathbb{R}^{n_{r}\times n_{r}}$ is a diagonal matrix whose $i$-th diagonal entry is $\frac{1}{\|\hat{U}(i,:)\|_{F}}$. For the real case, we use $\hat{J}_{*},\hat{Y}_{*}, \hat{Z}_{r}, \hat{Z}_{c}, \hat{\Pi}_{r}, \hat{\Pi}_{c}$ given in Algorithm \ref{alg:DiMSC} to estimate $J_{*},Y_{*},Z_{r},Z_{c}, \Pi_{r},\Pi_{c}$, respectively.  Algorithm \ref{alg:DiMSC} called Directed Mixed Simplex\&Cone (DiMSC for short) algorithm is a natural extension of the Ideal DiMSC to the real case.
\begin{algorithm}
\caption{\textbf{Directed Mixed Simplex \& Cone} (\textbf{DiMSC}) algorithm}
\label{alg:DiMSC}
\begin{algorithmic}[1]
\Require The adjacency matrix $A\in \mathbb{R}^{n_{r}\times n_{c}}$ of a directed network, the number of row communities (column communities) $K$.
\Ensure The estimated $n_{r}\times K$ row membership matrix $\hat{\Pi}_{r}$ and the estimated $n_{c}\times K$ column membership matrix $\hat{\Pi}_{c}$.
\State Obtain $\tilde{A}=\hat{U}\hat{\Lambda}\hat{V}'$, the $K$-dimensional SVD of $A$. Compute $\hat{U}_{*}$ from $\hat{U}$.
\State Apply SP algorithm (i.e., Algorithm 1 in the Supplementary Materials) on the rows of $\hat{V}$ assuming there are $K$ column communities to obtain $\mathcal{\hat{I}}_{c}$, the index set returned by SP algorithm. Similarly, apply SVM-cone algorithm (i.e., Algorithm 2 in the Supplementary Materials) on the rows of $\hat{U}_{*}$ with $K$ row communities to obtain $\mathcal{\hat{I}}_{r}$, the index set returned by SVM-cone algorithm.
\State Set $\hat{J}_{*}=\mathrm{diag}(\hat{U}_{*}(\hat{I}_{r},:)\hat{\Lambda}\hat{V}'(\hat{\mathcal{I}}_{c},:)), \hat{Y}_{*}=\hat{U}\hat{U}^{-1}_{*}(\hat{\mathcal{I}}_{r},:), \hat{Z}_{r}=\hat{Y}_{*}\hat{J}_{*}$ and $\hat{Z}_{c}=\hat{V}\hat{V}^{-1}(\hat{\mathcal{I}}_{c},:)$. Then set $\hat{Z}_{r}=\mathrm{max}(0, \hat{Z}_{r})$ and $\hat{Z}_{c}=\mathrm{max}(0, \hat{Z}_{c})$.
\State Estimate $\Pi_{r}(i,:)$ by $\hat{\Pi}_{r}(i,:)=\hat{Z}_{r}(i,:)/\|\hat{Z}_{r}(i,:)\|_{1}, 1\leq i\leq n_{r}$ and estimate $\Pi_{c}(j,:)$ by $\hat{\Pi}_{c}(j,:)=\hat{Z}_{c}(j,:)/\|\hat{Z}_{c}(j,:)\|_{1}, 1\leq j\leq n_{c}$.
\end{algorithmic}
\end{algorithm}

In the 3rd step, we set the negative entries of $\hat{Z}_{r}$ as 0 by setting $\hat{Z}_{r}=\mathrm{max}(0, \hat{Z}_{r})$ for the reason that weights for any row node should be nonnegative while there may exist some negative entries of $\hat{Y}_{*}\hat{J}_{*}$. Similar argument holds for $\hat{Z}_{c}$.
\subsection{Equivalence algorithm}
In this subsection, we design one algorithm DiMSC-equivalence which returns same estimations as Algorithm \ref{alg:DiMSC}.
Set $U_{2}=UU'\in\mathbb{R}^{n_{r}\times n_{r}}, \hat{U}_{2}=\hat{U}\hat{U}'\in\mathbb{R}^{n_{r}\times n_{r}}, V_{2}=VV'\in\mathbb{R}^{n_{c}\times n_{c}}, \hat{V}_{2}=\hat{V}\hat{V}'\in\mathbb{R}^{n_{c}\times n_{c}}$. Set $U_{*,2}\in\mathbb{R}^{n_{r}\times n_{r}}$ as  $U_{*,2}(i,:)=\frac{U_{2}(i,:)}{\|U_{2}(i,:)\|_{F}}$ for $1\leq i\leq n_{r}$. $\hat{U}_{*,2}$ is defined similarly. Next lemma guarantees that $V_{2}$ enjoys IS structure and $U_{*,2}$ enjoys IC structure.
\begin{lem}\label{ISIC2}
Under $DiDCMM_{n_{r},n_{c}}(K,P,\Pi_{r}, \Pi_{c}, \Theta_{r})$, we have $V_{2}=\Pi_{c}V_{2}(\mathcal{I}_{c},:)$, and $U_{*,2}=YU_{*,2}(\mathcal{I}_{r},:)$.
\end{lem}
Since $U_{*,2}(\mathcal{I}_{r},:)\in \mathbb{R}^{K\times n_{r}}$ and $V_{2}(\mathcal{I}_{c},:)\in \mathbb{R}^{K\times n_{c}}$, $U_{*,2}(\mathcal{I}_{r},:)$ and $V_{2}(\mathcal{I}_{c},:)$ are singular matrix with rank $K$ by condition (I1) while the inverses of $U_{*,2}(\mathcal{I}_{r},:)U'_{*,2}(\mathcal{I}_{r},:)$ and $V_{2}(\mathcal{I}_{c},:)V'_{2}(\mathcal{I}_{c},:)$ exist. Therefore, Lemma \ref{ISIC2} gives that
\begin{align*}
Y=U_{*,2}U'_{*,2}(\mathcal{I}_{r},:)(U_{*,2}(\mathcal{I}_{r},:)U'_{*,2}(\mathcal{I}_{r},:))^{-1}, \Pi_{c}=V_{2}V'_{2}(\mathcal{I}_{c},:)(V_{2}(\mathcal{I}_{c},:)V'_{2}(\mathcal{I}_{c},:))^{-1}.
\end{align*}
Since $U_{*,2}=N_{U}U_{2}$ and $Y=N_{M}\Pi_{r}\Theta^{-1}_{r}(,\mathcal{I}_{r},\mathcal{I}_{r})N^{-1}_{U}(\mathcal{I}_{r},\mathcal{I}_{r})$, we see that $Y_{*}$ also equals to $U_{2}U'_{*,2}(\mathcal{I}_{r},:)(U_{*,2}(\mathcal{I}_{r},:)U'_{*,2}(\mathcal{I}_{r},:))^{-1}$ by basic algebra.

Based on the above analysis, we are now ready to give the Ideal DiMSC-equivalence. Input $\Omega$. Output: $\Pi_{r}$ and $\Pi_{c}$.
\begin{itemize}
  \item Obtain $U,\Lambda,V, U_{*,2}, V_{2}$ from $\Omega$.
  \item Run SP algorithm on $V_{2}$ with $K$ column communities to obtain $V_{2}(\mathcal{I}_{c},:)$. Run SVM-cone algorithm on $U_{*,2}$ with $K$ row communities to obtain $\mathcal{I}_{r}$.
  \item Set $J_{*}=\mathrm{diag}(U_{*}(\mathcal{I}_{r},:)\Lambda V'(\mathcal{I}_{c},:)), Y_{*}=U_{2}U'_{*,2}(\mathcal{I}_{r},:)(U_{*,2}(\mathcal{I}_{r},:)U'_{*,2}(\mathcal{I}_{r},:))^{-1}, Z_{r}=Y_{*}J_{*}$ and $Z_{c}=V_{2}V'_{2}(\mathcal{I}_{c},:)(V_{2}(\mathcal{I}_{c},:)V'_{2}(\mathcal{I}_{c},:))^{-1}$.
  \item Recover $\Pi_{r}$ and $\Pi_{c}$ by setting $\Pi_{r}(i,:)=\frac{Z_{r}(i,:)}{\|Z_{r}(i,:)\|_{1}}$ for $1\leq i\leq n_{r}$, and $\Pi_{c}(j,:)=\frac{Z_{c}(j,:)}{\|Z_{c}(j,:)\|_{1}}$ for $1\leq j\leq n_{c}$.
\end{itemize}
For the real case, set $\hat{U}_{2}=\hat{U}\hat{U}', \hat{V}_{2}=\hat{V}\hat{V}', \hat{U}_{*,2}=N_{\hat{U}}\hat{U}_{2}$. We now extend the ideal case to the real one given by Algorithm \ref{alg:DiMSCequivalence}.
\begin{algorithm}
\caption{DiMSC-equivalence}
\label{alg:DiMSCequivalence}
\begin{algorithmic}[1]
\Require The adjacency matrix $A\in \mathbb{R}^{n_{r}\times n_{c}}$ of a directed network, the number of row communities (column communities) $K$.
\Ensure The estimated $n_{r}\times K$ row membership matrix $\hat{\Pi}_{r,2}$ and the estimated $n_{c}\times K$ column membership matrix $\hat{\Pi}_{c,2}$.
\State Obtain $\tilde{A}=\hat{U}\hat{\Lambda}\hat{V}'$, the $K$-dimensional SVD of $A$. Compute $\hat{U}_{*},\hat{U}_{2}, \hat{V}_{2}, \hat{U}_{*,2}$.
\State Apply SP algorithm on the rows of $\hat{V}_{2}$ assuming there are $K$ column communities to obtain $\mathcal{\hat{I}}_{c,2}$, the index set returned by SP algorithm. Apply SVM-cone algorithm on the rows of $\hat{U}_{*,2}$ with $K$ row communities to obtain $\mathcal{\hat{I}}_{r,2}$, the index set returned by SVM-cone algorithm.
\State Set $\hat{J}_{*,2}=\mathrm{diag}(\hat{U}_{*}(\hat{I}_{r,2},:)\hat{\Lambda}\hat{V}'(\hat{\mathcal{I}}_{c,2},:)), \hat{Y}_{*,2}=\hat{U}_{2}\hat{U}'_{*,2}(\hat{\mathcal{I}}_{r,2},:)(\hat{U}_{*,2}(\hat{\mathcal{I}}_{r,2},:)\hat{U}'_{*,2}(\hat{\mathcal{I}}_{r,2},:))^{-1}, \hat{Z}_{r,2}=\hat{Y}_{*,2}\hat{J}_{*,2}$ and $\hat{Z}_{c,2}=\hat{V}_{2}\hat{V}'_{2}(\hat{\mathcal{I}}_{c,2},:)(\hat{V}_{2}(\hat{\mathcal{I}}_{c,2},:)\hat{V}'_{2}(\hat{\mathcal{I}}_{c,2},:))^{-1}$. Then set $\hat{Z}_{r,2}=\mathrm{max}(0, \hat{Z}_{r,2})$ and $\hat{Z}_{c,2}=\mathrm{max}(0, \hat{Z}_{c,2})$.
\State Estimate $\Pi_{r}(i,:)$ by $\hat{\Pi}_{r,2}(i,:)=\hat{Z}_{r,2}(i,:)/\|\hat{Z}_{r,2}(i,:)\|_{1}, 1\leq i\leq n_{r}$ and estimate $\Pi_{c}(j,:)$ by $\hat{\Pi}_{c,2}(j,:)=\hat{Z}_{c,2}(j,:)/\|\hat{Z}_{c,2}(j,:)\|_{1}, 1\leq j\leq n_{c}$.
\end{algorithmic}
\end{algorithm}
\begin{lem}\label{Equivalence}
(Equivalence). For the empirical case, we have
$\mathcal{\hat{I}}_{r,2}\equiv \mathcal{\hat{I}}_{r}, \mathcal{\hat{I}}_{c,2}\equiv \mathcal{\hat{I}}_{c}, \hat{U}_{*,2}(\hat{\mathcal{I}}_{r,2},:)\hat{U}'_{*,2}(\hat{\mathcal{I}}_{r,2},:)\equiv \hat{U}_{*}(\hat{\mathcal{I}}_{r},:)\hat{U}'_{*}(\hat{\mathcal{I}}_{r},:), \hat{Y}_{*,2}\equiv\hat{Y}_{*},\hat{J}_{*,2}\equiv\hat{J}_{*}, \hat{Z}_{r,2}\equiv\hat{Z}_{r}, \hat{Z}_{c,2}\equiv\hat{Z}_{c}, \hat{\Pi}_{r,2}\equiv\hat{\Pi}_{r}$ and $\hat{\Pi}_{c,2}\equiv\hat{\Pi}_{c}$.
\end{lem}
Lemma \ref{Equivalence} guarantees that the two Algorithms \ref{alg:DiMSC} and \ref{alg:DiMSCequivalence} return same estimations for both row and column nodes's memberships. In this article, we introduce DiMSC-equivalence algorithm since it is helpful to build theoretical framework for DiMSC, see Remarks 9 and 10 in the Supplementary Materials for detail.
\section{Consistency results}\label{sec4}
In this section, we show the consistency of our algorithm for fitting the DiDCMM by proving that the sample-based estimates $\hat{\Pi}_{r}$ and $\hat{\Pi}_{c}$ concentrate around the true mixed membership matrices $\Pi_{r}$ and $\Pi_{c}$. Throughout this paper, $K$ is a known positive integer. Set $\theta_{r,\mathrm{max}}=\mathrm{max}_{1\leq i\leq n_{r}}\theta_{r}(i)$ and $\theta_{r,\mathrm{min}}=\mathrm{min}_{1\leq i\leq n_{r}}\theta_{r}(i)$. Assume that
\begin{assum}\label{a1}
$P_{\mathrm{max}}\mathrm{max}(\|\theta_{r}\|_{1},\theta_{r,\mathrm{max}}n_{c})\geq \mathrm{log}(n_{r}+n_{c})$.
\end{assum}
Assumption \ref{a1} means that the network can not be too sparse, and Assumption \ref{a1} also means that we allow $\theta_{r,\mathrm{max}}$ goes to zero with increasing numbers of row nodes and column nodes. Since we let $P_{\mathrm{max}}\leq C$, Assumption \ref{a1} equals $\mathrm{max}(\|\theta_{r}\|_{1},\theta_{r,\mathrm{max}}n_{c})\geq \mathrm{log}(n_{r}+n_{c})/C$.  Then we have the following lemma.
\begin{lem}\label{BoundAOmega}
Under $DiDCMM_{n_{r},n_{c}}(K,P,\Pi_{r}, \Pi_{c}, \Theta_{r})$, when Assumption \ref{a1} holds, with probability at least $1-o((n_{r}+n_{c})^{-3})$, we have
\begin{align*}
\|A-\Omega\|=O(\sqrt{\mathrm{max}(\|\theta_{r}\|_{1},\theta_{r,\mathrm{max}}n_{c})\mathrm{log}(n_{r}+n_{c})}).
\end{align*}
\end{lem}
In \cite{MixedSCORE, MaoSVM, mao2020estimating}, main theoretical results for  their proposed community detection methods hinge on a row-wise deviation bound for the eigenvectors of the adjacency matrix whether under MMSB or DCMM. Similarly, for our DiMSC, the main theoretical results (i.e., Theorem \ref{Main}) also rely on the row-wise deviation bounds for the singular eigenvectors of the adjacency matrix.  Different from the theoretical techniques in Theorem 3.1 in \cite{mao2020estimating} and Lemma C.3 in \cite{MixedSCORE}, to obtain the row-wise deviation bound for the singular eigenvector of $\Omega$, we use Theorem 4.3.1 in \cite{chen2020spectral}.
\begin{lem}\label{rowwiseerror}
	(Row-wise singular eigenvector error) Under $DiDCMM_{n_{r},n_{c}}(K,P,\Pi_{r}, \Pi_{c}, \Theta_{r})$, when Assumption \ref{a1} holds, suppose $\sigma_{K}(\Omega)\geq C\sqrt{\theta_{r,\mathrm{max}}(n_{r}+n_{c})\mathrm{log}(n_{r}+n_{c})}$, with probability at least $1-o((n_{r}+n_{c})^{-3})$, we have
\begin{align*}
&\mathrm{max}(\|\hat{U}\hat{U}'-UU'\|_{2\rightarrow\infty}, \|\hat{V}\hat{V}'-VV'\|_{2\rightarrow\infty})=O(\frac{\sqrt{\theta_{r,\mathrm{max}}K\mathrm{log}(n_{r}+n_{c})}}{\theta_{r,\mathrm{min}}\sigma_{K}(P)\sigma_{K}(\Pi_{r})\sigma_{K}(\Pi_{c})}).
\end{align*}
\end{lem}
When $\Theta_{r}=\rho I, n_{r}=n_{c}, \Pi_{r}=\Pi_{c}=\Pi$, and DiCCMM degenerates to MMSB, the bound in Lemma \ref{rowwiseerror} is $O(\frac{\sqrt{K\mathrm{log}(n)}}{\sigma_{K}(P)\sqrt{\rho}\lambda_{K}(\Pi'\Pi)})$. if we further assume that $\lambda_{K}(\Pi'\Pi)=O(\frac{n}{K})$ and $K=O(1)$, the bound is of order $O(\frac{1}{\sigma_{K}(P)}\frac{1}{\sqrt{n}}\sqrt{\frac{\mathrm{log}(n)}{\rho n}})$. Set the $\Theta$ in \cite{MixedSCORE} as $\sqrt{\rho}I$, their DCMM degenerates to MMSB, their assumptions are translated to our $\lambda_{K}(\Pi'\Pi)=O(\frac{n}{K})$, when $K=O(1)$, the row-wise deviation bound in the fourth bullet of Lemma 2.1 \cite{MixedSCORE} is $O(\frac{1}{\sigma_{K}(P)}\frac{1}{\sqrt{n}}\sqrt{\frac{\mathrm{log}(n)}{\rho n}})$, which is consistent with ours. Meanwhile, if we further assume that $\sigma_{K}(P)=O(1)$, the bound is of order $\frac{1}{\sqrt{n}}\sqrt{\frac{\mathrm{log}(n)}{\rho n}}$, which is consistent with the row-wise eigenvector deviation of \cite{lei2019unified}'s result shown in their Table 2.

For convenience, set $\varpi=\mathrm{max}(\|\hat{U}\hat{U}'-UU'\|_{2\rightarrow\infty}, \|\hat{V}\hat{V}'-VV'\|_{2\rightarrow\infty})$, and  $\pi_{r,\mathrm{min}}=\mathrm{min}_{1\leq k\leq K}\mathbf{1}'\Pi_{r} e_{k}$ , where $\pi_{r,\mathrm{min}}$ measures the minimum summation of row nodes belong to a certain row community. Increasing $\pi_{r,\mathrm{min}}$ makes the network tend to be more balanced, vice verse. Next lemma is the corner stone to characterize the behaviors of DiMSC.
\begin{lem}\label{boundC}
	Under $DiDCMM_{n_{r},n_{c}}(K,P,\Pi_{r}, \Pi_{c}, \Theta_{r})$, when conditions of Lemma \ref{rowwiseerror} hold, there exist two permutation matrices $\mathcal{P}_{r},\mathcal{P}_{c}\in\mathbb{R}^{K\times K}$ such that with probability at least $1-o((n_{r}+n_{c})^{-3})$, we have
\begin{align*}
&\mathrm{max}_{1\leq k\leq K}\|e'_{k}(\hat{U}_{*,2}(\mathcal{\hat{I}}_{r},:)-\mathcal{P}'_{r}U_{*,2}(\mathcal{I}_{r},:))\|_{F}=O(\frac{K^{3}\theta^{11}_{r,\mathrm{max}}\varpi\kappa^{3}(\Pi'_{r}\Pi_{r})\lambda^{1.5}_{1}(\Pi'_{r}\Pi_{r})}{\theta^{11}_{r,\mathrm{min}}\pi_{r,\mathrm{min}}}),\\
&\mathrm{max}_{1\leq k\leq K}\|e'_{k}(\hat{V}_{2}(\mathcal{\hat{I}}_{c},:)-\mathcal{P}'_{c}V_{2}(\mathcal{I}_{c},:))\|_{F}=O(\varpi\kappa(\Pi'_{c}\Pi_{c})).
\end{align*}
\end{lem}
\begin{lem}\label{boundZ}
Under $DiDCMM_{n_{r},n_{c}}(K,P,\Pi_{r}, \Pi_{c}, \Theta_{r})$, when conditions of Lemma \ref{rowwiseerror} hold,, with probability at least $1-o((n_{r}+n_{c})^{-3})$, we have
\begin{align*}
&\mathrm{max}_{1\leq i\leq n_{r}}\|e'_{i}(\hat{Z}_{r}-Z_{r}\mathcal{P}_{r})\|_{F}=O(\frac{K^{5}\theta^{15}_{r,\mathrm{max}}\varpi\kappa^{4.5}(\Pi'_{r}\Pi_{r})\kappa(\Pi_{c})\lambda^{1.5}_{1}(\Pi'_{r}\Pi_{r})}{\theta^{14}_{r,\mathrm{min}}\pi_{r,\mathrm{min}}}),\\
&\mathrm{max}_{1\leq j\leq n_{c}}\|e'_{j}(\hat{Z}_{c}-Z_{c}\mathcal{P}_{c})\|_{F}=O(\varpi\kappa(\Pi'_{c}\Pi_{c})\sqrt{K\lambda_{1}(\Pi'_{c}\Pi_{c})}).
\end{align*}
\end{lem}
Next theorem gives theoretical bounds on estimations of memberships for both row and column nodes, which is the main theoretical result for our DiMSC method.
\begin{thm}\label{Main}
Under $DiDCMM_{n_{r},n_{c}}(K,P,\Pi_{r}, \Pi_{c}, \Theta_{r})$, suppose conditions in Lemma \ref{rowwiseerror} hold, with probability at least $1-o((n_{r}+n_{c})^{-3})$, we have
\begin{align*}
&\mathrm{max}_{1\leq i\leq n_{r}}\|e'_{i}(\hat{\Pi}_{r}-\Pi_{r}\mathcal{P}_{r})\|_{1}=O(\frac{K^{5.5}\theta^{15}_{r,\mathrm{max}}\varpi\kappa^{4.5}(\Pi'_{r}\Pi_{r})\kappa(\Pi_{c})\lambda^{1.5}_{1}(\Pi'_{r}\Pi_{r})}{\theta^{15}_{r,\mathrm{min}}\pi_{r,\mathrm{min}}}),\\
&\mathrm{max}_{1\leq j\leq n_{c}}\|e'_{j}(\hat{\Pi}_{c}-\Pi_{c}\mathcal{P}_{c})\|_{1}=O(\varpi K\kappa(\Pi'_{c}\Pi_{c})\sqrt{\lambda_{1}(\Pi'_{c}\Pi_{c})}).
\end{align*}
\end{thm}
The following corollary is obtained by adding conditions on model parameters similar as Corollary 3.1 in \cite{mao2020estimating}.
\begin{cor}\label{AddConditions}
Under $DiDCMM_{n_{r},n_{c}}(K,P,\Pi_{r}, \Pi_{c}, \Theta_{r})$, when conditions of Lemma \ref{rowwiseerror} hold, suppose $\lambda_{K}(\Pi'_{r}\Pi_{r})=O(\frac{n_{r}}{K}), \lambda_{K}(\Pi'_{c}\Pi_{c})=O(\frac{n_{c}}{K}), \pi_{r,\mathrm{min}}=O(\frac{n_{r}}{K})$ and $K=O(1)$, with probability at least $1-o((n_{r}+n_{c})^{-3})$, we have
\begin{align*}
&\mathrm{max}_{1\leq i\leq n_{r}}\|e'_{i}(\hat{\Pi}_{r}-\Pi_{r}\mathcal{P}_{r})\|_{1}=O((\frac{\theta_{r,\mathrm{max}}}{\theta_{r,\mathrm{min}}})^{15.5}\frac{1}{\sigma_{K}(P)}\sqrt{\frac{\mathrm{log}(n_{r}+n_{c})}{\theta_{r,\mathrm{min}}n_{c}}}),\\
&\mathrm{max}_{1\leq j\leq n_{c}}\|e'_{j}(\hat{\Pi}_{c}-\Pi_{c}\mathcal{P}_{c})\|_{1}=O((\frac{\theta_{r,\mathrm{max}}}{\theta_{r,\mathrm{min}}})^{0.5}\frac{1}{\sigma_{K}(P)}\sqrt{\frac{\mathrm{log}(n_{r}+n_{c})}{\theta_{r,\mathrm{min}}n_{r}}}).
\end{align*}
Meanwhile,
\begin{itemize}
\item when $n_{r}=O(n), n_{c}=O(n)$ (i.e.,
    $\frac{n_{r}}{n_{c}}=O(1)$), we have
\begin{align*}
&\mathrm{max}_{1\leq i\leq n_{r}}\|e'_{i}(\hat{\Pi}_{r}-\Pi_{r}\mathcal{P}_{r})\|_{1}=O((\frac{\theta_{r,\mathrm{max}}}{\theta_{r,\mathrm{min}}})^{15.5}\frac{1}{\sigma_{K}(P)}\sqrt{\frac{\mathrm{log}(n)}{\theta_{r,\mathrm{min}}n}}),\\
&\mathrm{max}_{1\leq j\leq n_{c}}\|e'_{j}(\hat{\Pi}_{c}-\Pi_{c}\mathcal{P}_{c})\|_{1}=O((\frac{\theta_{r,\mathrm{max}}}{\theta_{r,\mathrm{min}}})^{0.5}\frac{1}{\sigma_{K}(P)}\sqrt{\frac{\mathrm{log}(n)}{\theta_{r,\mathrm{min}}n}}).
\end{align*}
\item when $\theta_{r,\mathrm{max}}=O(\rho), \theta_{r,\mathrm{min}}=O(\rho)$ (i.e.,
    $\frac{\theta_{r,\mathrm{min}}}{\theta_{r,\mathrm{max}}}=O(1)$), we have
\begin{align*}
&\mathrm{max}_{1\leq i\leq n_{r}}\|e'_{i}(\hat{\Pi}_{r}-\Pi_{r}\mathcal{P}_{r})\|_{1}=O(\frac{1}{\sigma_{K}(P)}\sqrt{\frac{\mathrm{log}(n_{r}+n_{c})}{\rho n_{c}}}),\\
&\mathrm{max}_{1\leq j\leq n_{c}}\|e'_{j}(\hat{\Pi}_{c}-\Pi_{c}\mathcal{P}_{c})\|_{1}=O(\frac{1}{\sigma_{K}(P)}\sqrt{\frac{\mathrm{log}(n_{r}+n_{c})}{\rho n_{r}}}).
\end{align*}
\item when $n_{r}=O(n), n_{c}=O(n)$ and $\theta_{r,\mathrm{max}}=O(\rho), \theta_{r,\mathrm{min}}=O(\rho)$, we have
\begin{align*}
&\mathrm{max}_{1\leq i\leq n_{r}}\|e'_{i}(\hat{\Pi}_{r}-\Pi_{r}\mathcal{P}_{r})\|_{1}=O(\frac{1}{\sigma_{K}(P)}\sqrt{\frac{\mathrm{log}(n)}{\rho n}}),\\
&\mathrm{max}_{1\leq j\leq n_{c}}\|e'_{j}(\hat{\Pi}_{c}-\Pi_{c}\mathcal{P}_{c})\|_{1}=O(\frac{1}{\sigma_{K}(P)}\sqrt{\frac{\mathrm{log}(n)}{\rho n}}).
\end{align*}
\end{itemize}
\end{cor}
Consider a bipartite mixed membership network  under the settings of Corollary \ref{AddConditions} when $\theta_{r,\mathrm{max}}=O(\rho), \theta_{r,\mathrm{min}}=O(\rho)$ where we call $\rho$ as sparsity parameter in this paper, to obtain consistency estimations for both row nodes and column nodes, by Corollary \ref{AddConditions}, $\sigma_{K}(P)$ should grow faster than $\sqrt{\frac{\mathrm{log}(n_{r}+n_{c})}{\rho \mathrm{min}(n_{r},n_{c})}}$. Especially, when $n_{r}=O(n)$ and $n_{c}=O(n)$, $\sigma_{K}(P)$ should grow faster than $\sqrt{\frac{\mathrm{log}(n)}{n}}$. We further assume that $P=(2-\beta)I_{K}+(\beta-1)\textbf{1}\textbf{1}'$ for $\beta\in [1,2)\cup(2,\infty)$, we see that this $P$ has unit diagonals and $\beta-1$ as non-diagonal entries. Meanwhile, $\sigma_{K}(P)=|\beta-2|\equiv P_{\mathrm{max}}-P_{\mathrm{min}}$ (here, $P_{\mathrm{min}}=\mathrm{min}_{1\leq k,l\leq K}P(k,l)$) and $P$ tends to be a singular matrix as $\beta$ is close to 2. To obtain consistency estimation, $P_{\mathrm{max}}-P_{\mathrm{min}}$ should grow faster than $\sqrt{\frac{\mathrm{log}(n_{r}+n_{c})}{\rho \mathrm{min}(n_{r},n_{c})}}$ by Corollary \ref{AddConditions}, and $P_{\mathrm{max}}-P_{\mathrm{min}}$ should grow faster than $\sqrt{\frac{\mathrm{log}(n)}{\rho n}}$ when $n_{r}=O(n)$ and $n_{c}=O(n)$.
\begin{rem}
	When the network is undirected (i.e., $n_{r}=n_{c}=n, \Pi_{r}=\Pi_{c}$) with $K=O(1)$ by setting $\theta_{r}(i)=\rho$ for $1\leq i\leq n_{r}$, DiDCMM degenerates to MMSB considered in \cite{mao2020estimating}, the upper bound of error rate for DiMSC is $O(\frac{1}{\sigma_{K}(P)}\sqrt{\frac{\mathrm{log}(n)}{\rho n}})$. Replacing the $\Theta$ in \cite{MixedSCORE} by $\Theta=\sqrt{\rho}I$, their DCMM model  degenerates to MMSB. Then their conditions in Theorem 2.2 are our Assumption \ref{a1} and $\lambda_{K}(\Pi'\Pi)=O(\frac{n}{K})$ where $\Pi=\Pi_{r}=\Pi_{c}$ for MMSB. When $K=O(1)$, the error bound in Theorem 2.2 in \cite{MixedSCORE} is
$O(\frac{1}{\sigma_{K}(P)}\sqrt{\frac{\mathrm{log}(n)}{\rho n}})$, which is consistent with ours.
\end{rem}
\section{Simulations}\label{sec5}
In this section, four experiments are conducted to investigate the performance of our DiMSC. We measure the performance of DiMSC by row-Mixed-Hamming error rate (rMHamm for short) and column-Mixed-Hamming error rate (cMHamm for short) defined below
\begin{align*}
\mathrm{rowMHamm}=\frac{\mathrm{min}_{\mathcal{P}\in S_{\mathcal{P}}}\|\hat{\Pi}_{r}\mathcal{P}-\Pi_{r}\|_{1}}{n_{r}}\mathrm{~and~} \mathrm{columnMHamm}=\frac{\mathrm{min}_{\mathcal{P}\in S}\|\hat{\Pi}_{c}\mathcal{P}-\Pi_{c}\|_{1}}{n_{c}},
\end{align*}
where $S_{\mathcal{P}}$ is the set of $K\times K$ permutation matrices.

For all simulations in this section, the parameters $(n_{r}, n_{c}, K, P, \Pi_{r}, \Pi_{c},\Theta_{r})$ under DiDCMM are set as follows. Set $n_{r}=500, n_{c}=600, K=3$. Let each row community and each column community have $n_{0}$ pure nodes. Let all the mixed row nodes have four different memberships $(0.4, 0.4, 0.2), (0.4, 0.2, 0.4), (0.2, 0.4, 0.4)$ and $(1/3,1/3,1/3)$, each with $\frac{500-3n_{0}}{4}$ number of row nodes. Let all the mixed column nodes also have the above four different memberships, each with $\frac{600-3n_{0}}{4}$ number of column nodes. Set $n_{0}$ as 80 except Experiment 1 where we study the fraction of pure nodes.  For $z\geq 1$, we generate the degree parameters for row nodes as below: let $\bar{\theta}_{r}\in\mathbb{R}^{n_{r}\times 1}$ such that $1/\bar{\theta}_{r}(i)\overset{iid}{\sim}U(1,z)$ for $1\leq i\leq n_{r}$, where $U(1,z)$ denotes the uniform distribution on $[1, z]$, and set $\theta_{r}=\rho \bar{\theta}_{r}$, where we use $\rho$ to control the sparsity of the network. $\rho$ is set as 1 except Experiment 4 where we study the sparsity.  Set $z$ as 5 except Experiment 2 where we study the degree heterogeneity. Except for Experiment 3 where we study the connectivity across communities, $P$ is set as
\[P=\begin{bmatrix}
    1&0.1&0.3\\
    0.2&1&0.4\\
    0.5&0.2&1\\
\end{bmatrix}.\]
After obtaining $P, \Pi_{r}, \Pi_{c},\theta_{r}$, similar as the five simulation steps in \cite{SCORE},  each simulation experiment contains the following steps:

(a) Let $\Theta_{r}$ be the $n_{r}\times n_{r}$ diagonal matrix such that $\Theta_{r}(i,i)=\theta_{r}(i),1\leq i\leq n_{r}$. Set $\Omega=\Theta_{r}\Pi_{r}P\Pi'_{c}$.

(b)  Let $W$ be an $n_{r}\times n_{c}$ matrix such that $W(i,j)$ are independent centered-Bernoulli with parameters $\Omega(i,j)$. Let $\tilde{A}=\Omega+W$.

(c)  Set $\tilde{S}_{r}=\{i: \sum_{j=1}^{n_{c}}\tilde{A}(i,j)=0\}$ and $\tilde{S}_{c}=\{j: \sum_{i=1}^{n_{r}}\tilde{A}(i,j)=0\}$, i.e., $\tilde{S}_{r}$ ($\tilde{S}_{c}$) is the set of row (column) nodes with 0 edges. Let $A$ be the adjacency matrix obtained by removing rows respective to nodes in $\tilde{S}_{r}$ and removing columns respective to nodes in $\tilde{S}_{c}$ from $\tilde{A}$. Similarly, update $\Pi_{r}$ by removing nodes in $\tilde{S}_{r}$ and update $\Pi_{c}$ by removing nodes in $\tilde{S}_{c}$.

(d) Apply DiMSC algorithm to $A$. Record rowMHamm  and columnMHamm under investigations.

(e) Repeat (b)-(d) for 50 times, and report the averaged rowMHamm and the averaged columnMHamm over the 50 repetitions.

Let $n_{r,A}$ be the number of rows of $A$ and $n_{c,A}$ be the number of columns of $A$. In our experiments, $n_{r,A}$ and $n_{c,A}$ are usually very close to $n_{r}$ and $n_{c}$, therefore we do not report the exact values of $n_{r,A}$ and $n_{c,A}$. After providing the above steps about how to generate $A$ numerically under DiDCMM and how to record the error rates, now we describe the four experiments in detail.

\begin{figure}
\centering
\subfigure[Errors against increasing $n_{0}$.]{\includegraphics[width=0.45\textwidth]{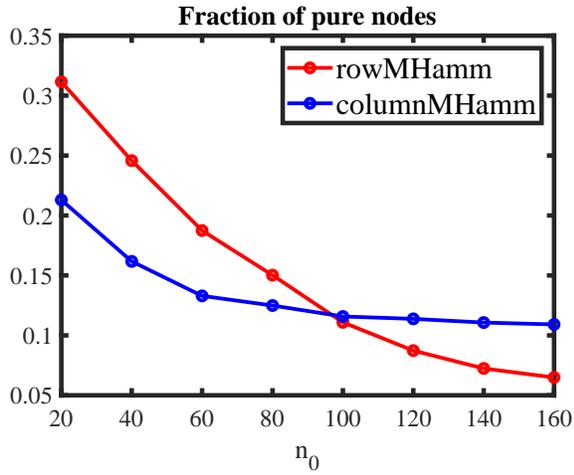}}
\subfigure[Errors against increasing $z$.]{\includegraphics[width=0.45\textwidth]{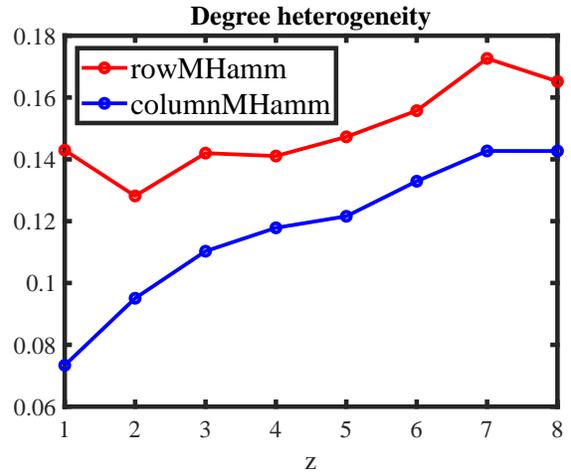}}
\subfigure[Errors against increasing $\beta$.]{\includegraphics[width=0.45\textwidth]{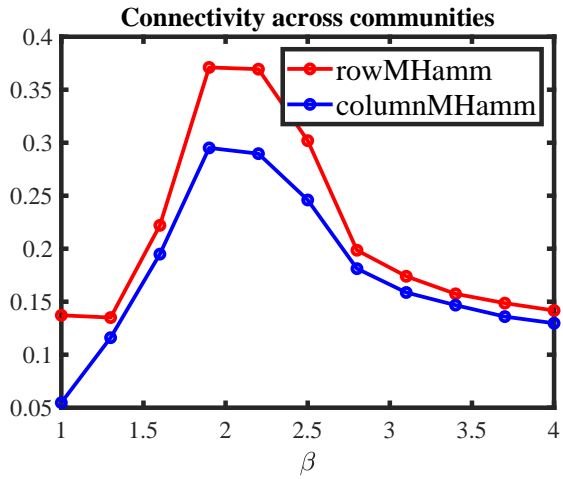}}
\subfigure[Errors against increasing $\rho$.]{\includegraphics[width=0.45\textwidth]{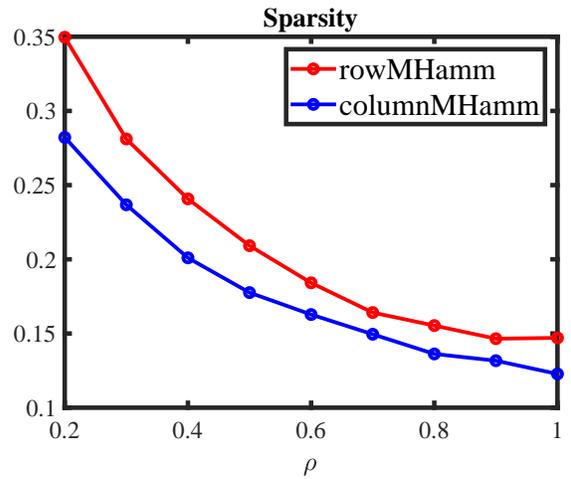}}
\caption{Estimation errors of DiMSC.}
\label{EX}
\end{figure}

\texttt{Experiment 1: Fraction of pure nodes.} Let $n_{0}$ range in $\{20, 40,\ldots,160\}$. Increasing $n_{0}$ increases the fraction of pure nodes for both row and column communities. The numerical results are shown in panel (a) of Figure \ref{EX}. The results show that as the fraction of pure nodes increases for both row and column communities, DiMSC performs better.

\texttt{Experiment 2: Degree heterogeneity.} Let $z$ range in $\{1,2,\ldots,8\}$. A lager $z$ generates lesser edges. The results are displayed in panel (b) of Figure \ref{EX}. The results suggest that the error rates of DiMSC for both row and column nodes tend to increase as $z$ increases. This phenomenon happens because decreasing degree heterogeneities for row nodes lowers the number of edges in the directed network, thus the network become harder to be detected for both row and column nodes.

\texttt{Experiment 3: Connectivity across communities.} Set \[P=\begin{bmatrix}
    1&\beta-1&\beta-1\\
    \beta-1&1&\beta-1\\
    \beta-1&\beta-1&1\\
\end{bmatrix}.
\]
and let $\beta$ range in $\{1,1.3,1.6,\ldots,4\}$. Decreasing $|\beta-2|$ increases the hardness for detecting such directed networks. Note that  $\mathbb{P}(A(i,j)=1)=\Omega(i,j)=\theta_{r}(i)\Pi_{r}(i,:)P\Pi'_{c}(j,:)$ gives $\mathrm{max}_{i,j}\Omega(i,j)=\theta_{r,\mathrm{max}}P_{\mathrm{max}}$ should be no larger than 1. Since $P_{\mathrm{max}}$ may be larger than 1 in this experiment, after obtaining $\theta_{r}$, we need to update $\theta_{r}$ as $\theta_{r}/P_{\mathrm{max}}$. The results are displayed in panel (c) of Figure \ref{EX} and they support the arguments given after Corollary \ref{AddConditions} such that DiMSC performs better when $|\beta-2|$ increases, and vice verse.

\texttt{Experiment 4: Sparsity.} Let $\rho$ range in $\{0.2, 0.3,\ldots, 1\}$. A larger $\rho$ indicates a denser network. Panel (d) in Figure \ref{EX} displays simulation results of this experiment. We see that DiMSC performs better as the simulated network becomes denser.
\section{Discussion and conclusion}\label{sec6}
In this paper, we propose directed degree corrected mixed membership (DiDCMM) model. DiDCMM models a directed network with mixed memberships for row nodes with degree heterogeneities and column nodes without degree heterogeneities. DiDCMM is identifiable when the two well-used conditions (I1) and (I2) hold. To fit the model, we propose a provably consistent spectral algorithm called DiMSC to infer community memberships for both row and column nodes in a directed network generated by a DiDCMM. DiMSC algorithm is designed based on the SVD of the adjacency matrix, where we apply SP algorithm to hunt for the corners in the simplex structure and SVM-cone algorithm to hunt for the corners in the cone structure. To show the consistency of estimations for DiMSC, we introduce the DiMSC-equivalence algorithm where we benefit a lot when building theoretical frameworks for DiMSC from this equivalence algorithm. Meanwhile, we take the advantage of the recent technique developed in \cite{chen2020spectral} to obtain the row-wise singular vector errors when building our theoretical results. The theoretical results of DiMSC show that it consistently recovers memberships of both row nodes and column nodes under mild conditions. Meanwhile, when DiDCMM degenerates to MMSB, our theoretical results match that of Theorem 2.2 \cite{MixedSCORE} when their DCMM degenerates to MMSB under mild conditions.

When designing DiMSC, to construct the Ideal Cone, we obtain the row-wise $l_{2}$-norms of $U$. Actually, similar as \cite{MixedSCORE}, we may obtain an Ideal Simplex from $U$ using the entry-wise ratios idea proposed in \cite{SCORE}. Meanwhile, DiMSC is designed based on the SVD of the adjacency matrix, and similar as \cite{rohe2011spectral,RSC,joseph2016impact,DISIM}, we may design spectral algorithms based on the regularized Laplacian matrix under DiDCMM. Furthermore, a limitation of our model is DiDCMM only models a directed mixed membership network with equal number of row communities and column communities, and how to overcome this limitation is an interesting problem. In this paper as well as \cite{DISIM,DSCORE,zhou2019analysis}, $K$ is assumed to be a known integer. An important future work is the estimation of $K$ for directed networks. Various techniques have been proposed to estimate $K$ for undirected network \cite{newman2016estimating,lei2016a,chen2018network,saldana2017how,bickel2016hypothesis,passino2020bayesian,hu2020using,jin2020estimating},  how to extend these techniques to determine $K$ for directed network leaves to be studied. \cite{zhao2012consistency,chen2018convexified,li2021convex} study the problem of community detection under SBM and DCSBM by modularity maximization approaches, and it is of interest to extend these techniques to study directed networks.



%
%
%
%



\bibliographystyle{agsm}
\bibliography{reference}
\vskip .65cm
\noindent
School of Mathematics, China University of Mining and Technology\\
Xuzhou 221116, China.
\vskip 2pt
\noindent
E-mail: qinghuan@cumt.edu.cn
\includepdf[pages=-]{Supplementary}
\end{document}


\def\spacingset#1{\renewcommand{\baselinestretch}%
		{#1}\small\normalsize} \spacingset{1}
	\if1\blind
	{
		\title{\bf Supplemental Materials for ``Directed degree corrected mixed membership model and estimating community memberships in directed networks''}
		\author{Huan Qing \thanks{School of Mathematics, China University of Mining and Technology  (E-mail: qinghuan@cumt.edu.cn).  }\hspace{.2cm}\\
			}
		\maketitle
	} \fi
	
	\if0\blind
	{
		\bigskip
		\bigskip
		\bigskip
		\begin{center}
			{\LARGE\bf Supplemental Materials for ``Directed degree corrected mixed membership model and estimating community memberships in directed networks''}
		\end{center}
		\medskip
	} \fi

In this document, we provide the technical proofs of lemmas and theorems in the main manuscript.
\spacingset{1.2}
\section{Proof for identifiability}
\subsection{Proof of Proposition 2.2}
\begin{proof}
Let $\Omega=U\Lambda V'$ be the compact singular value decomposition of $\Omega$. Lemma 3.1 gives $V=\Pi_{c}B_{c}\equiv\Pi_{c}V(\mathcal{I}_{c},:)$. Since $\Omega=\tilde{\Omega}$, $V$ also equals to $\tilde{\Pi}_{c}V(\mathcal{I}_{c},:)$, which gives that $\Pi_{c}=\tilde{\Pi}_{c}$.

Since $\Omega(\mathcal{I}_{r},\mathcal{I}_{c})=\Theta_{r}(\mathcal{I}_{r},\mathcal{I}_{r})\Pi_{r}(\mathcal{I}_{r},:)P\Pi'_{c}(\mathcal{I}_{c},:)=\Theta_{r}(\mathcal{I}_{r},\mathcal{I}_{r})P=U(\mathcal{I}_{r},:)\Lambda V'(\mathcal{I}_{c},:)$ by condition (I2), we have $\Theta_{r}(\mathcal{I}_{r},\mathcal{I}_{r})P=U(\mathcal{I}_{r},:)\Lambda V'(\mathcal{I}_{c},:)$, which gives that $\Theta_{r}(\mathcal{I}_{r},\mathcal{I}_{r})=\mathrm{diag}(U(\mathcal{I}_{r},:)\Lambda V'(\mathcal{I}_{c},:))$ \footnote{From this step, we see that if $P$'s diagonal entries are not ones, we can not obtain $\Theta_{r}(\mathcal{I}_{r},\mathcal{I}_{r})=\mathrm{diag}(U(\mathcal{I}_{r},:)\Lambda V'(\mathcal{I}_{c},:))$ which leads to a consequence that $\Theta_{r}(\mathcal{I}_{r},\mathcal{I}_{r})$ does not equal to $\tilde{\Theta}_{r}(\mathcal{I}_{r},\mathcal{I}_{r})$, hence condition (I1) is necessary.} by condition (I1). Since $\Omega=\tilde{\Omega}$, we also have $\tilde{\Theta}_{r}(\mathcal{I}_{r},\mathcal{I}_{r})=\mathrm{diag}(U(\mathcal{I}_{r},:)\Lambda V'(\mathcal{I}_{c},:))$, which gives that $\Theta_{r}(\mathcal{I}_{r},\mathcal{I}_{r})=\tilde{\Theta}_{r}(\mathcal{I}_{r},\mathcal{I}_{r})$. Since $\tilde{\Theta}_{r}(\mathcal{I}_{r},\mathcal{I}_{r})\tilde{P}$ also equals to $U(\mathcal{I}_{r},:)\Lambda V'(\mathcal{I}_{c},:)$, we have $P=\tilde{P}$.

Lemma 3.1 gives that $U=\Theta_{r}\Pi_{r}B_{r}$, where $B_{r}=\Theta^{-1}_{r}(\mathcal{I}_{r},\mathcal{I}_{r})U(\mathcal{I}_{r},:)$. Since $\Omega=\tilde{\Omega}$, we also have $U=\tilde{\Theta}_{r}\tilde{\Pi}_{r}\tilde{B}_{r}$. Since $\tilde{B}_{r}=\tilde{\Theta}^{-1}_{r}(\mathcal{I}_{r},\mathcal{I}_{r})U(\mathcal{I}_{r},:)=\Theta^{-1}_{r}(\mathcal{I}_{r},\mathcal{I}_{r})U(\mathcal{I}_{r},:)$, we have $\tilde{B}_{r}=B_{r}$. Since $U=\Theta_{r}\Pi_{r}B_{r}=\tilde{\Theta}_{r}\tilde{\Pi}_{r}\tilde{B}_{r}=\tilde{\Theta}_{r}\tilde{\Pi}_{r}B_{r}$, we have $\Theta_{r}\Pi_{r}=\tilde{\Theta}_{r}\tilde{\Pi}_{r}$. Since each row of $\Pi_{r}$ or $\tilde{\Pi}_{r}$ is a PMF, $\Theta_{r}=\tilde{\Theta}_{r}, \Pi_{r}=\tilde{\Pi}_{r}$, and the claim follows.
\begin{rem}\label{Reason}
(The reason that we do not model a directed network with mixed memberships where both row and column nodes have degree heterogeneities). Suppose both row and column nodes have degree heterogeneities in a mixed membership directed network. To model such directed network, the probability of generating an edge from row node $i$ to column node $j$ is
\begin{align*}
\mathbb{P}(A(i,j)=1)=\theta_{r}(i)\theta_{c}(j)\sum_{k=1}^{K}\sum_{l=1}^{K}\Pi_{r}(i,k)\Pi_{c}(j,l)P(k,l),
\end{align*}
where $\theta_{c}$ is an $n_{r}\times 1$ vector whose $j$-th entry is the degree heterogeneity of column node $j$. Set $\Omega=\mathbb{E}[A]$, then $\Omega=\Theta_{r}\Pi_{r}P\Pi'_{c}\Theta_{c}$, where $\Theta_{c}\in \mathbb{R}^{n_{c}\times n_{c}}$ is a diagonal matrix whose $j$-th diagonal entry $\theta_{c}(j)$. Set $\Omega=U\Lambda V'$ be the compact SVD of $\Omega$. Follow similar analysis as Lemma 3.1, we see that $U=\Theta_{r}\Pi_{r}B_{r}$ and $V=\Theta_{c}\Pi_{c}B_{c}$ (without causing confusion, we still use $B_{c}$ here for convenience.). For model identifiability, follow similar analysis as the proof of Proposition 2.2, since $\Omega(\mathcal{I}_{r},\mathcal{I}_{c})=\Theta_{r}(\mathcal{I}_{r},\mathcal{I}_{r})\Pi_{r}(\mathcal{I}_{r},;)P\Pi'_{c}(\mathcal{I}_{c},:)\Theta_{c}(\mathcal{I}_{c},\mathcal{I}_{c})=\Theta_{r}(\mathcal{I}_{r},\mathcal{I}_{r})P\Theta_{c}(\mathcal{I}_{c},\mathcal{I}_{c})=U(\mathcal{I}_{r},:)\Lambda V'(\mathcal{I}_{c},:)$, we see that $\Theta_{r}(\mathcal{I}_{r},\mathcal{I}_{r})P\Theta_{c}(\mathcal{I}_{c},\mathcal{I}_{c})=U(\mathcal{I}_{r},:)\Lambda V'(\mathcal{I}_{c},:)$. To obtain $\Theta_{r}(\mathcal{I}_{r},\mathcal{I}_{r})$ and $\Theta_{c}(\mathcal{I}_{c},\mathcal{I}_{c})$ from $U(\mathcal{I}_{r},:)\Lambda V'(\mathcal{I}_{c},:)$, when $P$ has unit diagonals, we see that it is impossible to recover  $\Theta_{r}(\mathcal{I}_{r},\mathcal{I}_{r})$ and $\Theta_{c}(\mathcal{I}_{c},\mathcal{I}_{c})$ unless we add a condition that $\Theta_{r}(\mathcal{I}_{r},\mathcal{I}_{r})=\Theta_{c}(\mathcal{I}_{c},\mathcal{I}_{c})$. Now suppose $\Theta_{r}(\mathcal{I}_{r},\mathcal{I}_{r})=\Theta_{c}(\mathcal{I}_{c},\mathcal{I}_{c})$ holds and call it condition (I3), we have $\Theta_{r}(\mathcal{I}_{r},\mathcal{I}_{r})P\Theta_{r}(\mathcal{I}_{r},\mathcal{I}_{r})=U(\mathcal{I}_{r},:)\Lambda V'(\mathcal{I}_{c},:)$, hence $\Theta_{r}(\mathcal{I}_{r},\mathcal{I}_{r})=\Theta_{c}(\mathcal{I}_{c},\mathcal{I}_{c})=\sqrt{\mathrm{diag}(U(\mathcal{I}_{r},:)\Lambda V'(\mathcal{I}_{c},:))}$ when $P$ has unit diagonals. However, condition (I3) is nontrivial since it requires $\Theta_{r}(\mathcal{I}_{r},\mathcal{I}_{r})=\Theta_{c}(\mathcal{I}_{c},\mathcal{I}_{c})$ and we always prefer a directed network that there is no connections between row nodes degree heterogeneities and column nodes degree heterogeneities. For example, when all nodes are pure in a directed network, \cite{DISIM} models such directed network using model DC-ScBM such that $\Omega=\Theta_{r}\Pi_{r}P\Pi'_{c}\Theta_{c}$ when all nodes are pure, and $\Theta_{r}$ and $\Theta_{c}$ are independent under DC-ScBM. Due to the fact that condition (I3) is nontrivial, we do not model a mixed membership directed network with all nodes have degree heterogeneities.
\end{rem}
\end{proof}
\section{Ideal Simplex, Ideal Cone}
\subsection{Proof of Lemma 3.1}
\begin{proof}
First, we consider $U$ and $V$. Since $\Omega=U\Lambda V'$, we have $U=\Omega V\Lambda^{-1}$ since $V'V=I_{K}$. Recall that $\Omega=\Theta_{r}\Pi_{r}P\Pi'_{c}$, we have $U=\Theta_{r}\Pi_{r}P\Pi'_{c}V\Lambda^{-1}=\Theta_{r}\Pi_{r}B_{r}$, where we set $B_{r}=P\Pi'_{c}V\Lambda^{-1}$ and sure it is unique. Since $U(\mathcal{I}_{r},:)=\Theta_{r}(\mathcal{I}_{r},\mathcal{I}_{r})\Pi_{r}(\mathcal{I}_{r},:)B_{r}=\Theta_{r}(\mathcal{I}_{r},\mathcal{I}_{r})B_{r}$, we have $B_{r}=\Theta^{-1}_{r}(\mathcal{I}_{r},\mathcal{I}_{r})U(\mathcal{I}_{r},:)$.

Similarly, since $\Omega=U\Lambda V'$, we have $V'=\Lambda^{-1}U'\Omega$ since $U'U=I_{K}$, hence $V=\Omega' U\Lambda^{-1}$. Recall that $\Omega=\Theta_{r}\Pi_{r}P\Pi'_{c}$, we have $V=(\Theta_{r}\Pi_{r}P\Pi'_{c})'U\Lambda^{-1}=\Pi_{c}P'\Pi'_{r}\Theta_{r}U\Lambda^{-1}=\Pi_{c}B_{c}$, where we set $B_{c}=P'\Pi'_{r}\Theta_{r}U\Lambda^{-1}$ and sure it is unique. Since $V(\mathcal{I}_{c},:)=\Pi_{c}(\mathcal{I}_{c},:)B_{c}=B_{c}$, we have $B_{c}=V(\mathcal{I}_{c},:)$. Meanwhile, for $1\leq j\leq n_{c}$, we have $V(j,:)=e'_{j}\Pi_{c}B_{c}=\Pi_{c}(j,:)B_{c}$. Hence, we have $V(j,:)=V(\bar{j},:)$ as long as $\Pi_{c}(j,:)=\Pi_{c}(\bar{j},:)$.

Now, we show the Ideal Cone structure appeared in $U_{*}$. For convenience, set $M=\Pi_{r}B_{r}$, hence $U=\Theta_{r}\Pi_{r}B_{r}$ gives $U=\Theta_{r}M$. Hence, we have $U(i,:)=e'_{i}U=\Theta_{r}(i,i)M(i,:)$. Therefore, $U_{*}(i,:)=\frac{U(i,:)}{\|U(i,:)\|_{F}}=\frac{M(i,:)}{\|M(i,:)\|_{F}}$, combine it with the fact that $B_{r}=\Theta^{-1}_{r}(\mathcal{I}_{r},\mathcal{I}_{r})U(\mathcal{I}_{r},:)$, we have
\begin{flalign*}
U_{*}&=\begin{bmatrix}
\frac{1}{\|M(1,:)\|_{F}} &  & & \\
& \frac{1}{\|M(2,:)\|_{F}}& &\\
& & \ddots&\\
&&&\frac{1}{\|M(n_{r},:)\|_{F}}
\end{bmatrix}\Pi_{r}B_{r}
=\begin{bmatrix}
\Pi_{r}(1,:)/\|M(1,:)\|_{F}\\
\Pi_{r}(2,:)/\|M(2,:)\|_{F}\\
\vdots\\
\Pi_{r}(n_{r},:)/\|M(n_{r},:)\|_{F} \end{bmatrix}B_{r}\\
&=\begin{bmatrix}
\Pi_{r}(1,:)/\|M(1,:)\|_{F}\\
\Pi_{r}(2,:)/\|M(2,:)\|_{F}\\
\vdots\\
\Pi_{r}(n_{r},:)/\|M(n_{r},:)\|_{F}	\end{bmatrix}\Theta^{-1}_{r}(\mathcal{I}_{r},\mathcal{I}_{r})N_{U}^{-1}(\mathcal{I}_{r},\mathcal{I}_{r})N_{U}(\mathcal{I}_{r},\mathcal{I}_{r})U(\mathcal{I}_{r},:)\\
&=\begin{bmatrix}
\Pi_{r}(1,:)/\|M(1,:)\|_{F}\\
\Pi_{r}(2,:)/\|M(2,:)\|_{F}\\
\vdots\\
\Pi_{r}(n_{r},:)/\|M(n_{r},:)\|_{F}
\end{bmatrix}\Theta^{-1}_{r}(\mathcal{I}_{r},\mathcal{I}_{r})N_{U}^{-1}(\mathcal{I}_{r},\mathcal{I}_{r})U_{*}(\mathcal{I}_{r},:).
\end{flalign*}
Therefore, we have
\begin{align*}
Y=\begin{bmatrix}
\Pi_{r}(1,:)/\|M(1,:)\|_{F}\\
\Pi_{r}(2,:)/\|M(2,:)\|_{F}\\
\vdots\\
\Pi_{r}(n_{r},:)/\|M(n_{r},:)\|_{F} \end{bmatrix}\Theta^{-1}_{r}(\mathcal{I}_{r},\mathcal{I}_{r})N_{U}^{-1}(\mathcal{I}_{r},\mathcal{I}_{r})=N_{M}\Pi_{r}\Theta^{-1}_{r}(\mathcal{I}_{r},\mathcal{I}_{r})N_{U}^{-1}(\mathcal{I}_{r},\mathcal{I}_{r}),
\end{align*}
where $N_{M}$ is a diagonal matrix with $N_{M}(i,i)=\frac{1}{\|M(i,:)\|_{F}}$ for $1\leq i\leq n_{r}$. Sure, all entries of $Y$ are nonnegative. And since we assume that each community has at least one pure node, no row of $Y$ is 0.
	
	Then we prove that $U_{*}(i,:)=U_{*}(\bar{i},:)$ when $\Pi_{r}(i,:)=\Pi_{r}(\bar{i},:)$. For $1\leq i\leq n_{r}$, we have
\begin{flalign*}
U_{*}(i,:)&=e'_{i}U_{*}=\frac{1}{\|M(i,:)\|_{F}}e'_{i}M=\frac{1}{\|\Pi_{r}(i,:)B_{r}\|_{F}}\Pi_{r}(i,:)B_{r}, \end{flalign*}
and the claim follows immediately.
\end{proof}
\subsection{Proof of Lemma 3.2}
\begin{proof}
Since $I=U'U=B'_{r}\Pi'_{r}\Theta^{2}_{r}\Pi_{r}B_{r}=U'(\mathcal{I}_{r},:)\Theta^{-1}_{r}(\mathcal{I}_{r},\mathcal{I}_{r})\Pi'_{r}\Theta^{2}_{r}\Pi_{r}\Theta^{-1}(\mathcal{I}_{r},\mathcal{I}_{r})U(\mathcal{I}_{r},:)$ and $\mathrm{rank}(U(\mathcal{I}_{r},:))=K$ (i.e., the inverse of $U(\mathcal{I}_{r},:)$ exists), we have $(U(\mathcal{I}_{r},:)U'(\mathcal{I}_{r},:))^{-1}=\Theta^{-1}_{r}(\mathcal{I}_{r},\mathcal{I}_{r})\Pi'_{r}\Theta^{2}_{r}\Pi_{r}\Theta^{-1}_{r}(\mathcal{I}_{r},\mathcal{I}_{r})$. Since $U_{*}(\mathcal{I}_{r},:)=N_{U}(\mathcal{I}_{r},\mathcal{I}_{r})U(\mathcal{I}_{r},:)$, we have
\begin{align*}
(U_{*}(\mathcal{I}_{r},:)U'_{*}(\mathcal{I}_{r},:))^{-1}=N_{U}^{-1}(\mathcal{I}_{r},\mathcal{I}_{r})\Theta^{-1}(\mathcal{I}_{r},\mathcal{I}_{r})\Pi'_{r}\Theta^{2}_{r}\Pi_{r}\Theta^{-1}_{r}(\mathcal{I}_{r},\mathcal{I}_{r})N_{U}^{-1}(\mathcal{I}_{r},\mathcal{I}_{r}).
\end{align*}
Since all entries of $N_{U}^{-1}(\mathcal{I}_{r},\mathcal{I}_{r}), \Pi_{r}, \Theta_{r}$ and nonnegative and $N,\Theta_{r}$ are diagonal matrices, we see that all entries of $(U_{*}(\mathcal{I}_{r},:)U'_{*}(\mathcal{I}_{r},:))^{-1}$ are nonnegative and its diagonal entries are strictly positive, hence we have $(U_{*}(\mathcal{I}_{r},:)U'_{*}(\mathcal{I}_{r},:))^{-1}\mathbf{1}>0$. \end{proof}
\subsection{Proof of Theorem 3.3}
\begin{proof}
For column nodes, Remark \ref{inputVinSP} guarantees that SP algorithm returns $\mathcal{I}_{c}$ when the input is $V$ with $K$ column communities, hence Ideal DiMSC recovers $\Pi_{c}$ exactly. For row nodes, Remark \ref{inputUstarinSVMCone} guarantees that SVM-cone algorithm returns $\mathcal{I}_{r}$ when the input is $U_{*}$ with $K$ row communities, hence Ideal DiMSC recovers $\Pi_{r}$ exactly, and this theorem follows.
\end{proof}
\subsection{Proof of Lemma 3.4}
\begin{proof}
By Lemma 3.1, we know that $V=\Pi_{c}V(\mathcal{I}_{c},:)$, which gives that $V_{2}=VV'=\Pi_{c}V(\mathcal{I}_{c},:)V'=\Pi_{c}(VV')(I_{c},:)=\Pi_{c}V_{2}(\mathcal{I}_{c},:)$. For $U$, since $U=\Theta_{r}\Pi_{r}\Theta^{-1}_{r}(\mathcal{I}_{r},\mathcal{I}_{r})U(\mathcal{I}_{r},:)$ by Lemma 3.1, we have $U_{2}=UU'=\Theta_{r}\Pi_{r}\Theta^{-1}_{r}(\mathcal{I}_{r},\mathcal{I}_{r})U(\mathcal{I}_{r},:)U'=\Theta_{r}\Pi_{r}\Theta^{-1}_{r}(\mathcal{I}_{r},\mathcal{I}_{r})(UU')(\mathcal{I}_{r},:)=\Theta_{r}\Pi_{r}\Theta^{-1}_{r}(\mathcal{I}_{r},\mathcal{I}_{r})U_{2}(\mathcal{I}_{r},:)$. Set $M_{2}=\Pi_{r}\Theta^{-1}_{r}(\mathcal{I}_{r},\mathcal{I}_{r})U_{2}(\mathcal{I}_{r},:)$, we have $U_{2}=\Theta_{r}M_{2}$. Then follow similar proof as Lemma 3.1, we have $U_{*,2}=Y_{2}U_{*,2}(\mathcal{I}_{r},:)$, where $Y_{2}=N_{M_{2}}\Pi_{r}\Theta^{-1}_{r}(\mathcal{I}_{r},\mathcal{I}_{r})N^{-1}_{U_{2}}(\mathcal{I}_{r},\mathcal{I}_{r})$, and $N_{M_{2}}, N_{U_{2}}$ are $n_{r}\times n_{r}$ diagonal matrices whose $i$-th diagonal entries are $\frac{1}{\|M_{2}(i,:)\|_{F}},\frac{1}{\|U_{2}(i,:)\|_{F}}$, respectively. Since $\|U_{2}(i,:)\|_{F}=\|U(i,:)U'\|_{F}=\|U(i,:)\|_{F}$, we have $N_{U_{2}}=N_{U}$. Since $\|M_{2}(i,:)|_{F}=\|\Pi_{r}\Theta^{-1}_{r}(\mathcal{I}_{r},\mathcal{I}_{r})U_{2}(\mathcal{I}_{r},:)\|_{F}=\|\Pi_{r}\Theta^{-1}_{r}(\mathcal{I}_{r},\mathcal{I}_{r})U(\mathcal{I}_{r},:)U'\|_{F}=\|M(i,:)\|_{F}$, we have $N_{M_{2}}=N_{M}$. Hence, $Y_{2}\equiv Y$ and the claim follows.
\end{proof}
\subsection{Proof of Lemma 3.5}
\begin{proof}
For column nodes, Lemma 3.2 \cite{mao2020estimating} gives $\hat{\mathcal{I}}_{c}=\hat{\mathcal{I}}_{c,2}$ (i.e., SP algorithm will return the same indices on both $\hat{V}$ and $\hat{V}_{2}$.), which gives that $\hat{V}_{2}\hat{V}'_{2}(\hat{\mathcal{I}}_{c,2},:)=\hat{V}_{2}\hat{V}'_{2}(\hat{\mathcal{I}}_{c},:)=\hat{V}\hat{V}'((\hat{V}\hat{V}')(\hat{\mathcal{I}}_{c},:))'=\hat{V}\hat{V}'(\hat{V}(\hat{\mathcal{I}}_{c},:)\hat{V}')'=\hat{V}\hat{V}'\hat{V}\hat{V}'(\hat{\mathcal{I}}_{c},:)=\hat{V}\hat{V}'(\hat{\mathcal{I}}_{c},:)$, and $\hat{V}_{2}(\hat{\mathcal{I}}_{c,2},:)\hat{V}'_{2}(\hat{\mathcal{I}}_{c,2},:)=\hat{V}_{2}(\hat{\mathcal{I}}_{c},:)\hat{V}'_{2}(\hat{\mathcal{I}}_{c},:)=\hat{V}(\hat{\mathcal{I}}_{c},:)\hat{V}'(\hat{V}(\hat{\mathcal{I}}_{c},:)\hat{V}')'=\hat{V}(\hat{\mathcal{I}}_{c},:)\hat{V}'(\hat{\mathcal{I}}_{c},:)$. Therefore, we have $\hat{Z}_{c,2}=\hat{Z}_{c}, \hat{\Pi}_{c,2}=\hat{\Pi}_{c}$.

For row nodes, Lemma G.1 \cite{MaoSVM} guarantees that $\hat{\mathcal{I}}_{r}=\hat{\mathcal{I}}_{r,2}$ (i.e., SVM-cone algorithm will return the same indices on both $\hat{U}_{*}$ and $\hat{U}_{*,2}$.), so immediately we have $\hat{J}_{*,2}=\hat{J}_{*}$. Since $\hat{U}_{*,2}(\hat{\mathcal{I}}_{r,2},:)=\hat{U}_{*,2}(\hat{\mathcal{I}}_{r},:)=N_{\hat{U}}(\hat{\mathcal{I}}_{r},\hat{\mathcal{I}}_{r})\hat{U}_{2}(\hat{\mathcal{I}}_{r},:)=N_{\hat{U}}(\hat{\mathcal{I}}_{r},\hat{\mathcal{I}}_{r})\hat{U}(\hat{\mathcal{I}}_{r},:)\hat{U}'=\hat{U}_{*}(\hat{\mathcal{I}}_{r},:)\hat{U}'$, we have $\hat{U}_{2}\hat{U}'_{*,2}(\hat{\mathcal{I}}_{r,2},:)=\hat{U}_{2}\hat{U}'_{*,2}(\hat{\mathcal{I}}_{r},:)=\hat{U}\hat{U}'\hat{U}\hat{U}'_{*}(\hat{\mathcal{I}}_{r},:)=\hat{U}\hat{U}'_{*}(\hat{\mathcal{I}}_{r},:)$ and $(\hat{U}_{*,2}(\hat{\mathcal{I}}_{r,2},:)\hat{U}'_{*,2}(\hat{\mathcal{I}}_{r,2},:))^{-1}=(\hat{U}_{*,2}(\hat{\mathcal{I}}_{r},:)\hat{U}'_{*,2}(\hat{\mathcal{I}}_{r},:))^{-1}=(\hat{U}_{*}(\hat{\mathcal{I}}_{r},:)\hat{U}'_{*}(\hat{\mathcal{I}}_{r},:))^{-1}$, which give that $\hat{Y}_{*,2}=\hat{Y}_{*}$, and the claim follows immediately.
\end{proof}
\section{Basic properties of $\Omega$}
\begin{lem}\label{P2}
Under $DiDCMM_{n_{r},n_{c}}(K,P,\Pi_{r}, \Pi_{c}, \Theta_{r})$, we have
\begin{align*}
&\frac{\theta_{r,\mathrm{min}}}{\theta_{r,\mathrm{max}}\sqrt{K\lambda_{1}(\Pi'_{r}\Pi_{r})}}\leq\|U(i,:)\|_{F}\leq \frac{\theta_{r,\mathrm{max}}}{\theta_{r,\mathrm{min}}\sqrt{\lambda_{K}(\Pi'_{r}\Pi_{r})}}, \qquad 1\leq i\leq n_{r},
\\
&\sqrt{\frac{1}{K\lambda_{1}(\Pi'_{c}\Pi_{c})}}\leq\|V(j,:)\|_{F}\leq \sqrt{\frac{1}{\lambda_{K}(\Pi'_{c}\Pi_{c})}}, \qquad 1\leq j\leq n_{c}.
\end{align*}
\end{lem}
\begin{proof}
Since $I=U'U=U'(\mathcal{I}_{r},:)\Theta^{-1}_{r}(\mathcal{I}_{r},\mathcal{I}_{r})\Pi'_{r}\Theta^{2}_{r}\Pi_{r}\Theta^{-1}(\mathcal{I}_{r},\mathcal{I}_{r})U(\mathcal{I}_{r},:)$,
we have
\begin{align*}
((\Theta^{-1}_{r}(\mathcal{I}_{r},\mathcal{I}_{r})U(\mathcal{I}_{r},:))((\Theta^{-1}_{r}(\mathcal{I}_{r},\mathcal{I}_{r})U(\mathcal{I}_{r},:))')^{-1}=\Pi'_{r}\Theta^{2}_{r}\Pi_{r},
\end{align*}
which gives that
\begin{align*}
\mathrm{max}_{k}\|e'_{k}(\Theta^{-1}_{r}(\mathcal{I}_{r},\mathcal{I}_{r})U(\mathcal{I}_{r},:))\|_{F}^{2}&=\mathrm{max}_{k}e'_{k}(\Theta^{-1}_{r}(\mathcal{I}_{r},\mathcal{I}_{r})U(\mathcal{I}_{r},:))(\Theta^{-1}_{r}(\mathcal{I}_{r},\mathcal{I}_{r})U(\mathcal{I}_{r},:))'e_{k}\\
&\leq \mathrm{max}_{\|x\|_{F}=1}x'(\Theta^{-1}_{r}(\mathcal{I}_{r},\mathcal{I}_{r})U(\mathcal{I}_{r},:))(\Theta^{-1}_{r}(\mathcal{I}_{r},\mathcal{I}_{r})U(\mathcal{I}_{r},:))'x\\
&=\lambda_{1}((\Theta^{-1}_{r}(\mathcal{I}_{r},\mathcal{I}_{r})U(\mathcal{I}_{r},:))(\Theta^{-1}_{r}(\mathcal{I}_{r},\mathcal{I}_{r})U(\mathcal{I}_{r},:))')\\
&=\frac{1}{\lambda_{K}(\Pi'_{r}\Theta^{2}_{r}\Pi_{r})}\leq \frac{1}{\theta^{2}_{r,\mathrm{min}}\lambda_{K}(\Pi'_{r}\Pi_{r})}.
\end{align*}
Similarly, we have
\begin{align*}
\mathrm{min}_{k}\|e'_{k}(\Theta^{-1}_{r}(\mathcal{I}_{r},\mathcal{I}_{r})U(\mathcal{I}_{r},:))\|_{F}^{2}\geq\frac{1}{\lambda_{1}(\Pi'_{r}\Theta^{2}_{r}\Pi_{r})}\geq \frac{1}{\theta^{2}_{r,\mathrm{max}}\lambda_{1}(\Pi'_{r}\Pi_{r})}.
\end{align*}
Since $U(i,:)=e'_{i}U=e'_{i}\Theta_{r}\Pi_{r}\Theta^{-1}_{r}(\mathcal{I}_{r},\mathcal{I}_{r})U(\mathcal{I}_{r},:)=\theta_{r}(i)\Pi_{r}(i,:)\Theta^{-1}_{r}(\mathcal{I}_{r},\mathcal{I}_{r})U(\mathcal{I}_{r},:)$ for $1\leq i\leq n_{r}$, we have
\begin{align*}
\|U(i,:)\|_{F}&=\|\theta_{r}(i)\Pi_{r}(i,:)\Theta^{-1}_{r}(\mathcal{I}_{r},\mathcal{I}_{r})U(\mathcal{I}_{r},:)\|_{F}=\theta_{r}(i)\|\Pi_{r}(i,:)\Theta^{-1}_{r}(\mathcal{I}_{r},\mathcal{I}_{r})U(\mathcal{I}_{r},:)\|_{F}\\
&\leq \theta_{r}(i) \mathrm{max}_{i}\|\Pi_{r}(i,:)\|_{F}\mathrm{max}_{i}\|e'_{i}(\Theta^{-1}_{r}(\mathcal{I}_{r},\mathcal{I}_{r})U(\mathcal{I}_{r},:))\|_{F}\leq \theta_{r}(i)\mathrm{max}_{i}\|e'_{i}(\Theta^{-1}_{r}(\mathcal{I}_{r},\mathcal{I}_{r})U(\mathcal{I}_{r},:))\|_{F}\\
&\leq\frac{\theta_{r,\mathrm{max}}}{\theta_{r,\mathrm{min}}\sqrt{\lambda_{K}(\Pi'_{r}\Pi_{r})}}.
\end{align*}
Similarly, we have
\begin{align*}
\|U(i,:)\|_{F}&\geq \theta_{r}(i) \mathrm{min}_{i}\|\Pi_{r}(i,:)\|_{F}\mathrm{min}_{i}\|e'_{i}(\Theta^{-1}_{r}(\mathcal{I}_{r},\mathcal{I}_{r})U(\mathcal{I}_{r},:))\|_{F}\\
&\geq \theta_{r}(i)\mathrm{min}_{i}\|e'_{i}(\Theta^{-1}_{r}(\mathcal{I}_{r},\mathcal{I}_{r})U(\mathcal{I}_{r},:))\|_{F}/\sqrt{K}\geq\frac{\theta_{r,\mathrm{min}}}{\theta_{r,\mathrm{max}}\sqrt{K\lambda_{1}(\Pi'_{r}\Pi_{r})}}.
\end{align*}
For $\|V(j,:)\|_{F}$, since $V=\Pi_{c}B_{c}$, we have
\begin{align*}
\mathrm{min}_{j}\|e'_{j}V\|^{2}_{F}&=\mathrm{min}_{j}e'_{j}VV'e_{j}=\mathrm{min}_{j}\Pi_{c}(j,:)B_{c}B'_{c}\Pi'_{c}(j,:)=\mathrm{min}_{j}\|\Pi_{c}(j,:)\|^{2}_{F}\frac{\Pi_{c}(j,:)}{\|\Pi_{c}(j,:)\|_{F}}B_{c}B'_{c}\frac{\Pi'_{c}(j,:)}{\|\Pi_{c}(j,:)\|_{F}}\\
&\geq\mathrm{min}_{j}\|\Pi_{c}(j,:)\|^{2}_{F}\mathrm{min}_{\|x\|_{F}=1}x'B_{c}B'_{c}x=\mathrm{min}_{j}\|\Pi_{c}(j,:)\|^{2}_{F}\lambda_{K}(B_{c}B'_{c})\\
&\overset{\mathrm{By~Lemma~}\ref{P3}}{=}\frac{\mathrm{min}_{j}\|\Pi_{c}(j,:)\|^{2}_{F}}{\lambda_{1}(\Pi'_{c}\Pi_{c})}\geq \frac{1}{K\lambda_{1}(\Pi'_{c}\Pi_{c})}.
\end{align*}
Meanwhile,
\begin{align*}
\mathrm{max}_{j}\|e'_{j}V\|^{2}_{F}&=\mathrm{max}_{j}\|\Pi_{c}(j,:)\|^{2}_{F}\frac{\Pi_{c}(j,:)}{\|\Pi_{c}(j,:)\|_{F}}B_{c}B'_{c}\frac{\Pi'_{c}(j,:)}{\|\Pi_{c}(j,:)\|_{F}}\\
&\leq\mathrm{max}_{j}\|\Pi_{c}(j,:)\|^{2}_{F}\mathrm{max}_{\|x\|_{F}=1}x'B_{c}B'_{c}x=\mathrm{max}_{j}\|\Pi_{c}(j,:)\|^{2}_{F}\lambda_{K}(B_{c}B'_{c})\\
&\overset{\mathrm{By~Lemma~}\ref{P3}}{=}\frac{\mathrm{max}_{j}\|\Pi_{c}(j,:)\|^{2}_{F}}{\lambda_{K}(\Pi'_{c}\Pi_{c})}\leq \frac{1}{\lambda_{K}(\Pi'_{c}\Pi_{c})}.
\end{align*}
\end{proof}
\begin{lem}\label{P3}
Under $DiDCMM_{n_{r},n_{c}}(K,P,\Pi_{r}, \Pi_{c}, \Theta_{r})$, we have
\begin{align*}	
&\frac{\theta^{2}_{r,\mathrm{min}}\lambda_{K}(\Pi'_{r}\Pi_{r})}{\theta^{2}_{r,\mathrm{max}}\lambda_{1}(\Pi'_{r}\Pi_{r})}\leq \lambda_{K}(U_{*}(\mathcal{I}_{r},:)U'_{*}(\mathcal{I}_{r},:)), \lambda_{1}(U_{*}(\mathcal{I}_{r},:)U'_{*}(\mathcal{I}_{r},:))\leq\frac{\theta^{2}_{r,\mathrm{max}}K\lambda_{1}(\Pi'_{r}\Pi_{r})}{\theta^{2}_{r,\mathrm{min}}\lambda_{K}(\Pi'_{r}\Pi_{r})},\\
&\mathrm{and~}\lambda_{1}(B_{c}B'_{c})=\frac{1}{\lambda_{K}(\Pi'_{c}\Pi_{c})},\lambda_{K}(B_{c}B'_{c})=\frac{1}{\lambda_{1}(\Pi'_{c}\Pi_{c})}.
\end{align*}
\end{lem}
\begin{proof}
Recall that $V=\Pi_{c}B_{c}$ and $V'V=I$, we have $I=B'_{c}\Pi'_{c}\Pi_{c}B_{c}$. As $B_{c}$ is full rank, we have $\Pi'_{c}\Pi_{c}=(B_{c}B'_{c})^{-1}$, which gives
\begin{align*}
\lambda_{1}(B_{c}B'_{c})=\frac{1}{\lambda_{K}(\Pi'_{c}\Pi_{c})},\lambda_{K}(B_{c}B'_{c})=\frac{1}{\lambda_{1}(\Pi'_{c}\Pi_{c})}.
\end{align*}
By the proof of Lemma 3.2, we know that
\begin{align*}
(U_{*}(\mathcal{I}_{r},:)U'_{*}(\mathcal{I}_{r},:))^{-1}=N_{U}^{-1}(\mathcal{I}_{r},\mathcal{I}_{r})\Theta^{-1}(\mathcal{I}_{r},\mathcal{I}_{r})\Pi'_{r}\Theta^{2}_{r}\Pi_{r}\Theta^{-1}_{r}(\mathcal{I}_{r},\mathcal{I}_{r})N_{U}^{-1}(\mathcal{I}_{r},\mathcal{I}_{r}),
\end{align*}
which gives that
\begin{align*}
U_{*}(\mathcal{I}_{r},:)U'_{*}(\mathcal{I}_{r},:)=N_{U}(\mathcal{I}_{r},\mathcal{I}_{r})\Theta(\mathcal{I}_{r},\mathcal{I}_{r})(\Pi'_{r}\Theta^{2}_{r}\Pi_{r})^{-1}\Theta_{r}(\mathcal{I}_{r},\mathcal{I}_{r})N_{U}(\mathcal{I}_{r},\mathcal{I}_{r}).
\end{align*}
Then we have
\begin{align*} &\lambda_{1}(U_{*}(\mathcal{I}_{r},:)U'_{*}(\mathcal{I}_{r},:))=\lambda_{1}(N_{U}(\mathcal{I}_{r},\mathcal{I}_{r})\Theta(\mathcal{I}_{r},\mathcal{I}_{r})(\Pi'_{r}\Theta^{2}_{r}\Pi_{r})^{-1}\Theta_{r}(\mathcal{I}_{r},\mathcal{I}_{r})N_{U}(\mathcal{I}_{r},\mathcal{I}_{r}))\\
&=\lambda_{1}(N^{2}_{U}(\mathcal{I}_{r},\mathcal{I}_{r})\Theta^{2}_{r}(\mathcal{I}_{r},\mathcal{I}_{r})(\Pi'_{r}\Theta^{2}_{r}\Pi_{r})^{-1})\leq\lambda^{2}_{1}(N_{U}(\mathcal{I}_{r},\mathcal{I}_{r})\Theta_{r}(\mathcal{I}_{r},\mathcal{I}_{r}))\lambda_{1}((\Pi'_{r}\Theta^{2}_{r}\Pi_{r})^{-1})\\ &=\lambda^{2}_{1}(N_{U}(\mathcal{I}_{r},\mathcal{I}_{r})\Theta_{r}(\mathcal{I}_{r},\mathcal{I}_{r}))/\lambda_{K}(\Pi'_{r}\Theta^{2}_{r}\Pi_{r})\leq(\mathrm{max}_{i\in \mathcal{I}_{r}}\theta_{r}(i)/\|U(i,:)\|_{F})^{2}/\lambda_{K}(\Pi'_{r}\Theta^{2}_{r}\Pi_{r})\\ &\leq \frac{\theta^{2}_{r,\mathrm{max}}K\lambda_{1}(\Pi'_{r}\Pi_{r})}{\lambda_{K}(\Pi'_{r}\Theta^{2}_{r}\Pi_{r})}\leq \frac{\theta^{2}_{r,\mathrm{max}}K\lambda_{1}(\Pi'_{r}\Pi_{r})}{\theta^{2}_{r,\mathrm{min}}\lambda_{K}(\Pi'_{r}\Pi_{r})}.
\end{align*}
Similarly, we have
\begin{align*} &\lambda_{K}(U_{*}(\mathcal{I}_{r},:)U'_{*}(\mathcal{I}_{r},:))=\lambda_{K}(N_{U}(\mathcal{I}_{r},\mathcal{I}_{r})\Theta(\mathcal{I}_{r},\mathcal{I}_{r})(\Pi'_{r}\Theta^{2}_{r}\Pi_{r})^{-1}\Theta_{r}(\mathcal{I}_{r},\mathcal{I}_{r})N_{U}(\mathcal{I}_{r},\mathcal{I}_{r}))\\
&=\lambda_{K}(N^{2}_{U}(\mathcal{I}_{r},\mathcal{I}_{r})\Theta^{2}_{r}(\mathcal{I}_{r},\mathcal{I}_{r})(\Pi'_{r}\Theta^{2}_{r}\Pi_{r})^{-1})\geq\lambda^{2}_{K}(N_{U}(\mathcal{I}_{r},\mathcal{I}_{r})\Theta_{r}(\mathcal{I}_{r},\mathcal{I}_{r}))\lambda_{K}((\Pi'_{r}\Theta^{2}_{r}\Pi_{r})^{-1})\\ &=\lambda^{2}_{K}(N_{U}(\mathcal{I}_{r},\mathcal{I}_{r})\Theta_{r}(\mathcal{I}_{r},\mathcal{I}_{r}))/\lambda_{1}(\Pi'_{r}\Theta^{2}_{r}\Pi_{r})\geq(\mathrm{min}_{i\in \mathcal{I}_{r}}\theta_{r}(i)/\|U(i,:)\|_{F})^{2}/\lambda_{1}(\Pi'_{r}\Theta^{2}_{r}\Pi_{r})\\ &\geq \frac{\theta^{2}_{r,\mathrm{min}}\lambda_{K}(\Pi'_{r}\Pi_{r})}{\lambda_{1}(\Pi'_{r}\Theta^{2}_{r}\Pi_{r})}\geq \frac{\theta^{2}_{r,\mathrm{min}}\lambda_{K}(\Pi'_{r}\Pi_{r})}{\theta^{2}_{r,\mathrm{max}}\lambda_{1}(\Pi'_{r}\Pi_{r})}.
\end{align*}
\end{proof}
\begin{lem}\label{P4}
Under $DiDCMM_{n_{r},n_{c}}(K,P,\Pi_{r}, \Pi_{c}, \Theta_{r})$, we have
\begin{align*} \sigma_{K}(\Omega)\geq\theta_{r,\mathrm{min}}\sigma_{K}(P)\sigma_{K}(\Pi_{r})\sigma_{K}(\Pi_{c}), \sigma_{1}(\Omega)\leq\theta_{r,\mathrm{max}}\sigma_{1}(P)\sigma_{1}(\Pi_{r})\sigma_{1}(\Pi_{c}). \end{align*}
\end{lem}
\begin{proof}
For $\sigma_{K}(\Omega)$, we have
\begin{align*}
\sigma^{2}_{K}(\Omega)=\lambda_{K}(\Omega\Omega')&=\lambda_{K}(\Theta_{r}\Pi_{r}P\Pi'_{c}\Pi_{c}P'\Pi'_{r}\Theta_{r})=\lambda_{K}(\Theta^{2}_{r}\Pi_{r}P\Pi'_{c}\Pi_{c}P'\Pi'_{r})\geq\theta^{2}_{r,\mathrm{min}}\lambda_{K}(\Pi_{r}'\Pi_{r}P\Pi'_{c}\Pi_{c}P')\\
&\geq \theta^{2}_{r,\mathrm{min}}\lambda_{K}(\Pi'_{r}\Pi_{r})\lambda_{K}(P\Pi'_{c}\Pi_{c}P')=\theta^{2}_{r,\mathrm{min}}\lambda_{K}(\Pi'_{r}\Pi_{r})\lambda_{K}(\Pi'_{c}\Pi_{c}P'P)\\
&\geq\theta^{2}_{r,\mathrm{min}}\lambda_{K}(\Pi'_{r}\Pi_{r})\lambda_{K}(\Pi'_{c}\Pi_{c})\lambda_{K}(PP')=\theta^{2}_{r,\mathrm{min}}\sigma^{2}_{K}(\Pi_{r})\sigma^{2}_{K}(\Pi_{c})\sigma^{2}_{K}(P),
\end{align*}
where we have used the fact for any matrices $X, Y$, the nonzero eigenvalues of $XY$ are the same as the nonzero eigenvalues of $YX$. Follow similar analysis, the lemma follows.
\end{proof}
\section{Proof of consistency for DiMSC}
\subsection{Proof of Lemma 4.2}
\begin{proof}
Let $e_{i}$ be an $n_{r}\times 1$ vector, where $e_{i}(i)=1$ and 0 elsewhere, for row nodes $1\leq i\leq n_{r}$, and $\tilde{e}_{j}$ be an $n_{c}\times 1$ vector, where $\tilde{e}_{j}(j)=1$ and 0 elsewhere, for column nodes $1\leq j\leq n_{c}$. Set $W=\sum_{i=1}^{n^{r}}\sum_{j=1}^{n_{c}}W(i,j)e_{i}\tilde{e}'_{j}$, where $W=A-\Omega$. Set $W^{(i,j)}=W(i,j)e_{i}\tilde{e}'_{j}$, for $1\leq i\leq n_{r}, 1\leq j\leq n_{c}$. Then, we have $\mathbb{E}(W^{(i,j)})=0$. For $1\leq i\leq n_{r}, 1\leq j\leq n_{c}$, we have
\begin{align*}
\|W^{(i,j)}\|&=\|W(i,j)e_{i}\tilde{e}'_{j}\|=|A(i,j)-\Omega(i,j)|\leq 1.
\end{align*}
Next  we consider the variance parameter
\begin{align*}
\sigma^{2}:=\mathrm{max}(\|\sum_{i=1}^{n_{r}}\sum_{j=1}^{n_{c}}\mathbb{E}(W^{(i,j)}(W^{(i,j)})')\|,\|\sum_{i=1}^{n_{r}}\sum_{j=1}^{n_{c}}\mathbb{E}((W^{(i,j)})'W^{(i,j)})\|).
\end{align*}
Since
\begin{align*}
\mathbb{E}(W^{2}(i,j))=\mathbb{E}((A(i,j)-\Omega(i,j))^{2})=\mathrm{Var}(A(i,j)),
\end{align*}
where $\mathrm{Var}(A(i,j))$ denotes the variance of Bernoulli random variable $A(i,j)$, we have
\begin{align*}
\mathbb{E}(W^{2}(i,j))&=\mathrm{Var}(A(i,j))=\mathbb{P}(A(i,j)=1)(1-\mathbb{P}(A(i,j)=1))\\
&\leq\mathbb{P}(A(i,j)=1)=\Omega(i,j)=e'_{i}\Theta_{r}\Pi_{r}P\Pi'_{c}\tilde{e}_{j}=\theta_{r}(i)e'_{i}\Pi_{r}P\Pi'_{c}\tilde{e}_{j}\leq \theta_{r}(i)P_{\mathrm{max}}.
\end{align*}
Since $e_{i}e'_{i}$ is an $n_{r}\times n_{r}$ diagonal matrix with $(i,i)$-th entry being 1 and others entries being 0, we have
\begin{align*}
\|\sum_{i=1}^{n_{r}}\sum_{j=1}^{n_{c}}\mathbb{E}(W^{(i,j)}(W^{(i,j)})')\|=\|\sum_{i=1}^{n_{r}}\sum_{j=1}^{n_{c}}\mathbb{E}(W^{2}(i,j))e_{i}e'_{i}\|=\underset{1\leq i\leq n_{r}}{\mathrm{max}}|\sum_{j=1}^{n_{c}}\mathbb{E}(W^{2}(i,j))|\leq \theta_{r,\mathrm{max}}P_{\mathrm{max}}n_{c}.
\end{align*}
Similarly, we have $\|\sum_{i=1}^{n_{r}}\sum_{j=1}^{n_{c}}\mathbb{E}((W^{(i,j)})'W^{(i,j)})\|\leq P_{\mathrm{max}}\|\theta_{r}\|_{1}$, which gives that
\begin{align*}
\sigma^{2}=\mathrm{max}(\|\sum_{i=1}^{n_{r}}\sum_{j=1}^{n_{c}}\mathbb{E}(W^{(i,j)}(W^{(i,j)})')\|,\|\sum_{i=1}^{n_{r}}\sum_{j=1}^{n_{c}}\mathbb{E}((W^{(i,j)})'W^{(i,j)})\|)\leq
P_{\mathrm{max}}\mathrm{max}(\|\theta_{r}\|_{1},\theta_{r,\mathrm{max}}n_{c}).
\end{align*}
By  the rectangular version of Bernstein inequality \cite{tropp2012user}, combining with \\
$\sigma^{2}\leq P_{\mathrm{max}}\mathrm{max}(\|\theta_{r}\|_{1},\theta_{r,\mathrm{max}}n_{c}), R=1, d_{1}+d_{2}=n_{r}+n_{c}$, set\\
$t=\frac{\alpha+1+\sqrt{\alpha^{2}+20\alpha+19}}{3}\sqrt{P_{\mathrm{max}}\mathrm{max}(\|\theta_{r}\|_{1},\theta_{r,\mathrm{max}}n_{c})\mathrm{log}(n_{r}+n_{c})}$ for any $\alpha>0$, we have
\begin{align*}
\mathbb{P}(\|W\|\geq t)&=\mathbb{P}(\|\sum_{i=1}^{n^{r}}\sum_{j=1}^{n_{c}}W^{(i,j)}\|\geq t)\leq(n_{r}+n_{c})\mathrm{exp}(-\frac{t^{2}/2}{\sigma^{2}+\frac{Rt}{3}})\\
&\leq (n_{r}+n_{c})\mathrm{exp}(-\frac{t^{2}/2}{P_{\mathrm{max}}\mathrm{max}(\|\theta_{r}\|_{1},\theta_{r,\mathrm{max}}n_{c})+t/3})\\
&=(n_{r}+n_{c})\mathrm{exp}(-(\alpha+1)\mathrm{log}(n_{r}+n_{c})\cdot \frac{1}{\frac{2(\alpha+1)P_{\mathrm{max}}\mathrm{max}(\|\theta_{r}\|_{1},\theta_{r,\mathrm{max}}n_{c})\mathrm{log}(n_{r}+n_{c})}{t^{2}}+\frac{2(\alpha+1)}{3}\frac{\mathrm{log}(n_{r}+n_{c})}{t}})\\
&=(n_{r}+n_{c})\mathrm{exp}(-(\alpha+1)\mathrm{log}(n_{r}+n_{c})\cdot \frac{1}{\frac{18}{(\sqrt{\alpha+19}+\sqrt{\alpha+1})^{2}}+\frac{2\sqrt{\alpha+1}}{\sqrt{\alpha+19}+\sqrt{\alpha+1}}\sqrt{\frac{\mathrm{log}(n_{r}+n_{c})}{P_{\mathrm{max}}\mathrm{max}(\|\theta_{r}\|_{1},\theta_{r,\mathrm{max}}n_{c})}}})\\
&\leq (n_{r}+n_{c})\mathrm{exp}(-(\alpha+1)\mathrm{log}(n_{r}+n_{c}))=\frac{1}{(n_{r}+n_{c})^{\alpha}},
\end{align*}
where we have used the Assumption 4.1 in the last inequality. Set $\alpha=3$, the claim follows.
\begin{rem}
Consider a special case when $n_{r}=n_{c}=n$ and all entries of $\theta_{r}$ are identical, the upper bound of $\|A-\Omega\|$ is consistent with Corollary 6.5 in \cite{cai2015robust}.
\end{rem}
\end{proof}
\subsection{Proof of Lemma 4.3}
\begin{proof}
Let $H_{\hat{U}}=\hat{U}'U$, and $H_{\hat{U}}=U_{H_{\hat{U}}}\Sigma_{H_{\hat{U}}}V'_{H_{\hat{U}}}$ be the SVD decomposition of $H_{\hat{U}}$ with $U_{H_{\hat{U}}},V_{H_{\hat{U}}}\in \mathbb{R}^{n_{r}\times K}$, where $U_{H_{\hat{U}}}$ and $V_{H_{\hat{U}}}$ represent respectively the left and right singular matrices of $H_{\hat{U}}$. Define $\mathrm{sgn}(H_{\hat{U}})=U_{H_{\hat{U}}}V'_{H_{\hat{U}}}$. $\mathrm{sgn}(H_{\hat{V}})$ is defined similarly. Since $\mathbb{E}(A(i,j)-\Omega(i,j))=0$, $\mathbb{E}[(A(i,j)-\Omega(i,j))^{2}]\leq \theta_{r}(i)P_{\mathrm{max}}\leq \theta_{r,\mathrm{max}}P_{\mathrm{max}}$ by the proof of Lemma 4.2, $\frac{1}{\sqrt{\theta_{r,\mathrm{max}}P_{\mathrm{max}}\mathrm{min}(n_{r},n_{c})/(\mu \mathrm{log}(n_{r}+n_{c}))}}\leq O(1)$ holds by Assumption 4.1 where $\mu$ is the incoherence parameter defined as $\mu=\mathrm{max}(\frac{n_{r}\|U\|^{2}_{2\rightarrow\infty}}{K},\frac{n_{c}\|V\|^{2}_{2\rightarrow\infty}}{K})$. By Theorem 4.3.1. \cite{chen2020spectral}, with high probability, we have below row-wise singular vector deviation
\begin{align}\label{errorhatUhatV}
\mathrm{max}(\|\hat{U}\mathrm{sgn}(H_{\hat{U}})-U\|_{2\rightarrow\infty},\|\hat{V}\mathrm{sgn}(H_{\hat{V}})-V\|_{2\rightarrow\infty})&\leq C\frac{\sqrt{\theta_{r,\mathrm{max}}K}(\kappa(\Omega)\sqrt{\frac{\mathrm{max}(n_{r},n_{c})\mu}{\mathrm{min}(n_{r},n_{c})}}+\sqrt{\mathrm{log}(n_{r}+n_{c})})}{\sigma_{K}(\Omega)}\notag\\
&\leq C\frac{\sqrt{\theta_{r,\mathrm{max}}K\mathrm{log}(n_{r}+n_{c})}}{\sigma_{K}(\Omega)},
\end{align}
provided that $c_{1}\sigma_{K}(\Omega)\geq \sqrt{\theta_{r,\mathrm{max}}(n_{r}+n_{c})\mathrm{log}(n_{r}+n_{c})}$ for some sufficiently small constant $c_{1}$, and here we set $\sqrt{\frac{\mathrm{max}(n_{r},n_{c})\mu}{\mathrm{min}(n_{r},n_{c})}}=O(1)$ for convenience since this term has little effect on the error bounds of DiMSC, especially for the case when $\frac{n_{r}}{n_{c}}=O(1)$.

Since $U'U=I,\hat{U}'\hat{U}=I$, we have   $\|\hat{U}\hat{U}'-UU'\|_{2\rightarrow\infty}\leq 2\|U-\hat{U}\mathrm{sgn}(H_{\hat{U}})\|_{2\rightarrow\infty}$ by basic algebra. Now we are ready to bound $\|\hat{U}\hat{U}'-UU'\|_{2\rightarrow\infty}$:
\begin{align*}	
&\|\hat{U}\hat{U}'-UU'\|_{2\rightarrow\infty}=\mathrm{max}_{1\leq i\leq n_{r}}\|e'_{i}(UU'-\hat{U}\hat{U}')\|_{F}\leq2\|U-\hat{U}\mathrm{sgn}(H_{\hat{U}})\|_{2\rightarrow\infty}\\
&\leq C\frac{\sqrt{\theta_{r,\mathrm{max}}K\mathrm{log}(n_{r}+n_{c})}}{\sigma_{K}(\Omega)}\overset{\mathrm{By~Lemma~}\ref{P4}}{\leq}C\frac{\sqrt{\theta_{r,\mathrm{max}}K\mathrm{log}(n_{r}+n_{c})}}{\theta_{r,\mathrm{min}}\sigma_{K}(P)\sigma_{K}(\Pi_{r})\sigma_{K}(\Pi_{c})}.
\end{align*}
The lemma holds by following similar proof for $\|\hat{V}\hat{V}'-VV'\|_{2\rightarrow\infty}$.
\end{proof}
\subsection{Proof of Lemma 4.4}
\begin{proof}
First, we consider column nodes.  We write down the SP algorithm as below.
	\begin{algorithm}
		\caption{\textbf{Successive Projection (SP)} \citep{gillis2015semidefinite}}
		\label{alg:SP}
		\begin{algorithmic}[1]
			\Require Near-separable matrix $Y_{sp}=S_{sp}M_{sp}+Z_{sp}\in\mathbb{R}^{m\times n}_{+}$ , where $S_{sp}, M_{sp}$ should satisfy Assumption 1 \cite{gillis2015semidefinite}, the number $r$ of columns to be extracted.
			\Ensure Set of indices $\mathcal{K}$ such that $Y(\mathcal{K},:)\approx S$ (up to permutation)
			\State Let $R=Y_{sp}, \mathcal{K}=\{\}, k=1$.
			\State \textbf{While} $R\neq 0$ and $k\leq r$ \textbf{do}
			\State ~~~~~~~$k_{*}=\mathrm{argmax}_{k}\|R(k,:)\|_{F}$.
			\State ~~~~~~$u_{k}=R(k_{*},:)$.
			\State ~~~~~~$R\leftarrow (I-\frac{u_{k}u'_{k}}{\|u_{k}\|^{2}_{F}})R$.
			\State ~~~~~~$\mathcal{K}=\mathcal{K}\cup \{k_{*}\}$.
			\State ~~~~~~k=k+1.
			\State \textbf{end while}
		\end{algorithmic}
	\end{algorithm}
Based on Algorithm \ref{alg:SP}, the following theorem is Theorem 1.1 in \cite{gillis2015semidefinite}.
\begin{thm}\label{gillis2015siamSP}
Fix $m\geq r$ and $n\geq r$. Consider a matrix $Y_{sp}=S_{sp}M_{sp}+Z_{sp}$, where $S_{sp}\in\mathbb{R}^{m\times r}$ has a full column rank, $M_{sp}\in \mathbb{R}^{r\times n}$ is a nonnegative matrix such that the sum of each column is at most 1, and $Z_{sp}=[Z_{sp,1},\ldots, Z_{sp,n}]\in \mathbb{R}^{m\times n}$. Suppose $M_{sp}$ has a submatrix equal to $I_{r}$. Write $\epsilon\leq \mathrm{max}_{1\leq i\leq n}\|Z_{sp,i}\|_{F}$. Suppose $\epsilon=O(\frac{\sigma_{\mathrm{min}}(S_{sp})}{\sqrt{r}\kappa^{2}(S_{sp})})$, where $\sigma_{\mathrm{min}}(S_{sp})$ and $\kappa(S_{sp})$ are the minimum singular value and condition number of $S_{sp}$, respectively. If we apply the SP algorithm to columns of $Y_{sp}$, then it outputs an index set $\mathcal{K}\subset \{1,2,\ldots, n\}$ such that $|\mathcal{K}|=r$ and $\mathrm{max}_{1\leq k\leq r}\mathrm{min}_{j\in\mathcal{K}}\|S_{sp}(:,k)-Y_{sp}(:,j)\|_{F}=O(\epsilon \kappa^{2}(S_{sp}))$, where $S_{sp}(:,k)$ is the $k$-th column of $S_{sp}$.
\end{thm}
Let $m=K, r=K, n=n_{c}, Y_{sp}=\hat{V}'_{2}, Z_{sp}=\hat{V}'_{2}-V'_{2}, S_{sp}=V'_{2}(\mathcal{I}_{c},:),$ and $M_{sp}=\Pi_{c}'$. By condition (I2), $M_{sp}$ has an identity submatrix $I_{K}$. By Lemma 4.3, we have
\begin{align*}
\epsilon_{c}=\mathrm{max}_{1\leq j\leq n_{c}}\|\hat{V}_{2}(j,:)-V_{2}(j,:)\|_{F}=\|\hat{V}_{2}(j,:)-V_{2}(j,:)\|_{2\rightarrow\infty}\leq \varpi.
\end{align*}
By Theorem \ref{gillis2015siamSP}, there exists a permutation matrix $\mathcal{P}_{c}$ such that
\begin{align*} \mathrm{max}_{1\leq k\leq K}\|e'_{k}(\hat{V}_{2}(\mathcal{\hat{I}}_{c},:)-\mathcal{P}'_{c}V_{2}(\mathcal{I}_{c},:))\|_{F}=O(\epsilon_{c}\kappa^{2}(V_{2}(\mathcal{I}_{c},:))\sqrt{K})=O(\varpi\kappa^{2}(V_{2}(\mathcal{I}_{c},:))).
\end{align*}
Since $\kappa^{2}(V_{2}(\mathcal{I}_{c},:))=\kappa(V_{2}(\mathcal{I}_{c},:)V'_{2}(\mathcal{I}_{c},:))=\kappa(V(\mathcal{I}_{c},:)V'(\mathcal{I}_{c},:))=\kappa(\Pi'_{c}\Pi_{c})$ where the last equality holds by Lemma \ref{P3}, we have
\begin{align*}
\mathrm{max}_{1\leq k\leq K}\|e'_{k}(\hat{V}_{2}(\mathcal{\hat{I}}_{c},:)-\mathcal{P}'_{c}V_{2}(\mathcal{I}_{c},:))\|_{F}=O(\varpi\kappa(\Pi'_{c}\Pi_{c})).
\end{align*}
\begin{rem}\label{inputVinSP}
For the ideal case, let $m=K, r=K, n=n_{c}, Y_{sp}=V', Z_{sp}=V'-V'\equiv0, S_{sp}=V'(\mathcal{I}_{c},:),$ and $M_{sp}=\Pi_{c}'$. Then, we have $\mathrm{max}_{1\leq j\leq n_{c}}\|V(j,:)-V(j,:)\|_{F}=0$. By Theorem \ref{gillis2015siamSP}, SP algorithm returns $\mathcal{I}_{c}$ when the input is $V$ assuming there are $K$ column communities.
\end{rem}
Now, we consider row nodes. From Lemma 3.2, we see that $U_{*}(\mathcal{I}_{r},:)$ satisfies condition 1 in \cite{MaoSVM}. Meanwhile, since $(U_{*}(\mathcal{I}_{r},:)U'_{*}(\mathcal{I}_{r},:))^{-1}\mathbf{1}>0$, we have $(U_{*}(\mathcal{I}_{r},:)U'_{*}(\mathcal{I}_{r},:))^{-1}\mathbf{1}\geq \eta\mathbf{1}$, hence $U_{*}(\mathcal{I}_{r},:)$ satisfies condition 2 in \cite{MaoSVM}. Now, we give a lower bound for $\eta$ to show that $\eta$ is strictly positive. By the proof of Lemma \ref{P3}, we have
\begin{align*}
(U_{*}(\mathcal{I}_{r},:)U'_{*}(\mathcal{I}_{r},:))^{-1}&=N_{U}^{-1}(\mathcal{I}_{r},\mathcal{I}_{r})\Theta^{-1}(\mathcal{I}_{r},\mathcal{I}_{r})\Pi'_{r}\Theta^{2}_{r}\Pi_{r}\Theta^{-1}_{r}(\mathcal{I}_{r},\mathcal{I}_{r})N_{U}^{-1}(\mathcal{I}_{r},\mathcal{I}_{r})\\
&\geq \frac{\theta^{2}_{r,\mathrm{min}}}{\theta^{2}_{r,\mathrm{max}}N^{2}_{U,\mathrm{max}}}\Pi'_{r}\Pi_{r}\geq \frac{\theta^{4}_{r,\mathrm{min}}}{\theta^{4}_{r,\mathrm{max}}K\lambda_{1}(\Pi'_{r}\Pi_{r})}\Pi'_{r}\Pi_{r},
\end{align*}
where we set $N_{U,\mathrm{max}}=\mathrm{max}_{1\leq i\leq n_{r}}N_{U}(i,i)$ and we have used the facts that $N_{U}, \Theta_{r}$ are diagonal matrices, and $N_{U,\mathrm{max}}\leq \frac{\theta_{r,\mathrm{max}}\sqrt{K\lambda_{1}(\Pi'_{r}\Pi_{r})}}{\theta_{r,\mathrm{min}}}$ by Lemma \ref{P2}. Then we have
\begin{align*}
&\eta=\mathrm{min}_{1\leq k\leq K}((U_{*}(\mathcal{I}_{r},:)U'_{*}(\mathcal{I}_{r},:))^{-1}\mathbf{1})(k)\geq \frac{\theta^{4}_{r,\mathrm{min}}}{\theta^{4}_{r,\mathrm{max}}K\lambda_{1}(\Pi'_{r}\Pi_{r})}\mathrm{min}_{1\leq k\leq K}e'_{k}\Pi'_{r}\Pi_{r}\mathbf{1}\\
&=\frac{\theta^{4}_{r,\mathrm{min}}}{\theta^{4}_{r,\mathrm{max}}K\lambda_{1}(\Pi'_{r}\Pi_{r})}\mathrm{min}_{1\leq k\leq K}e'_{k}\Pi'_{r}\mathbf{1}=\frac{\theta^{4}_{r,\mathrm{min}}\pi_{r,\mathrm{min}}}{\theta^{4}_{r,\mathrm{max}}K\lambda_{1}(\Pi'_{r}\Pi_{r})},
\end{align*}
i.e., $\eta$ is strictly positive. Since $U_{*,2}(\mathcal{I}_{r},:)U'_{*,2}(\mathcal{I}_{r},:)\equiv U_{*}(\mathcal{I}_{r},:)U'_{*}(\mathcal{I}_{r},:)$, we have $U_{*,2}(\mathcal{I}_{r},:)$ also satisfies conditions 1 and 2 in \cite{MaoSVM}.  The above analysis shows that we can directly apply Lemma F.1 of \cite{MaoSVM} since the Ideal DiMSC algorithm satisfies conditions 1 and 2 in \cite{MaoSVM}, therefore there exists a permutation matrix $\mathcal{P}_{r}\in\mathbb{R}^{K\times K}$ such that
\begin{align*}		
\mathrm{max}_{1\leq k\leq K}\|e'_{k}(\hat{U}_{*,2}(\mathcal{\hat{I}}_{r},:)-\mathcal{P}'_{r}U_{*,2}(\mathcal{I}_{r},:))\|_{F}= O(\frac{\sqrt{K}\zeta\epsilon_{r}}{\lambda^{1.5}_{K}(U_{*,2}(\mathcal{I}_{r},:))U'_{*,2}(\mathcal{I}_{r},:)}),
\end{align*}
where $\zeta\leq\frac{4K}{\eta\lambda^{1.5}_{K}(U_{*,2}(\mathcal{I}_{r},:)U'_{*,2}(\mathcal{I}_{r},:))}=O(\frac{K}{\eta\lambda^{1.5}_{K}(U_{*}(\mathcal{I}_{r},:)U'_{*}(\mathcal{I}_{r},:))})$, and $\epsilon_{r}=\mathrm{max}_{1\leq i\leq n_{r}}\|\hat{U}_{*,2}(i,:)-U_{*,2}(i,:)\|$. Next we bound $\epsilon_{r}$ as below
\begin{align*}		&\|\hat{U}_{*,2}(i,:)-U_{*,2}(i,:)\|_{F}=\|\frac{\hat{U}_{2}(i,:)\|U_{2}(i,:)\|_{F}-U_{2}(i,:)\|\hat{U}_{2}(i,:)\|_{F}}{\|\hat{U}_{2}(i,:)\|_{F}\|U_{2}(i,:)\|_{F}}\|_{F}\leq\frac{2\|\hat{U}_{2}(i,:)-U_{2}(i,:)\|_{F}}{\|U_{2}(i,:)\|_{F}}\\
&\leq \frac{2\|\hat{U}_{2}-U_{2}\|_{2\rightarrow\infty}}{\|U_{2}(i,:)\|_{F}}\leq\frac{2\varpi}{\|U_{2}(i,:)\|_{F}}=\frac{2\varpi}{\|(UU')(i,:)\|_{F}}=\frac{2\varpi}{\|U(i,:)U'\|_{F}}=\frac{2\varpi}{\|U(i,:)\|_{F}}\\
&\leq 2\varpi\frac{\theta_{r,\mathrm{max}}\sqrt{K\lambda_{1}(\Pi'_{r}\Pi_{r})}}{\theta_{r,\mathrm{min}}},
\end{align*}
where the last inequality holds by Lemma \ref{P2}. Then, we have $\epsilon_{r}=O(\varpi\frac{\theta_{r,\mathrm{max}}\sqrt{K\lambda_{1}(\Pi'_{r}\Pi_{r})}}{\theta_{r,\mathrm{min}}})$. Finally, by Lemma \ref{P3}, we have
\begin{align*} \mathrm{max}_{1\leq k\leq K}\|e'_{k}(\hat{U}_{*,2}(\mathcal{\hat{I}}_{r},:)-\mathcal{P}'_{r}U_{*,2}(\mathcal{I}_{r},:))\|_{F}=O(\frac{K^{3}\theta^{11}_{r,\mathrm{max}}\varpi\kappa^{3}(\Pi'_{r}\Pi_{r})\lambda^{1.5}_{1}(\Pi'_{r}\Pi_{r})}{\theta^{11}_{r,\mathrm{min}}\pi_{r,\mathrm{min}}}).
\end{align*}
\begin{rem}\label{inputUstarinSVMCone}
For the ideal case, when setting $U_{*}$ as the input of SVM-cone algorithm assuming there are $K$ row communities, since $\|U_{*}-U_{*}\|_{2\rightarrow\infty}=0$, Lemma F.1. \cite{MaoSVM} guarantees that SVM-cone algorithm returns $\mathcal{I}_{r}$ exactly. Meanwhile, another view to see that SVM-cone algorithm exactly obtains $\mathcal{I}_{r}$ when the input is $U_{*}$ (also $U_{2,*}$) is given in Appendix \ref{introduceSVMcone}, which focuses on following the three steps of SVM-cone algorithm to show that it returns $\mathcal{I}_{r}$ with input $U_{*}$ (also $U_{*,2}$), instead of simply applying Lemma F.1. \cite{MaoSVM}.
\end{rem}
\end{proof}
\subsection{Proof of Lemma 4.5}
\begin{proof}
First, we consider column nodes. Recall that $V(\mathcal{I}_{c},:)=B_{c}$. For convenience, set $\hat{V}(\mathcal{\hat{I}}_{c},:)=\hat{B}_{c}, V_{2}(\mathcal{I}_{c},:)=B_{2c}, \hat{V}_{2}(\mathcal{\hat{I}}_{c},:)=\hat{B}_{2c}$. We bound $\|e'_{j}(\hat{Z}_{c}-Z_{c}\mathcal{P}_{c})\|_{F}$ when the input is $\hat{V}$ in the SP algorithm.  Recall that $Z_{c}=\mathrm{max}(VV'(\mathcal{I}_{c},:)(V(\mathcal{I}_{c},:)V'(\mathcal{I}_{c},:))^{-1},0)\equiv \Pi_{c}$,  for $1\leq j\leq n_{c}$, we have
\begin{align*} &\|e'_{j}(\hat{Z}_{c}-Z_{c}\mathcal{P}_{c})\|_{F}=\|e'_{j}(\mathrm{max}(0,\hat{V}\hat{B}'_{c}(\hat{B}_{c}\hat{B}'_{c})^{-1})-VB'_{c}(B_{c}B'_{c})^{-1}\mathcal{P}_{c})\|_{F}\\
&\leq\|e'_{j}(\hat{V}\hat{B}'_{c}(\hat{B}_{c}\hat{B}'_{c})^{-1}-VB'_{c}(B_{c}B'_{c})^{-1}\mathcal{P}_{c})\|_{F}\\ &=\|e'_{j}(\hat{V}-V(V'\hat{V}))\hat{B}'_{c}(\hat{B}_{c}\hat{B}'_{c})^{-1}+e'_{j}(V(V'\hat{V})\hat{B}'_{c}(\hat{B}_{c}\hat{B}'_{c})^{-1}-V(V'\hat{V})(\mathcal{P}'_{c}(B_{c}B'_{c})(B'_{c})^{-1}(V'\hat{V}))^{-1})\|_{F}\\ &\leq\|e'_{j}(\hat{V}-V(V'\hat{V}))\hat{B}'_{c}(\hat{B}_{c}\hat{B}'_{c})^{-1}\|_{F}+\|e'_{j}V(V'\hat{V})(\hat{B}'_{c}(\hat{B}_{c}\hat{B}'_{c})^{-1}-(\mathcal{P}'_{c}(B_{c}B'_{c})(B'_{c})^{-1}(V'\hat{V}))^{-1})\|_{F}\\
&\leq\|e'_{j}(\hat{V}-V(V'\hat{V}))\|_{F}\|\hat{B}^{-1}_{c}\|_{F}+\|e'_{j}V(V'\hat{V})(\hat{B}'_{c}(\hat{B}_{c}\hat{B}'_{c})^{-1}-(\mathcal{P}'_{c}(B_{c}B'_{c})(B'_{c})^{-1}(V'\hat{V}))^{-1})\|_{F}\\	&\leq \sqrt{K}\|e'_{j}(\hat{V}-V(V'\hat{V}))\|_{F}/\sqrt{\lambda_{K}(\hat{B}_{c}\hat{B}'_{c})}+\|e'_{j}V(V'\hat{V})(\hat{B}^{-1}_{c}-(\mathcal{P}_{c}'B_{c}(V'\hat{V}))^{-1})\|_{F}\\
&=\sqrt{K}\|e'_{j}(\hat{V}\hat{V}'-VV')\hat{V}\|_{F}O(\sqrt{\lambda_{1}(\Pi'_{c}\Pi_{c})})+\|e'_{j}V(V'\hat{V})(\hat{B}^{-1}_{c}-(\mathcal{P}_{c}'B_{c}(V'\hat{V}))^{-1})\|_{F}\\
&\leq \sqrt{K}\|e'_{j}(\hat{V}\hat{V}'-VV')\|_{F}O(\sqrt{\lambda_{1}(\Pi'_{c}\Pi_{c})})+\|e'_{j}V(V'\hat{V})(\hat{B}^{-1}_{c}-(\mathcal{P}'_{c}B_{c}(V'\hat{V}))^{-1})\|_{F}\\
&\leq \sqrt{K}\varpi O(\sqrt{\lambda_{1}(\Pi'_{c}\Pi_{c})})+\|e'_{j}V(V'\hat{V})(\hat{B}^{-1}_{c}-(\mathcal{P}'_{c}B_{c}(V'\hat{V}))^{-1})\|_{F}\\	&=O(\varpi\sqrt{K\lambda_{1}(\Pi'_{c}\Pi_{c})})+\|e'_{j}V(V'\hat{V})(\hat{B}^{-1}_{c}-(\mathcal{P}'_{c}B_{c}(V'\hat{V}))^{-1})\|_{F},
\end{align*}
where we have used similar idea in the proof of Lemma VII.3 in \cite{mao2020estimating} such that apply $O(\frac{1}{\lambda_{K}(B_{c}B'_{c})})$ to estimate $\frac{1}{\lambda_{K}(\hat{B}_{c}\hat{B}'_{c})}$, then by Lemma \ref{P3}, we have $\frac{1}{\lambda_{K}(\hat{B}_{c}\hat{B}'_{c})}=O(\lambda_{1}(\Pi'_{c}\Pi_{c}))$.

Now we aim to bound $\|e'_{j}V(V'\hat{V})(\hat{B}^{-1}_{c}-(\mathcal{P}_{c}'B_{c}(V'\hat{V}))^{-1})\|_{F}$. For convenience, set $T_{c}=V'\hat{V}, S_{c}=\mathcal{P}_{c}'B_{c}T_{c}$. We have
\begin{align}	&\|e'_{j}V(V'\hat{V})(\hat{B}^{-1}_{c}-(\mathcal{P}'_{c}B_{c}(V'\hat{V}))^{-1})\|_{F}=\|e'_{j}VT_{c}S^{-1}_{c}(S_{c}-\hat{B}_{c})\hat{B}^{-1}_{c}\|_{F}\notag\\	&\leq\|e'_{j}VT_{c}S^{-1}_{c}(S_{c}-\hat{B}_{c})\|_{F}\|\hat{B}^{-1}_{c}\|_{F}\leq\|e'_{j}VT_{c}S^{-1}_{c}(S_{c}-\hat{B}_{c})\|_{F}\frac{\sqrt{K}}{|\lambda_{K}(\hat{B}_{c})|}\notag\\	&=\|e'_{j}VT_{c}S^{-1}_{c}(S_{c}-\hat{B}_{c})\|_{F}\frac{\sqrt{K}}{\sqrt{\lambda_{K}(\hat{B}_{c}\hat{B}'_{c})}}\leq\|e'_{j}VT_{c}S^{-1}_{c}(S_{c}-\hat{B}_{c})\|_{F}O(\sqrt{K\lambda_{1}(\Pi'_{c}\Pi_{c})})\notag\\	&=\|e'_{j}VT_{c}T^{-1}_{c}B'_{c}(B_{c}B'_{c})^{-1}\mathcal{P}_{c}(S_{c}-\hat{B}_{c})\|_{F}O(\sqrt{K\lambda_{1}(\Pi'_{c}\Pi_{c})})\notag\\	&=\|e'_{j}VB'_{c}(B_{c}B'_{c})^{-1}\mathcal{P}_{c}(S_{c}-\hat{B}_{c})\|_{F}O(\sqrt{K\lambda_{1}(\Pi'_{c}\Pi_{c})})\notag\\ &=\|e'_{j}Z_{c}\mathcal{P}_{c}(S_{c}-\hat{B}_{c})\|_{F}O(\sqrt{K\lambda_{1}(\Pi'_{c}\Pi_{c})})\overset{\mathrm{By~}Z_{c}=\Pi_{c}}{\leq}\mathrm{max}_{1\leq k\leq K}\|e'_{k}(S_{c}-\hat{B}_{c})\|_{F}O(\sqrt{K\lambda_{1}(\Pi'_{c}\Pi_{c})})\notag\\ &=\mathrm{max}_{1\leq k\leq K}\|e'_{k}(\hat{B}_{c}-\mathcal{P}_{c}'B_{c}V'\hat{V})\|_{F}O(\sqrt{K\lambda_{1}(\Pi'_{c}\Pi_{c})})\notag\\
&=\mathrm{max}_{1\leq k\leq K}\|e'_{k}(\hat{B}_{c}\hat{V}'-\mathcal{P}'_{c}B_{c}V')\hat{V}\|_{F}O(\sqrt{K\lambda_{1}(\Pi'_{c}\Pi_{c})})\notag\\ &\leq\mathrm{max}_{1\leq k\leq K}\|e'_{k}(\hat{B}_{c}\hat{V}'-\mathcal{P}_{c}'B_{c}V')\|_{F}O(\sqrt{K\lambda_{1}(\Pi'_{c}\Pi_{c})})\notag\\
&=\mathrm{max}_{1\leq k\leq K}\|e'_{k}(\hat{B}_{2c}-\mathcal{P}_{c}'B_{2c})\|_{F}O(\sqrt{K\lambda_{1}(\Pi'_{c}\Pi_{c})})\label{BenefitV}\\
&=O(\varpi\kappa(\Pi'_{c}\Pi_{c})\sqrt{K\lambda_{1}(\Pi'_{c}\Pi_{c})})\notag.
\end{align}
\begin{rem}\label{BenefitEquivalence}
Eq (\ref{BenefitV}) supports our statement that building the theoretical framework of DiMSC benefits a lot by introducing DiMSC-equivalence algorithm since $\|\hat{B}_{2c}-\mathcal{P}'_{c}B_{2c}\|_{2\rightarrow\infty}$ is obtained from DiMSC-equivalence (i.e., inputing $\hat{V}_{2}$ in the SP algorithm obtains $\|\hat{B}_{2c}-\mathcal{P}'_{c}B_{2c}\|_{2\rightarrow\infty}$).
\end{rem}
Then, we have
\begin{align*}
\|e'_{j}(\hat{Z}_{c}-Z_{c}\mathcal{P}_{c})\|_{F}&\leq O(\varpi\sqrt{K\lambda_{1}(\Pi'_{c}\Pi_{c})})+\|e'_{j}V(V'\hat{V})(\hat{B}^{-1}_{c}-(\mathcal{P}'_{c}B_{c}(V'\hat{V}))^{-1})\|_{F}\\
&\leq O(\varpi\sqrt{K\lambda_{1}(\Pi'_{c}\Pi_{c})})+O(\varpi\kappa(\Pi'_{c}\Pi_{c})\sqrt{K\lambda_{1}(\Pi'_{c}\Pi_{c})})\\
&=O(\varpi\kappa(\Pi'_{c}\Pi_{c})\sqrt{K\lambda_{1}(\Pi'_{c}\Pi_{c})}).
\end{align*}
Next, we consider row nodes. For $1\leq i\leq n_{r}$, since $Z_{r}=Y_{*}J_{*}, \hat{Z}_{r}=\hat{Y}_{*}\hat{J}_{*}$, we have
\begin{align*} &\|e'_{i}(\hat{Z}_{r}-Z_{r}\mathcal{P}_{r})\|_{F}=\|e'_{i}(\mathrm{max}(0, \hat{Y}_{*}\hat{J}_{*})-Y_{*}J_{*}\mathcal{P}_{r})\|_{F}\leq \|e'_{i}(\hat{Y}_{*}\hat{J}_{*}-Y_{*}J_{*}\mathcal{P}_{r})\|_{F}\\
&=\|e'_{i}(\hat{Y}_{*}-Y_{*}\mathcal{P}_{r})\hat{J}_{*}+e'_{i}Y_{*}\mathcal{P}_{r}(\hat{J}_{*}-\mathcal{P}'_{r}J_{*}\mathcal{P}_{r})\|_{F}\leq\|e'_{i}(\hat{Y}_{*}-Y_{*}\mathcal{P}_{r})\|_{F}\|\hat{J}_{*}\|_{F}+\|e'_{i}Y_{*}\mathcal{P}_{r}\|_{F}\|\hat{J}_{*}-\mathcal{P}'_{r}J_{*}\mathcal{P}_{r}\|_{F}\\
&=\|e'_{i}(\hat{Y}_{*}-Y_{*}\mathcal{P}_{r})\|_{F}\|\hat{J}_{*}\|_{F}+\|e'_{i}Y_{*}\|_{F}\|\hat{J}_{*}-\mathcal{P}'_{r}J_{*}\mathcal{P}_{r}\|_{F}. \end{align*}
Therefore, the bound of $\|e'_{i}(\hat{Z}_{r}-Z_{r}\mathcal{P}_{r})\|_{F}$ can be obtained as long as we bound $\|e'_{i}(\hat{Y}_{*}-Y_{*}\mathcal{P}_{r})\|_{F}, \|\hat{J}_{*}\|_{F}, \|e'_{i}Y_{*}\|_{F}$ and $\|\hat{J}_{*}-\mathcal{P}'_{r}J_{*}\mathcal{P}_{r}\|_{F}$. We bound the four terms as below:
\begin{itemize}
  \item we bound $\|e'_{i}(\hat{Y}_{*}-Y_{*}\mathcal{P}_{r})\|_{F}$ first. Similar as bounding $\|e'_{j}(\hat{Z}_{c}-Z_{c}\mathcal{P}_{c})\|$, we set $U_{*}(\mathcal{I}_{r},:)=B_{R},\hat{U}_{*}(\mathcal{\hat{I}}_{r},:)=\hat{B}_{R}, U_{*,2}(\mathcal{I}_{r},:)=B_{2R}, \hat{U}_{*,2}(\mathcal{\hat{I}}_{r},:)=\hat{B}_{2R}$ for convenience. We bound $\|e'_{i}(\hat{Y}_{*}-Y_{*}\mathcal{P}_{r})\|_{F}$ when the input is $\hat{U}_{*}$ in the SVM-cone algorithm.  For $1\leq i\leq n_{r}$, we have
\begin{align*} &\|e'_{i}(\hat{Y}_{*}-Y_{*}\mathcal{P}_{r})\|_{F}=\|e'_{i}(\hat{U}\hat{B}'_{R}(\hat{B}_{R}\hat{B}'_{R})^{-1}-UB'_{R}(B_{R}B'_{R})^{-1}\mathcal{P}_{r})\|_{F}\\ &=\|e'_{i}(\hat{U}-U(U'\hat{U}))\hat{B}'_{R}(\hat{B}_{R}\hat{B}'_{R})^{-1}\\
&~~~+e'_{i}(U(U'\hat{U})\hat{B}'_{R}(\hat{B}_{R}\hat{B}'_{R})^{-1}-U(U'\hat{U})(\mathcal{P}'_{r}(B_{R}B'_{R})(B'_{R})^{-1}(U'\hat{U}))^{-1})\|_{F}\\ &\leq\|e'_{i}(\hat{U}-U(U'\hat{U}))\hat{B}'_{R}(\hat{B}_{R}\hat{B}'_{R})^{-1}\|_{F}+\|e'_{i}U(U'\hat{U})(\hat{B}'_{R}(\hat{B}_{R}\hat{B}'_{R})^{-1}-(\mathcal{P}'_{r}(B_{R}B'_{R})(B'_{R})^{-1}(U'\hat{U}))^{-1})\|_{F}\\
&\leq\|e'_{i}(\hat{U}-U(U'\hat{U}))\|_{F}\|\hat{B}^{-1}_{R}\|_{F}+\|e'_{i}U(U'\hat{U})(\hat{B}'_{R}(\hat{B}_{R}\hat{B}'_{R})^{-1}-(\mathcal{P}'_{r}(B_{R}B'_{R})(B'_{R})^{-1}(U'\hat{U}))^{-1})\|_{F}\\	&\leq \sqrt{K}\|e'_{i}(\hat{U}-U(U'\hat{U}))\|_{F}/\sqrt{\lambda_{K}(\hat{B}_{R}\hat{B}'_{R})}+\|e'_{i}U(U'\hat{U})(\hat{B}^{-1}_{R}-(\mathcal{P}_{r}'B_{R}(U'\hat{U}))^{-1})\|_{F}\\
&\overset{(i)}{=}\sqrt{K}\|e'_{i}(\hat{U}\hat{U}'-UU')\hat{U}\|_{F}O(\frac{\theta_{r,\mathrm{max}}\sqrt{\kappa(\Pi'_{r}\Pi_{r})}}{\theta_{r,\mathrm{min}}})+\|e'_{i}U(U'\hat{U})(\hat{B}^{-1}_{R}-(\mathcal{P}_{r}'B_{R}(U'\hat{U}))^{-1})\|_{F}\\
&\leq \sqrt{K}\|e'_{i}(\hat{U}\hat{U}'-UU')\|_{F}O(\frac{\theta_{r,\mathrm{max}}\sqrt{\kappa(\Pi'_{r}\Pi_{r})}}{\theta_{r,\mathrm{min}}})+\|e'_{i}U(U'\hat{U})(\hat{B}^{-1}_{R}-(\mathcal{P}'_{r}B_{R}(U'\hat{U}))^{-1})\|_{F}\\
&\leq \sqrt{K}\varpi O(\frac{\theta_{r,\mathrm{max}}\sqrt{\kappa(\Pi'_{r}\Pi_{r})}}{\theta_{r,\mathrm{min}}})+\|e'_{i}U(U'\hat{U})(\hat{B}^{-1}_{R}-(\mathcal{P}'_{r}B_{R}(U'\hat{U}))^{-1})\|_{F}\\	&=O(\varpi\frac{\theta_{r,\mathrm{max}}\sqrt{K\kappa(\Pi'_{r}\Pi_{r})}}{\theta_{r,\mathrm{min}}})+\|e'_{i}U(U'\hat{U})(\hat{B}^{-1}_{R}-(\mathcal{P}'_{r}B_{R}(U'\hat{U}))^{-1})\|_{F},
\end{align*}
where we have used similar idea in the proof of Lemma VII.3 in \cite{mao2020estimating} such that apply $O(\frac{1}{\lambda_{K}(B_{R}B'_{R})})$ to estimate $\frac{1}{\lambda_{K}(\hat{B}_{R}\hat{B}'_{R})}$, hence (i) holds by Lemma \ref{P3}.

Now we aim to bound $\|e'_{i}U(U'\hat{U})(\hat{B}^{-1}_{R}-(\mathcal{P}_{r}'B_{R}(U'\hat{U}))^{-1})\|_{F}$. For convenience, set $T_{r}=U'\hat{U}, S_{r}=\mathcal{P}_{r}'B_{R}T_{r}$. We have
\begin{align}	&\|e'_{i}U(U'\hat{U})(\hat{B}^{-1}_{R}-(\mathcal{P}'_{r}B_{R}(U'\hat{U}))^{-1})\|_{F}=\|e'_{i}UT_{r}S^{-1}_{r}(S_{r}-\hat{B}_{R})\hat{B}^{-1}_{R}\|_{F}\notag\\	&\leq\|e'_{i}UT_{r}S^{-1}_{r}(S_{r}-\hat{B}_{R})\|_{F}\|\hat{B}^{-1}_{R}\|_{F}\leq\|e'_{i}UT_{r}S^{-1}_{r}(S_{r}-\hat{B}_{R})\|_{F}\frac{\sqrt{K}}{|\lambda_{K}(\hat{B}_{R})|}\notag\\	&=\|e'_{i}UT_{r}S^{-1}_{r}(S_{r}-\hat{B}_{R})\|_{F}\frac{\sqrt{K}}{\sqrt{\lambda_{K}(\hat{B}_{R}\hat{B}'_{R})}}\leq\|e'_{i}UT_{r}S^{-1}_{r}(S_{r}-\hat{B}_{R})\|_{F}O(\frac{\theta_{r,\mathrm{max}}\sqrt{K\kappa(\Pi'_{r}\Pi_{r})}}{\theta_{r,\mathrm{min}}})\notag\\	&=\|e'_{i}UT_{r}T^{-1}_{r}B'_{R}(B_{R}B'_{R})^{-1}\mathcal{P}_{r}(S_{r}-\hat{B}_{R})\|_{F}O(\frac{\theta_{r,\mathrm{max}}\sqrt{K\kappa(\Pi'_{r}\Pi_{r})}}{\theta_{r,\mathrm{min}}})\notag\\	&=\|e'_{i}UB'_{R}(B_{R}B'_{R})^{-1}\mathcal{P}_{r}(S_{r}-\hat{B}_{R})\|_{F}O(\frac{\theta_{r,\mathrm{max}}\sqrt{K\kappa(\Pi'_{r}\Pi_{r})}}{\theta_{r,\mathrm{min}}})\notag\\ &=\|e'_{i}Y_{*}\mathcal{P}_{r}(S_{r}-\hat{B}_{R})\|_{F}O(\frac{\theta_{r,\mathrm{max}}\sqrt{K\lambda_{1}(\Pi'_{r}\Pi_{r})}}{\theta_{r,\mathrm{min}}})\leq \|e'_{i}Y_{*}\|_{F}\|S_{r}-\hat{B}_{R}\|_{F}O(\frac{\theta_{r,\mathrm{max}}\sqrt{K\lambda_{1}(\Pi'_{r}\Pi_{r})}}{\theta_{r,\mathrm{min}}})\notag\\
&\overset{\mathrm{By~Eq}(\ref{boundYstar})}{\leq}\frac{\theta^{2}_{r,\mathrm{max}}\sqrt{K\lambda_{1}(\Pi'_{r}\Pi_{r})}}{\theta^{2}_{r,\mathrm{min}}\lambda_{K}(\Pi'_{r}\Pi_{r})}\mathrm{max}_{1\leq k\leq K}\|e'_{k}(S_{r}-\hat{B}_{R})\|_{F}O(\frac{\theta_{r,\mathrm{max}}K\sqrt{\kappa(\Pi'_{r}\Pi_{r})}}{\theta_{r,\mathrm{min}}})\notag\\ &=\mathrm{max}_{1\leq k\leq K}\|e'_{k}(\hat{B}_{R}-\mathcal{P}_{r}'B_{R}U'\hat{U})\|_{F}O(\frac{\theta^{3}_{r,\mathrm{max}}K^{1.5}\kappa(\Pi'_{r}\Pi_{r})}{\theta^{3}_{r,\mathrm{min}}\sqrt{\lambda_{K}(\Pi'_{r}\Pi_{r})}})\notag\\
&=\mathrm{max}_{1\leq k\leq K}\|e'_{k}(\hat{B}_{R}\hat{U}'-\mathcal{P}'_{r}B_{R}U')\hat{U}\|_{F}O(\frac{\theta^{3}_{r,\mathrm{max}}K^{1.5}\kappa(\Pi'_{r}\Pi_{r})}{\theta^{3}_{r,\mathrm{min}}\sqrt{\lambda_{K}(\Pi'_{r}\Pi_{r})}})\notag\\ &\leq\mathrm{max}_{1\leq k\leq K}\|e'_{k}(\hat{B}_{R}\hat{U}'-\mathcal{P}_{r}'B_{R}U')\|_{F}O(\frac{\theta^{3}_{r,\mathrm{max}}K^{1.5}\kappa(\Pi'_{r}\Pi_{r})}{\theta^{3}_{r,\mathrm{min}}\sqrt{\lambda_{K}(\Pi'_{r}\Pi_{r})}})\notag\\
&=\mathrm{max}_{1\leq k\leq K}\|e'_{k}(\hat{B}_{2R}-\mathcal{P}_{r}'B_{2R})\|_{F}O(\frac{\theta^{3}_{r,\mathrm{max}}K^{1.5}\kappa(\Pi'_{r}\Pi_{r})}{\theta^{3}_{r,\mathrm{min}}\sqrt{\lambda_{K}(\Pi'_{r}\Pi_{r})}})\label{BenefitU}\\
&\overset{\mathrm{By~Lemma~4.4}}{=}O(\frac{K^{4.5}\theta^{14}_{r,\mathrm{max}}\varpi\kappa^{4.5}(\Pi'_{r}\Pi_{r})\lambda_{1}(\Pi'_{r}\Pi_{r})}{\theta^{14}_{r,\mathrm{min}}\pi_{r,\mathrm{min}}})\notag.
\end{align}
\begin{rem}\label{BenefitEquivalenceU}
Similar as Eq (\ref{BenefitV}), Eq (\ref{BenefitU}) supports our statement that building the theoretical framework of DiMSC benefits a lot by introducing DiMSC-equivalence algorithm since $\|\hat{B}_{2R}-\mathcal{P}'_{r}B_{2R}\|_{2\rightarrow\infty}$ is obtained from DiMSC-equivalence (i.e., inputing $\hat{U}_{*,2}$ in the SVM-cone algorithm obtains $\|\hat{B}_{2R}-\mathcal{P}'_{r}B_{2R}\|_{2\rightarrow\infty}$).
\end{rem}
Then, we have
\begin{align*}
\|e'_{i}(\hat{Y}_{*}-Y_{*}\mathcal{P}_{r})\|_{F}&\leq O(\varpi\frac{\theta_{r,\mathrm{max}}\sqrt{K\kappa(\Pi'_{r}\Pi_{r})}}{\theta_{r,\mathrm{min}}})+\|e'_{i}U(U'\hat{U})(\hat{B}^{-1}_{R}-(\mathcal{P}'_{r}B_{R}U'\hat{U}))^{-1})\|_{F}\\
&\leq O(\varpi\frac{\theta_{r,\mathrm{max}}\sqrt{K\kappa(\Pi'_{r}\Pi_{r})}}{\theta_{r,\mathrm{min}}})+O(\frac{K^{4.5}\theta^{14}_{r,\mathrm{max}}\varpi\kappa^{4.5}(\Pi'_{r}\Pi_{r})\lambda_{1}(\Pi'_{r}\Pi_{r})}{\theta^{14}_{r,\mathrm{min}}\pi_{r,\mathrm{min}}})\\
&=O(\frac{K^{4.5}\theta^{14}_{r,\mathrm{max}}\varpi\kappa^{4.5}(\Pi'_{r}\Pi_{r})\lambda_{1}(\Pi'_{r}\Pi_{r})}{\theta^{14}_{r,\mathrm{min}}\pi_{r,\mathrm{min}}}).
\end{align*}
  \item for $\|e'_{i}Y_{*}\|_{F}$, since $Y_{*}=UU^{-1}_{*}(\mathcal{I}_{r},:)$, by Lemmas \ref{P2} and \ref{P3}, we have
  \begin{align}\label{boundYstar}
        \|e'_{i}Y_{*}\|_{F}\leq \|U(i,:)\|_{F}\|U^{-1}_{*}(\mathcal{I}_{r},:)\|_{F}\leq \frac{\sqrt{K}\|U(i,:)\|_{F}}{\sqrt{\lambda_{K}(U_{*}(\mathcal{I}_{r},:)U'_{*}(\mathcal{I}_{r},:))}}\leq \frac{\theta^{2}_{r,\mathrm{max}}\sqrt{K\lambda_{1}(\Pi'_{r}\Pi_{r})}}{\theta^{2}_{r,\mathrm{min}}\lambda_{K}(\Pi'_{r}\Pi_{r})}.
      \end{align}
  \item for $\|\hat{J}_{*}\|_{F}$, recall that $\hat{J}_{*}=\mathrm{diag}(\hat{U}_{*}(\hat{I}_{r},:)\hat{\Lambda}\hat{V}'(\hat{\mathcal{I}}_{c},:))$, we have
      \begin{align*}
      \|\hat{J}_{*}\|&=\mathrm{max}_{1\leq k\leq K}\hat{J}_{*}(k,k)=\mathrm{max}_{1\leq k\leq K}e'_{k}\hat{U}_{*}(\hat{I}_{r},:)\hat{\Lambda}\hat{V}'(\hat{\mathcal{I}}_{c},:)e_{k}=\mathrm{max}_{1\leq k\leq K}\|e'_{k}\hat{U}_{*}(\hat{I}_{r},:)\hat{\Lambda}\hat{V}'(\hat{\mathcal{I}}_{c},:)e_{k}\|\\
      &\leq \mathrm{max}_{1\leq k\leq K}\|e'_{k}\hat{U}_{*}(\hat{I}_{r},:)\|\|\hat{\Lambda}\|\|\hat{V}'(\hat{\mathcal{I}}_{c},:)e_{k}\|\leq \mathrm{max}_{1\leq k\leq K}\|e'_{k}\hat{U}_{*}(\hat{I}_{r},:)\|_{F}\|\hat{\Lambda}\|\|\hat{V}'(\hat{\mathcal{I}}_{c},:)e_{k}\|\\
      &=\mathrm{max}_{1\leq k\leq K}\|A\|\|\hat{V}'(\hat{\mathcal{I}}_{c},:)e_{k}\|=\mathrm{max}_{1\leq k\leq K}\|A\|\|(e'_{k}\hat{V}(\hat{\mathcal{I}}_{c},:))'\|=\mathrm{max}_{1\leq k\leq K}\|A\|\|e'_{k}\hat{V}(\hat{\mathcal{I}}_{c},:)\|\\
      &\leq\mathrm{max}_{1\leq k\leq K}\|A\|\|e'_{k}\hat{V}(\hat{\mathcal{I}}_{c},:)\|_{F}\leq \|A\|\|\hat{V}\|_{2\rightarrow\infty}=\|A\|\|\hat{V}\mathrm{sgn}(H_{\hat{V}})-V+V\|_{2\rightarrow\infty}\\
      &\leq \|A\|(\|\hat{V}\mathrm{sgn}(H_{\hat{V}})-V\|_{2\rightarrow\infty}+\|V\|_{2\rightarrow\infty}).
      \end{align*}
      By Lemma 4.2 and Lemma \ref{P4}, $\|A\|=\|A-\Omega+\Omega\|\leq \|A-\Omega\|+\sigma_{1}(\Omega)\leq\|A-\Omega\|+\theta_{r,\mathrm{max}}\sigma_{1}(P)\sigma_{1}(\Pi_{r})\sigma_{1}(\Pi_{c})=O(\theta_{r,\mathrm{max}}\sigma_{1}(\Pi_{r})\sigma_{1}(\Pi_{c}))$. By Lemma \ref{P4} and Eq (\ref{errorhatUhatV}),
$\|\hat{V}\mathrm{sgn}(H_{\hat{V}})-V\|_{2\rightarrow\infty}\leq
C\frac{\sqrt{\theta_{r,\mathrm{max}}K\mathrm{log}(n_{r}+n_{c})}}{\theta_{r,\mathrm{min}}\sigma_{K}(P)\sigma_{K}(\Pi_{r})\sigma_{K}(\Pi_{c})}$. By Lemma \ref{P2}, $\|V\|_{2\rightarrow\infty}\leq \sqrt{\frac{1}{\lambda_{K}(\Pi'_{c}\Pi_{c})}}$, which gives $\|\hat{V}\mathrm{sgn}(H_{\hat{V}})-V\|_{2\rightarrow\infty}+\|V\|_{2\rightarrow\infty}=O(\sqrt{\frac{1}{\lambda_{K}(\Pi'_{c}\Pi_{c})}})$ (this can be seen as simply using $\|V\|_{2\rightarrow\infty}$ to estimate $\|\hat{V}\|_{2\rightarrow\infty}$ since $\sqrt{\frac{1}{\lambda_{K}(\Pi'_{c}\Pi_{c})}}$ is the same order as $\frac{\sqrt{\theta_{r,\mathrm{max}}K\mathrm{log}(n_{r}+n_{c})}}{\theta_{r,\mathrm{min}}\sigma_{K}(P)\sigma_{K}(\Pi_{r})\sigma_{K}(\Pi_{c})}$). Then we have $\|\hat{J}_{*}\|=O(\theta_{r,\mathrm{max}}\sigma_{1}(\Pi_{r})\kappa(\Pi_{c}))$, which gives that $\|\hat{J}_{*}\|_{F}=O(\theta_{r,\mathrm{max}}\sqrt{K}\sigma_{1}(\Pi_{r})\kappa(\Pi_{c}))$.
\item for $\|\hat{J}_{*}-\mathcal{P}'_{r}J_{*}\mathcal{P}_{r}\|_{F}$, since $J_{*}=N_{U}(\mathcal{I}_{r},\mathcal{I}_{r})\Theta_{r}(\mathcal{I}_{r},\mathcal{I}_{r})$, we have $\|J_{*}\|\leq N_{U,\mathrm{max}}\theta_{r,\mathrm{max}}\leq \frac{\theta^{2}_{r,\mathrm{max}}\sqrt{K\lambda_{1}(\Pi'_{r}\Pi_{r})}}{\theta_{r,\mathrm{min}}}$, which gives that $\|J_{*}\|_{F}\leq \frac{\theta^{2}_{r,\mathrm{max}}K\sigma_{1}(\Pi_{r})}{\theta_{r,\mathrm{min}}}$. Thus, we have $\|\hat{J}_{*}-\mathcal{P}'_{r}J_{*}\mathcal{P}_{r}\|_{F}=O(\frac{\theta^{2}_{r,\mathrm{max}}K\sigma_{1}(\Pi_{r})}{\theta_{r,\mathrm{min}}})$.
\end{itemize}
Combine the above results, we have
\begin{align*} &\|e'_{i}(\hat{Z}_{r}-Z_{r}\mathcal{P}_{r})\|_{F}\leq\|e'_{i}(\hat{Y}_{*}-Y_{*}\mathcal{P}_{r})\|_{F}\|\hat{J}_{*}\|_{F}+\|e'_{i}Y_{*}\|_{F}\|\hat{J}_{*}-\mathcal{P}'_{r}J_{*}\mathcal{P}_{r}\|_{F}\\
&=O(\frac{K^{4.5}\theta^{14}_{r,\mathrm{max}}\varpi\kappa^{4.5}(\Pi'_{r}\Pi_{r})\lambda_{1}(\Pi'_{r}\Pi_{r})}{\theta^{14}_{r,\mathrm{min}}\pi_{r,\mathrm{min}}})O(\theta_{r,\mathrm{max}}\sqrt{K}\sigma_{1}(\Pi_{r})\kappa(\Pi_{c}))\\
&~~~+\frac{\theta^{2}_{r,\mathrm{max}}\sqrt{K\lambda_{1}(\Pi'_{r}\Pi_{r})}}{\theta^{2}_{r,\mathrm{min}}\lambda_{K}(\Pi_{r}\Pi_{r})}O(\frac{\theta^{2}_{r,\mathrm{max}}K\sigma_{1}(\Pi_{r})}{\theta_{r,\mathrm{min}}})=O(\frac{K^{5}\theta^{15}_{r,\mathrm{max}}\varpi\kappa^{4.5}(\Pi'_{r}\Pi_{r})\kappa(\Pi_{c})\lambda^{1.5}_{1}(\Pi'_{r}\Pi_{r})}{\theta^{14}_{r,\mathrm{min}}\pi_{r,\mathrm{min}}}).
\end{align*}
\end{proof}
\subsection{Proof of Theorem 4.6}
\begin{proof}
We bound $\|e'_{j}(\hat{\Pi}_{c}-\Pi_{c}\mathcal{P}_{c})\|_{1}$ first. Recall that $Z_{c}=\Pi_{c},\Pi_{c}(j,:)=\frac{Z_{c}(j,:)}{\|Z_{c}(j,:)\|_{1}}, \hat{\Pi}_{c}(i,:)=\frac{\hat{Z}_{c}(j,:)}{\|\hat{Z}_{c}(j,:)\|_{1}}$, for $1\leq j\leq n_{c}$, since
\begin{align*}	\|e'_{j}(\hat{\Pi}_{c}-\Pi_{c}\mathcal{P}_{c})\|_{1}&=\|\frac{e'_{j}\hat{Z}_{c}}{\|e'_{j}\hat{Z}_{c}\|_{1}}-\frac{e'_{j}Z_{c}\mathcal{P}_{c}}{\|e'_{j}Z_{c}\mathcal{P}_{c}\|_{1}}\|_{1}=\|\frac{e'_{j}\hat{Z}_{c}\|e'_{j}Z_{c}\|_{1}-e'_{j}Z_{c}\mathcal{P}_{c}\|e'_{j}\hat{Z}_{c}\|_{1}}{\|e'_{j}\hat{Z}_{c}\|_{1}\|e'_{j}Z_{c}\|_{1}}\|_{1}\\	&=\|\frac{e'_{j}\hat{Z}_{c}\|e'_{j}Z_{c}\|_{1}-e'_{j}\hat{Z}_{c}\|e'_{j}\hat{Z}_{c}\|_{1}+e'_{j}\hat{Z}_{c}\|e'_{j}\hat{Z}_{c}\|_{1}-e'_{j}Z_{c}\mathcal{P}\|e'_{j}\hat{Z}_{c}\|_{1}}{\|e'_{j}\hat{Z}_{c}\|_{1}\|e'_{j}Z_{c}\|_{1}}\|_{1}\\
&\leq\frac{\|e'_{j}\hat{Z}_{c}\|e'_{j}Z_{c}\|_{1}-e'_{j}\hat{Z}_{c}\|e'_{j}\hat{Z}_{c}\|_{1}\|_{1}+\|e'_{j}\hat{Z}_{c}\|e'_{j}\hat{Z}_{c}\|_{1}-e'_{j}Z_{c}\mathcal{P}_{c}\|e'_{j}\hat{Z}_{c}\|_{1}\|_{1}}{\|e'_{j}\hat{Z}_{c}\|_{1}\|e'_{j}Z_{c}\|_{1}}\\	&=\frac{\|e'_{j}\hat{Z}_{c}\|_{1}|\|e'_{j}Z_{c}\|_{1}-\|e'_{j}\hat{Z}_{c}\|_{1}|+\|e'_{j}\hat{Z}_{c}\|_{1}\|e'_{j}\hat{Z}_{c}-e'_{j}Z_{c}\mathcal{P}_{c}\|_{1}}{\|e'_{j}\hat{Z}_{c}\|_{1}\|e'_{j}Z_{c}\|_{1}}\\ &=\frac{|\|e'_{j}Z_{c}\|_{1}-\|e'_{j}\hat{Z}_{c}\|_{1}|+\|e'_{j}\hat{Z}_{c}-e'_{j}Z_{c}\mathcal{P}_{c}\|_{1}}{\|e'_{j}Z_{c}\|_{1}}\leq\frac{2\|e'_{j}(\hat{Z}_{c}-Z_{c}\mathcal{P}_{c})\|_{1}}{\|e'_{j}Z_{c}\|_{1}}\\
&=\frac{2\|e'_{j}(\hat{Z}_{c}-Z_{c}\mathcal{P}_{c})\|_{1}}{\|e'_{j}\Pi_{c}\|_{1}}=2\|e'_{j}(\hat{Z}_{c}-Z_{c}\mathcal{P}_{c})\|_{1}\leq 2\sqrt{K}\|e'_{j}(\hat{Z}_{c}-Z_{c}\mathcal{P}_{c})\|_{F},
\end{align*}
by Lemma 4.5, we have
\begin{align*}	\|e'_{j}(\hat{\Pi}_{c}-\Pi_{c}\mathcal{P}_{c})\|_{1}=O(\sqrt{K}\|e'_{j}(\hat{Z}_{c}-Z_{c}\mathcal{P}_{c})\|_{F})=O(\varpi K\kappa(\Pi'_{c}\Pi_{c})\sqrt{\lambda_{1}(\Pi'_{c}\Pi_{c})}).
\end{align*}
For row nodes $1\leq i\leq n_{r}$, recall that $Z_{r}=Y_{*}J_{*}\equiv N^{-1}_{U}N_{M}\Pi_{r}, \hat{Z}_{r}=\hat{Y}_{*}\hat{J}_{*}, \Pi_{r}(i,:)=\frac{Z_{r}(i,:)}{\|Z_{r}(i,:)\|_{1}}$ and $\hat{\Pi}_{r}(i,:)=\frac{\hat{Z}_{r}(i,:)}{\|\hat{Z}_{r}(i,:)\|_{1}}$ where $N_{M}$ and $M$ are defined in the proof of Lemma 3.1 such that $U=\Theta_{r}M\equiv\Theta_{r}\Pi_{r}B_{r}$ and $N_{M}(i,i)=\frac{1}{\|M(i,:)\|_{F}}$, similar as the proof for column nodes, we have
\begin{align*}	\|e'_{i}(\hat{\Pi}_{r}-\Pi_{r}\mathcal{P}_{r})\|_{1}\leq\frac{2\|e'_{i}(\hat{Z}_{r}-Z_{r}\mathcal{P}_{r})\|_{1}}{\|e'_{i}Z_{r}\|_{1}}\leq \frac{2\sqrt{K}\|e'_{i}(\hat{Z}_{r}-Z_{r}\mathcal{P}_{r})\|_{F}}{\|e'_{i}Z_{r}\|_{1}}.
\end{align*}
Now, we provide a lower bound of $\|e'_{i}Z_{r}\|_{1}$ as below
\begin{align*}
\|e'_{i}Z_{r}\|_{1}&=\|e'_{i}N^{-1}_{U}N_{M}\Pi_{r}\|_{1}=\|N_{U}^{-1}(i,i)e'_{i}N_{M}\Pi_{r}\|_{1}=N^{-1}_{U}(i,i)\|N_{M}(i,i)e'_{i}\Pi_{r}\|_{1}=\frac{N_{M}(i,i)}{N_{U}(i,i)}\\
&=\|U(i,:)\|_{F}N_{M}(i,i)=\|U(i,:)\|_{F}\frac{1}{\|M(i,:)\|_{F}}=\|U(i,:)\|_{F}\frac{1}{\|e'_{i}M\|_{F}}=\|U(i,:)\|_{F}\frac{1}{\|e'_{i}\Theta^{-1}_{r}U\|_{F}}\\
&=\|U(i,:)\|_{F}\frac{1}{\|\Theta^{-1}_{r}(i,i)e'_{i}U\|_{F}}=\theta_{r}(i)\geq \theta_{r,\mathrm{min}}.
\end{align*}
Therefore, by Lemma 4.5, we have
\begin{align*}	\|e'_{i}(\hat{\Pi}_{r}-\Pi_{r}\mathcal{P}_{r})\|_{1}&\leq \frac{2\sqrt{K}\|e'_{i}(\hat{Z}_{r}-Z_{r}\mathcal{P}_{r})\|_{F}}{\|e'_{i}Z_{r}\|_{1}}\leq\frac{2\sqrt{K}\|e'_{i}(\hat{Z}_{r}-Z_{r}\mathcal{P}_{r})\|_{F}}{\theta_{r,\mathrm{min}}}\\
&=O(\frac{K^{5.5}\theta^{15}_{r,\mathrm{max}}\varpi\kappa^{4.5}(\Pi'_{r}\Pi_{r})\kappa(\Pi_{c})\lambda^{1.5}_{1}(\Pi'_{r}\Pi_{r})}{\theta^{15}_{r,\mathrm{min}}\pi_{r,\mathrm{min}}}).
\end{align*}
\end{proof}
\subsection{Proof of Corollary 4.7}
\begin{proof}
Under  conditions of Corollary 4.7, we have
\begin{align*}
&\|e'_{i}(\hat{\Pi}_{r}-\Pi_{r}\mathcal{P}_{r})\|_{1}=O(\frac{K^{5.5}\theta^{15}_{r,\mathrm{max}}\varpi\kappa^{4.5}(\Pi'_{r}\Pi_{r})\kappa(\Pi_{c})\lambda^{1.5}_{1}(\Pi'_{r}\Pi_{r})}{\theta^{15}_{r,\mathrm{min}}\pi_{r,\mathrm{min}}})=O(\frac{\theta^{15}_{r,\mathrm{max}}\varpi\sqrt{n_{r}}}{\theta^{15}_{r,\mathrm{min}}}),\\
&\|e'_{j}(\hat{\Pi}_{c}-\Pi_{c}\mathcal{P}_{c})\|_{1}=O(\varpi K\kappa(\Pi'_{c}\Pi_{c})\sqrt{\lambda_{1}(\Pi'_{c}\Pi_{c})})=O(\varpi\sqrt{n_{c}}).
\end{align*}
Under  conditions of Corollary 4.7, Lemma 4.3 gives $\varpi=O(\frac{\sqrt{\theta_{r,\mathrm{max}}\mathrm{log}(n_{r}+n_{c})}}{\theta_{r,\mathrm{min}}\sigma_{K}(P)\sqrt{n_{r}n_{c}}})$,
which gives that
\begin{align*}
&\|e'_{i}(\hat{\Pi}_{r}-\Pi_{r}\mathcal{P}_{r})\|_{1}=O(\frac{\theta^{15}_{r,\mathrm{max}}\varpi\sqrt{n_{r}}}{\theta^{15}_{r,\mathrm{min}}})=O(\frac{\theta^{15.5}_{r,\mathrm{max}}\sqrt{\mathrm{log}(n_{r}+n_{c})}}{\theta^{16}_{r,\mathrm{min}}\sigma_{K}(P)\sqrt{n_{c}}}),\\
&\|e'_{j}(\hat{\Pi}_{c}-\Pi_{c}\mathcal{P}_{c})\|_{1}=O(\varpi\sqrt{n_{c}})=O(\frac{\sqrt{\theta_{r,\mathrm{max}}\mathrm{log}(n_{r}+n_{c})}}{\theta_{r,\mathrm{min}}\sigma_{K}(P)\sqrt{n_{r}}}).
\end{align*}
By basic algebra, this corollary follows.
\end{proof}
\section{SVM-cone algorithm}\label{introduceSVMcone}
For readers' convenience, we briefly introduce the SVM-cone algorithm given in \cite{MaoSVM} and provide another view that SVM-cone algorithm exactly recovers $\Pi_{r}$ when the input is $U_{*}$ (or $U_{*,2}$). Let $S$ be a matrix whose rows have unit $l_{2}$ norm and $S$ can be written as
$S=HS_{C}$, where $H\in\mathrm{R}^{n\times K}$ with nonnegative entries, no row of $H$ is 0, and $S_{C}\in\mathbb{R}^{K\times m}$ corresponding to $K$ rows of $S$ (i.e., there exists an index set $\mathcal{I}$ with $K$ entries such that $S_{C}=S(\mathcal{I},:)$). Inferring $H$ from $S$ is called the Ideal Cone problem, i.e., Problem 1 in \cite{MaoSVM}. The Ideal Cone problem can be solved by applying one-class SVM to the rows of $S$, and the $K$ rows of $S_{C}$ are the support vectors found by one-class SVM:
\begin{align}\label{OneClassSVM}
\mathrm{maximize~}b~~\mathrm{s.t.}~~\textbf{w}'S(i,:)\geq b(\mathrm{~for~}i=1,2,\ldots,n)~\mathrm{and~~}\|\textbf{w}\|_{F}\leq 1.
\end{align}
The solution  $(\textbf{w}, b)$ for the Ideal Cone problem when $(S_{C}S'_{C})^{-1}\mathbf{1}>0$ is given by
\begin{align}\label{SolutionOfOneClassSVM}
\textbf{w}=b^{-1}\cdot S'_{C}\frac{(S_{C}S'_{C})^{-1}\mathbf{1}}{\mathbf{1}'(S_{C}S'_{C})^{-1}\mathbf{1}},~~~ b=\frac{1}{\sqrt{\mathbf{1}'(S_{C}S'_{C})^{-1}\mathbf{1}}}.
\end{align}
For the empirical case, let $\hat{S}\in\mathbb{R}^{n\times m}$ be a matrix with all rows have unit $l_{2}$ norm, infer $H$ from $\hat{S}$ with given $K$ is called the empirical cone problem i.e., Problem 2 in \cite{MaoSVM}. For the empirical cone problem, one-class SVM is applied to all rows of $\hat{S}$ to obtain $\textbf{w}$ and $b$'s estimations $\hat{\textbf{w}}$ and $\hat{b}$. Then apply K-means algorithm to rows of $\hat{S}$ that are close to the hyperplane into $K$ clusters, and an estimation of the  index set $\mathcal{I}$ can be obtained from the $K$ clusters provides. Algorithm \ref{alg:SVMcone} below is the SVM-cone algorithm provided in \cite{MaoSVM}.
\begin{algorithm}
\caption{\textbf{SVM-cone}\cite{MaoSVM}}
\label{alg:SVMcone}
\begin{algorithmic}[1]
\Require $\hat{S}\in \mathbb{R}^{n\times m}$ with rows have unit $l_{2}$ norm, number of corners $K$, estimated distance corners from hyperplane $\gamma$.
\Ensure The near-corner index set $\mathcal{\hat{I}}$.
\State Run one-class SVM on $\hat{S}(i,:)$ to get $\hat{\textbf{w}}$ and $\hat{b}$
\State Run K-means algorithm to the set $\{\hat{S}(i,:)| \hat{S}(i,:)\hat{\textbf{w}}\leq \hat{b}+\gamma\}$ that are close to the hyperplane into $K$ clusters
\State Pick one point from each cluster to get the near-corner set $\mathcal{\hat{I}}$
\end{algorithmic}
\end{algorithm}
As suggested in \cite{MaoSVM}, we can start $\gamma=0$ and incrementally increase it until $K$ distinct clusters are found.

Now turn to our DiMSC algorithm and focus on estimating $\mathcal{I}_{r}$ with given $U_{*}, U_{*,2}$ and $K$. By Lemmas 3.1 and 3.4, we know that $U_{*}$ and $U_{*,2}$ enjoy the Ideal Cone structure, and Lemma 3.2 guarantees that one-class SVM can be applied to rows of  $U_{*}$ and $U_{*,2}$. Set $\textbf{w}_{1}=b_{1}^{-1}U'_{*}(\mathcal{I}_{r},:)\frac{(U_{*}(\mathcal{I}_{r},:)U'_{*}(\mathcal{I}_{r},:))^{-1}\mathbf{1}}{\mathbf{1}'(U_{*}(\mathcal{I}_{r},:)U'_{*}(\mathcal{I}_{r},:))^{-1}}, b_{1}=\frac{1}{\sqrt{\mathbf{1}'(U_{*}(\mathcal{I}_{r},:)U'_{*}(\mathcal{I}_{r},:))^{-1}\mathbf{1}}}$, and $\textbf{w}_{2}=b_{2}^{-1}U'_{*,2}(\mathcal{I}_{r},:)\frac{(U_{*,2}(\mathcal{I}_{r},:)U'_{*,2}(\mathcal{I}_{r},:))^{-1}\mathbf{1}}{\mathbf{1}'(U_{*,2}(\mathcal{I}_{r},:)U'_{*,2}(\mathcal{I}_{r},:))^{-1}}, b_{2}=\frac{1}{\sqrt{\mathbf{1}'(U_{*,2}(\mathcal{I}_{r},:)U'_{*,2}(\mathcal{I}_{r},:))^{-1}\mathbf{1}}}$. Now that $\textbf{w}_{1}$ and $b_{1}$ are solutions of the one-class SVM in Eq (\ref{OneClassSVM}) by setting $S=U_{*}$, and $\textbf{w}_{2}$ and $b_{2}$ are solutions of the one-class SVM in Eq (\ref{OneClassSVM}) by setting $S=U_{*,2}$ . Lemma \ref{WhyUseKmeansInSVMcone} says that if row node $i$ is a pure node, we have $U_{*}(i,:)\textbf{w}_{1}=b_{1}$, and this suggests that in the SVM-cone algorithm, if the input matrix is $U_{*}$, by setting $\gamma=0$, we can find all pure row nodes, i.e., the set $\{U_{*}(i,:)|U_{*}(i,:)\textbf{w}_{1}=b_{1}\}$ contains all rows of $U_{*}$ respective to pure row nodes while including mixed row nodes. By Lemma 3.1, these  pure row nodes belong to the $K$ distinct row communities such that if row nodes $i,\bar{i}$ are in the same row community, then we have $U_{*}(i,:)=U_{*}(\bar{i},:)$, and this is the reason that we need to apply K-means algorithm on the set obtained in step 2 in the SVM-cone algorithm to obtain the $K$ distinct row communities, and this is also the reason that we said SVM-cone algorithm returns the index set $\mathcal{I}$ exactly when the input is $U_{*}$. These conclusions also hold when we set the input in the SVM-cone algorithm as $U_{*,2}$.
\begin{lem}\label{P1}
Under $DiDCMM_{n_{r},n_{c}}(K,P,\Pi_{r}, \Pi_{c}, \Theta_{r})$, for $1\leq i\leq n_{r}$, $U_{*}(i,:)$ can be written as $U_{*}(i,:)=r_{1}(i)\Phi_{1}(i,:)U_{*}(\mathcal{I}_{r},:)$, where $r_{1}(i)\geq 1$. Meanwhile, $r_{1}(i)=1$ and $\Phi_{1}(i,:)=e'_{k}$ if $i$ is a pure node such that $\Pi_{r}(i,k)=1$; $r_{1}(i)>1$ and $\Phi_{1}(i,:)\neq e'_{k}$ if $\Pi_{r}(i,k)<1$ for $1\leq k\leq K$. Similarly, $U_{*,2}(i,:)$ can be written as $U_{*,2}(i,:)=r_{2}(i)\Phi_{2}(i,:)U_{*,2}(\mathcal{I}_{r},:)$, where $r_{2}(i)\geq 1$. Meanwhile, $r_{2}(i)=1$ and $\Phi_{2}(i,:)=e'_{k}$ if $\Pi_{r}(i,k)=1$; $r_{2}(i)>1$ and $\Phi_{2}(i,:)\neq e'_{k}$ if $\Pi_{r}(i,k)<1$ for $1\leq k\leq K$.
\end{lem}
\begin{proof}
Since $U_{*}=YU_{*}(\mathcal{I}_{r},:)$ by Lemma 3.1, for $1\leq i\leq n_{r}$, we have
\begin{align*}
U_{*}(i,:)=Y(i,:)U_{*}(\mathcal{I}_{r},:)=Y(i,:)\mathbf{1}\frac{Y(i,:)}{Y(i,:)\mathbf{1}}U_{*}(\mathcal{I}_{r},:)=r_{1}(i)\Phi_{1}(i,:)U_{*}(\mathcal{I}_{r},:),
\end{align*}
where we set $r_{1}(i)=Y(i,:)\textbf{1}$, $\Phi_{1}(i,:)=\frac{Y(i,:)}{Y(i,:)\mathbf{1}}$, and $\mathbf{1}$ is a $K\times 1$  vector with all entries being ones.

By the proof of Lemma 3.1, $Y(i,:)=\frac{\Pi_{r}(i,:)}{\|M(i,:)\|_{F}}\Theta^{-1}_{r}(\mathcal{I}_{r},\mathcal{I}_{r})N^{-1}_{U}(\mathcal{I}_{r},\mathcal{I}_{r})$, where $M=\Pi_{r}\Theta^{-1}_{r}(\mathcal{I}_{r},\mathcal{I}_{r})U(\mathcal{I}_{r},:)$.  For convenience, set $T=\Theta^{-1}_{r}(\mathcal{I}_{r},\mathcal{I}_{r}), Q=N^{-1}_{U}(\mathcal{I}_{r},\mathcal{I}_{r})$, and $R=U(\mathcal{I}_{r},:)$ (such setting of $T,Q, R$ is only for used for notation convenience for the proof of Lemma \ref{P1}).

On the one hand, if row node $i$ is pure such that $\Pi_{r}(i,k)=1$ for certain $k$ among $\{1,2,\ldots,K\}$ (i.e., $\Pi_{r}(i,:)=e_{k}$ if $\Pi_{r}(i,k)=1$), we have $M(i,:)=\Pi_{r}(i,:)\Theta^{-1}_{r}(\mathcal{I}_{r},\mathcal{I}_{r})U(\mathcal{I}_{r},:)=T(k,k)R(k,:)$, and $\Pi_{r}(i,:)TQ=T(k,k)Q(k,:)$, which give that $Y(i,:)=\frac{T(k,k)Q(k,:)}{\|T(k,k)R(k,:)\|_{F}}=\frac{Q(k,:)}{\|R(k,:)\|_{F}}$. Recall that the $k$-th diagonal entry of $N^{-1}_{U}(\mathcal{I}_{r},\mathcal{I}_{r})$ is $\|[U(\mathcal{I}_{r},:)](k,:)\|_{F}$, i.e., $Q(k,:)\mathbf{1}=\|R(k,:)\|_{F}$, which gives that $r_{1}(i)=Y(i,:)\mathbf{1}=1$ and $\Phi_{1}(i,:)=e'_{k}$ when $\Pi_{r}(i,k)=1$.

On the other hand, if $i$ is a mixed node, since $\|M(i,:)\|_{F}=\|\Pi_{r}(i,:)\Theta^{-1}_{r}(\mathcal{I}_{r},\mathcal{I}_{r})U(\mathcal{I}_{r},:)\|_{F}=\|\sum_{k=1}^{K}\Pi_{r}(i,k)T(k,k)R(k,:)\|_{F}< \sum_{k=1}^{K}\Pi_{r}(i,k)T(k,k)\|R(k,:)\|_{F}=\sum_{k=1}^{K}\Pi_{r}(i,k)T(k,k)Q(k,k)$, combine it with $\Pi_{r}(i,:)TQ\mathbf{1}=\sum_{k=1}^{K}\Pi_{r}(i,k)T(k,k)Q(k,k)$, so $r_{1}(i)=Y(i,:)\mathbf{1}=\frac{\Pi_{r}(i,:)TQ\mathbf{1}}{\|M(i,:)\|_{F}}> 1$. The lemma follows by similar analysis for $U_{*,2}$.
\end{proof}
\begin{lem}\label{WhyUseKmeansInSVMcone}
Under $DiDCMM_{n_{r},n_{c}}(K,P,\Pi_{r}, \Pi_{c}, \Theta_{r})$, for $1\leq i\leq n_{r}$, if row node $i$ is a pure node such that $\Pi_{r}(i,k)=1$ for certain $k$, we have
\begin{align*}
U_{*}(i,:)\textbf{w}_{1}=b_{1}\mathrm{~~~and~~~}U_{*,2}(i,:)\textbf{w}_{2}=b_{2},
\end{align*}
Meanwhile, if row node $i$ is a mixed node, the above equalities do not hold.
\end{lem}
\begin{proof}
For the claim that $U_{*}(i,:)\textbf{w}_{1}=b_{1}$ holds when $i$ is pure, by Lemma \ref{P1}, when $i$ is a pure node such that $\Pi_{r}(i,k)=1$, $U_{*}(i,:)$ can be written as $U_{*}(i,:)=e'_{k}U_{*}(\mathcal{I}_{r},:)$, so $U_{*}(i,:)\textbf{w}_{1}=b_{1}$ holds surely. When $i$ is a mixed node, by Lemma \ref{P1}, $r_{1}(i)>1$ and $\Phi_{1}(i,:)\neq e_{k}$ for any $k=1,2,\ldots, K$, hence $U_{*}(i,:)\neq e'_{k}U_{*}(\mathcal{I}_{r},:)$ if $i$ is mixed, which gives the result. Follow similar analysis, we obtain the results associated with $U_{*,2}$, and the lemma follows.
\end{proof}

\bibliographystyle{agsm}
\bibliography{reference}